\documentclass[journal=jcisd8,manuscript=article]{achemso}

\usepackage{chemformula} 
\usepackage[T1]{fontenc} 
\usepackage{amssymb}
\usepackage{amsmath}
\usepackage{bbm}
\usepackage{subcaption,float,placeins}
\usepackage[flushleft]{threeparttable}
\usepackage{multirow,booktabs}
\usepackage{xspace}
\usepackage{verbatim} 
\usepackage{comment, mathtools}
\usepackage{pgfplots, pgfplotstable}
\pgfplotsset{width=7.5cm,compat=1.12}
\pgfplotsset{compat=1.15, width=\textwidth}
\usepackage{longtable, threeparttablex}
\usepackage{graphicx}
\usepackage{soul}
\usepackage{siunitx}

\usepackage[colorinlistoftodos]{todonotes}

\SectionNumbersOn

\author{Vishal Dey}
\affiliation
{Department of Computer Science and Engineering, The Ohio State University, Columbus, OH}
\author{Raghu Machiraju}
\affiliation
{Department of Computer Science and Engineering, The Ohio State University, Columbus, OH}
\alsoaffiliation[The Ohio State University, Columbus, OH]
{Biomedical Informatics, The Ohio State University, Columbus, OH}
\alsoaffiliation
{Translational Data Analytics Institute, The Ohio State University, 
Columbus, OH}
\author{Xia Ning}
\affiliation
{Department of Computer Science and Engineering, The Ohio State University, Columbus, OH}
\alsoaffiliation[The Ohio State University, Columbus, OH]
{Biomedical Informatics, The Ohio State University, Columbus, OH}
\alsoaffiliation
{Translational Data Analytics Institute, The Ohio State University, 
Columbus, OH}
\email{ning.104@osu.edu}
\title[An \textsf{achemso} demo]
{Improving Compound Activity Classification via Deep Transfer and Representation Learning}

\newcommand{\RankMethod}{\mbox{$\mathop{\mathtt{gnnCP}}\limits$}\xspace}
\newcommand{\OurMethod}{\mbox{$\mathop{\mathtt{TAc}}\limits$}\xspace}
\newcommand{\OurMethodWithDisc}{\mbox{$\mathop{\mathtt{TAc\text{-}fc}}\limits$}\xspace}

\newcommand{\OurMethodWithLocalDisc}{\mbox{$\mathop{\mathtt{TAc\text{-}f}}\limits$}\xspace}
\newcommand{\OurMethodWithGlobalDisc}{\mbox{$\mathop{\mathtt{TAc\text{-}c}}\limits$}\xspace}

\newcommand{\OurMethodWithDMPNNAWithGlobalDisc}{\mbox{$\mathop{\mathtt{TAc\text{-}c\text{-}\newgnn}}\limits$}\xspace}

\newcommand{\OurMethodWithDMPNN}{\mbox{$\mathop{\OurMethod\text{-}\dmpnn}\limits$}\xspace}
\newcommand{\OurMethodWithDMPNNA}{\mbox{$\mathop{\OurMethod\text{-}\newgnn}\limits$}\xspace}
\newcommand{\OurMethodWithDMPNNWithDisc}{\mbox{$\mathop{\OurMethodWithDisc\text{-}\dmpnn}\limits$}\xspace}
\newcommand{\OurMethodWithDMPNNAWithDisc}{\mbox{$\mathop{\OurMethodWithDisc\text{-}\newgnn}\limits$}\xspace}
\newcommand{\gnn}{\mbox{$\mathop{\mathtt{GNN}}\limits$}\xspace}
\newcommand{\gnns}{\mbox{$\mathop{\mathtt{GNNs}}\limits$}\xspace}

\newcommand{\morganFCN}{\mbox{$\mathop{\mathtt{FCN}\text{-}\mathtt{morgan}}\limits$}\xspace}
\newcommand{\morgancFCN}{\mbox{$\mathop{\mathtt{FCN}\text{-}\mathtt{morganc}}\limits$}\xspace}
\newcommand{\dmpnnFCN}{\mbox{$\mathop{\mathtt{FCN}\text{-}\mathtt{\dmpnn}}\limits$}\xspace}
\newcommand{\newgnnFCN}{\mbox{$\mathop{\mathtt{FCN}\text{-}\mathtt{\newgnn}}\limits$}\xspace} 

\newcommand{\dmpnnFCNDT}{\mbox{$\mathop{\mathtt{FCN}\text{-}\mathtt{\dmpnn} ~(\datatransfer)}\limits$}\xspace}
\newcommand{\newgnnFCNDT}{\mbox{$\mathop{\mathtt{FCN}\text{-}\mathtt{\newgnn} ~(\datatransfer)}\limits$}\xspace} 

\newcommand{\DANN}{\mbox{$\mathop{\mathtt{DANN}}\limits$}\xspace}
\newcommand{\DANNWithDMPNN}{\mbox{$\mathop{\DANN\text{-}\dmpnn}\limits$}\xspace}
\newcommand{\DANNWithDMPNNA}{\mbox{$\mathop{\DANN\text{-}\newgnn}\limits$}\xspace}

\newcommand{\datatransfer}{\mbox{$\mathop{\mathtt{DT}}\limits$}\xspace}
\newcommand{\notransfer}{\mbox{$\mathop{\mathtt{noT}}\limits$}\xspace}

\newcommand{\SourceDomain}{\mbox{$\mathop{\mathcal{D}^{(S)}}\limits$}\xspace}
\newcommand{\TargetDomain}{\mbox{$\mathop{\mathcal{D}^{(T)}}\limits$}\xspace}
\newcommand{\SourceTask}{\mbox{$\mathop{\mathcal{T}^{(S)}}\limits$}\xspace}
\newcommand{\TargetTask}{\mbox{$\mathop{\mathcal{T}^{(T)}}\limits$}\xspace}

\newcommand{\source}{\mbox{$\mathop{\mathtt{Src}}\limits$}\xspace}
\newcommand{\target}{\mbox{$\mathop{\mathtt{Tgt}}\limits$}\xspace}

\newcommand{\prauc}{\mbox{$\mathop{\mathtt{PR\text{-}AUC}}\limits$}\xspace}
\newcommand{\rocauc}{\mbox{$\mathop{\mathtt{ROC\text{-}AUC}}\limits$}\xspace}
\newcommand{\precision}{\mbox{$\mathop{\mathtt{precision}}\limits$}\xspace}

\newcommand{\sensitivity}{\mbox{$\mathop{\mathtt{sens}}\limits$}\xspace}

\newcommand{\accuracy}{\mbox{$\mathop{\mathtt{accuracy}}\limits$}\xspace}
\newcommand{\fone}{\mbox{$\mathop{\mathtt{F1}}\limits$}\xspace}

\newcommand{\dmpnn}{\mbox{$\mathop{\mathtt{dmpn}}\limits$}\xspace}
\newcommand{\MORGAN}{\mbox{$\mathop{\mathtt{morgan}}\limits$}\xspace}
\newcommand{\MORGANC}{\mbox{$\mathop{\mathtt{morgan\textrm{-}c}}\limits$}\xspace}
\newcommand{\MORGANBA}{\mbox{$\mathop{\mathtt{morgan\textrm{-}ba}}\limits$}\xspace}
\newcommand{\RDKITGD}{\mbox{$\mathop{\mathtt{RDKit200}}\limits$}\xspace}

\newcommand{\newgnn}{\mbox{$\mathop{\mathtt{dmpna}}\limits$}\xspace}

\newcommand{\fgen}{\mbox{$\mathtt{F}$}\xspace}
\newcommand{\ldisc}{\mbox{$\mathtt{L}$}\xspace}
\newcommand{\gdisc}{\mbox{$\mathtt{G}$}\xspace}
\newcommand{\classifier}{\mbox{$\mathtt{S}$}\xspace}
\newcommand{\fatt}{\mbox{$f_a$}\xspace}

\newcommand{\ParGnn}{\mbox{$\Omega$}\xspace}
\newcommand{\ParLdisc}{\mbox{$\Theta$}\xspace}
\newcommand{\ParGdisc}{\mbox{$\Psi$}\xspace}
\newcommand{\ParClassifier}{\mbox{$\Phi$}\xspace}

\newcommand{\MolGraph}{\mbox{${\mathcal{G}}$}\xspace}
\newcommand{\compound}{\mbox{$c$}\xspace}
\newcommand{\bioassay}{\mbox{$B$}\xspace}
\newcommand{\SetAtoms}{\mbox{$\mathcal{A}$}\xspace}
\newcommand{\SetBonds}{\mbox{$\mathcal{E}$}\xspace}
\newcommand{\SetCompounds}{\mbox{$\mathcal{X}$}\xspace}
\newcommand{\SetLabels}{\mbox{$\mathcal{Y}$}\xspace}

\newcommand{\InputFeature}{\mbox{${\mathbf{x}}$}\xspace}
\newcommand{\Label}{\mbox{$y$}\xspace}

\newcommand{\FeatureSpace}{\mbox{${\mathbb{X}}$}\xspace}
\newcommand{\LabelSpace}{\mbox{${\mathbb{Y}}$}\xspace}

\newcommand{\domain}{\mbox{${\mathcal{D}}$}\xspace}
\newcommand{\task}{\mbox{${\mathcal{T}}$}\xspace}
\newcommand{\entropy}{\mbox{${H}$}\xspace}

\newcommand{\embedding}{\mbox{${\mathbf{s}}$}\xspace}

\newcommand{\hidden}{\mbox{${\mathbf{h}}$}\xspace}
\newcommand{\compr}{\mbox{${\mathbf{r}}$}\xspace}
\newcommand{\scompr}{\mbox{${\mathbf{z}}$}\xspace}

\newcommand{\FinalLoss}{\mbox{$\mathcal{L}$}\xspace}
\newcommand{\ClassifierLoss}{\mbox{$\mathcal{L}_{(c)}$}\xspace}
\newcommand{\LocalDiscLoss}{\mbox{$\mathcal{L}_{(l)}$}\xspace}
\newcommand{\GlobalDiscLoss}{\mbox{$\mathcal{L}_{(g)}$}\xspace}

\newcommand{\numImpv}{\mbox{$\mathop{\mathtt{N\textrm{-}impv}}\limits$}\xspace}
\newcommand{\taskImpv}{\mbox{$\mathop{\mathtt{t\textrm{-}impv\%}}\limits$}\xspace}
\newcommand{\diff}{\mbox{$\mathop{\mathtt{diff\%}}\limits$}\xspace}
\newcommand{\taskDiff}{\mbox{$\mathop{\mathtt{t\textrm{-}diff\%}}\limits$}\xspace}

\newcommand{\pvalue}{\mbox{$\mathop{\mathtt{p\textrm{-}value}}\limits$}\xspace}

\usepackage{xr}
\begin{document}

%
%
%
%
%


\begin{abstract}
Recent advances in molecular machine learning, 
especially deep neural networks such as Graph Neural Networks (\gnns) 
for predicting structure activity relationships (SAR)
have shown tremendous potential in computer-aided drug discovery.
%
However, the applicability of such deep neural networks are limited by 
the requirement of large amounts of training data.
In order to cope with limited training data for a target task,
transfer learning for SAR modeling has been recently adopted
to leverage information from data of related tasks.
%
In this work, in contrast to the popular parameter-based transfer learning such as pretraining,
we develop novel deep transfer learning methods \OurMethod and \OurMethodWithDisc
to leverage source domain data
and transfer useful information to the target domain.
\OurMethod learns to generate effective molecular features that can generalize well
from one domain to another, and increase the classification performance in the target domain.
Additionally, \OurMethodWithDisc extends \OurMethod by incorporating novel components
to selectively learn feature-wise and compound-wise transferability.
We used the bioassay screening data from PubChem, and identified 120 pairs of bioassays
such that the active compounds in each pair are more similar to each other
compared to its inactive compounds.
Overall, \OurMethod achieves the best performance with average \rocauc of 0.801;
it significantly improves \rocauc of 83\% target tasks with 
average task-wise performance improvement of 7.102\%,
compared to the best baseline \newgnnFCNDT. 
Our experiments clearly demonstrate that \OurMethod achieves significant 
improvement over all baselines across a large number of target tasks. 
Furthermore, although \OurMethodWithDisc achieves slightly worse
\rocauc on average compared to \OurMethod (0.798 vs 0.801), 
\OurMethodWithDisc still achieves the best performance on more tasks in terms of \prauc and \fone
compared to other methods.
In summary, \OurMethodWithDisc is also found to be a strong model with competitive or
even better performance than \OurMethod on a notable number of target tasks.
\end{abstract}

\section{Introduction}
\label{sec:intro}

Drug discovery is a time consuming and expensive process~\cite{dickson2009cost} --
it takes at least 10 years and at least \$1 billion to fully develop a drug~\cite{DiMasi2003}.
During the initial stages of this process, promising drug candidates
are identified by screening large library of chemical compounds,
and are then further investigated for specific properties.
In order to speed up this process, 
computational approaches~\cite{Terstappen2001, Sliwoski2014} have been adopted, particularly
for identifying potential drug candidates during the initial stages of
the drug discovery.
Computational approaches explore a much larger space of chemical compounds
to predict their physio-chemical properties and/or biological activities towards the target.
In this paper, we consider the problem of compound bioactivity classification,
where a compound is classified as active or inactive based on whether that compound
binds to the protein target. 
Biological activities of compounds are initially examined in a bioassay by measuring 
their binding affinities or dissociation constants toward the target.
Significant research~\cite{Hansch1962, Debnath1991, SupertiFurga1995} have established the relationship between
the chemical structures and biological activities of compounds, also known as
Structure-Activity relationships (SAR)~\cite{Hansch1962}.
Several computational approaches~\cite{Dudek2006} have been developed to model
SAR
and to predict compound bioactivities from their 2D/3D structures.	
However, most popular approaches such as deep neural networks require
large amounts of labeled data for effective SAR modeling.
Thus, limited availability of bioassay data for specific targets still
pose a major challenge in effective SAR modeling~\cite{Imrie2018}.
%

%
Over the years, several methods~\cite{Ning2009, Liu2017Multi, Liu2017Diff} 
aimed to improve SAR predictions for specific targets
by leveraging activity information from related targets. 
These methods consider targets to be related,
based on the principles from 
chemogenomics~\cite{Murcko2001, Kubinyi2005, Harris2006, Klabunde2007}.
\emph{The key principle behind these methods is that
similar proteins tend to bind to structurally similar compounds.}
In this work, we consider proteins belonging to the same protein family to be similar.
%
Thus, leveraging compound activity information
from bioassays corresponding to a set of proteins from the same protein family
(e.g., G-coupled protein receptors, kinases, peptidases, etc)
collectively might better inform the SAR model than the individual bioassays.
In essence, transfer learning can enable better SAR modeling by
leveraging information from such related bioassays.
%
%
However, 
existing methods are instance-based transfer learning methods~\cite{Pan2010}.
They select a subset of data from related bioassays,
and then augment the training data for the target task with the selected subset.
Existing deep transfer learning based methods~\cite{Cai2020}
for SAR modeling are either parameter-based (such as finetuning), 
or feature-based; out of which parameter-based methods
are more popular.
However, such methods can lead to overfitting and negative transfer~\cite{Pan2010},
especially when the targets are not related.
In this regard, we believe that feature-based methods are better in 
that they can learn the similarity/relatedness between the targets in the
latent space in a data-driven manner.

Primarily, we develop an instance-based transfer learning method \OurMethod
that leverages target information from related bioassays,
based on the key principle of chemogenomics as mentioned earlier.
%
%
We further extend \OurMethod to a novel feature-based 
deep transfer learning method \OurMethodWithDisc
that quantitatively measures transferability and explicitly learns what to
transfer in a fully data-driven manner.
To this end, we develop novel components to learn feature-wise and compound-wise
transferability in order to effectively encode the commonalities among compounds
of different tasks.
%
%
In order to represent compounds,
we leveraged the popular idea of Directed Message Passing Neural Network (\dmpnn)~\cite{Yang2019},
and add an attention-based pooling mechanism, denoted as \newgnn.
We collected a set of confirmatory bioassays from PubChem~\cite{kim2021pubchem} 
that have a single protein target,
and are tested on chemical substances.
We identified 120 bioassay pairs involving 59 protein targets 
such that the active compounds in each pair
are more similar to each other compared to the inactive compounds.
We compared our methods \OurMethod and \OurMethodWithDisc 
with several baselines with respect to two aspects: compound representation and
transfer mechanisms.
Overall, \OurMethodWithDMPNNA achieves the best performance compared to all other methods.
Compared to \OurMethodWithDMPNNA, \OurMethodWithDMPNNAWithDisc 
performs slightly worse, but the latter
still provides significant performance improvement on some target tasks.
This suggests that although the transfer mechanism in \OurMethod performs the best overall,
the deep transfer mechanism with learned feature-wise and compound-wise transferability
can actually benefit some targets.
Furthermore, experimental results demonstrate the efficacy of our proposed attention mechanism
of \newgnn in learning better compound features.
We provide additional experiments on the compound prioritization problem~\cite{Liu2017Diff}
where \newgnn clearly outperforms all other compound representation methods. 


The rest of the paper is organized as follows. Section~\ref{sec:related} presents the related works
in drug discovery, and transfer learning with applications in SAR predictions.
Section~\ref{sec:results} presents the materials used for experimental evaluation,
experimental results and detailed analyses with discussions.
Section~\ref{sec:conclusions} presents the conclusions.
Section~\ref{sec:comp_methods} presents the 
notations and definitions used in this paper,
and the proposed methods of transfer learning for activity prediction.


\subsection{Related Work}
\label{sec:related}


In this section, we provide a brief overview of existing works and divide them
across 3 subsections as follows. 
In section{~\ref{sec:related:comp}}, we summarize notable works on computational approaches in drug discovery. 
In section{~\ref{sec:related:transfer}}, we provide a brief overview of existing works in transfer learning. 
In section{~\ref{sec:related:transfer_sar}}, we provide an overview of existing methods
that uses transfer learning for better SAR modeling.

\subsubsection{Computational Methods in Drug Discovery}
\label{sec:related:comp}
The first step in the drug discovery process is to conduct bioassays~\cite{Atta-ur-Rahman2001} 
that screens a large set of compounds for desirable properties
(e.g., activity, solubility, toxicity).
The findings from these bioassays guide the later steps 
of the drug discovery process.
In order to speed up initial stages of the drug discovery process, 
computational approaches have been adopted. 
Computational approaches to predict activities/properties of
compounds from their molecular structures have been a significant
research area in cheminformatics~\cite{Dudek2006, TareqHassanKhan2010, Lo2018}.
These approaches rely on the quantitative structure activity/property relationship
(QSAR/QSPR)~\cite{Hansch1962, Katritzky2001} 
to predict compound activities/properties as expressed in bioassays.
In order to predict such activities/properties, 
machine learning methods such as classification and regression are typically used.
Binary/real-valued observations from bioassay data 
are used to train these classification/regression methods.
Popular conventional classification and regression methods 
to predict compound activities/properties
consist of 
support vector machines~\cite{Czermiski2001, Zernov2003, Hou2007, Alvarsson2016}, 
random forests~\cite{Svetnik2003, Zhang2007}, 
Bayesian models~\cite{Xia2004, Chen2011}, etc.
In these methods, compounds are typically represented by
hand-crafted molecular fingerprints~\cite{rogers2010extended, Durant2002} or descriptors~\cite{Randic1993}.
Recently, deep learning methods~\cite{Ma2015, Gawehn2016, Zhang2017, Chen2018, Hop2018} 
have demonstrated significant
performance improvement over conventional methods
across several activity/property prediction tasks~\cite{Klambauer2017, Wenzel2019, Wu2019}.
Unlike conventional methods, these methods do not require
careful and expensive design of
hand-crafted molecular fingerprints or descriptors by domain experts.
These methods learn the compound representations from 
molecular graphs~\cite{duvenaud2015convolutional, Kearnes2016, Gilmer2017, Yang2019, Sun2020, Withnall2020} and
SMILES strings~\cite{Zheng2019, Chakravarti2019, Karpov2020}, 
in a fully data-driven manner for each task.
Such learned representations are task-specific and can better encode relevant structures
for each task.
Thus, such learned representations are often more effective than
molecular fingerprints or descriptors.
%
%
While these deep learning models have achieved the state-of-the-art
performance on several molecular activity/property prediction tasks,
these models require large amount of labeled training data to encode
relevant patterns into learned representations.
%
Training these models with limited labeled data
for certain prediction tasks
often leads to sub-par performance. 

\subsubsection{Transfer Learning}
\label{sec:related:transfer}
In order to effectively train models with limited labeled data
for certain prediction tasks, 
transfer learning between related tasks 
has been widely explored in Computer Vision(CV)
and Natural Language Processing (NLP)~\cite{Tan2018, Zhuang2021}.
Transfer learning~\cite{Pan2010} is an emerging research area 
in which knowledge gained from auxiliary tasks
is transferred to improve the predictive performance 
of the target task. 
Instead of training a model for the target task from scratch, 
a popular transfer learning technique, called finetuning~\cite{Yosinski2014}, finetunes the model
pretrained from other related tasks.
Pretraining does not explicitly learn what/when to transfer
and rather relies on the model parameters to encode and
transfer information across different tasks.
Although pretraining is the most popular transfer learning method,
it does not guarantee improvement (due to `negative transfer'~\cite{Wang2019}). 
Moreover, finetuning a highly parameterized model with limited data
may lead to overfitting to the training data and 
thus, the finetuned model might not generalize well to the test data. 
Apart from pretraining, another area of deep transfer learning, called 
domain adaptation has gained a lot of attention~\cite{Pan2011, Long2013, Wang2018}.
Domain adaptation methods reduces the effect of
the domain shift
by learning domain-invariant representations that can generalize
well across different tasks.
In order to learn such representations, 
domain adaptation methods 
either minimize statistical measures~\cite{Tzeng2014,Long2015,Long2016} of domain shift 
or use adversarial training~\cite{Liu2019}.
Following the success of 
adversarial training in Generative Adversarial Networks (GANs)~\cite{Goodfellow2014},
adversarial domain adaptation methods~\cite{Ganin2015, Long2017, Sankaranarayanan2018} gained more attention and demonstrated
state-of-the-art performances
over benchmark CV and NLP datasets.
Adversarial domain adaptation methods use
adversarial training to learn domain-invariant representations
via a mini-max optimization
using a feature extractor, a domain classifier and a label predictor.
%
The principle of adversarial training is used to train the feature extractor
to learn domain-invariant representations which are indistinguishable by the domain classifier.
Seminal methods in adversarial domain adaptation~\cite{Ganin2015, Long2017, Tzeng2017} 
differ in the design choices such as 
adversarial loss functions, optimization, 
coupling of weights etc.
Other existing methods focus on 
conditional feature alignment~\cite{Long2018,Monteiro2021},
multi-source transfer~\cite{Pei2018, Zhao2018}, etc.
However, these methods have been specifically developed
for image domain adaptation or image translation problems. 
To the best of our knowledge, none of these methods have been
widely adapted for graph-structured data.
In this work, following the idea of adversarial domain adaptation,
we proposed a novel transfer learning method that learns effective compound representations
from graph-structured data and transfers relevant information
from a related task to the target task.
\subsubsection{Transfer Learning in SAR predictions}
\label{sec:related:transfer_sar}
To alleviate the limited data problem in cheminformatics,
various transfer learning~\cite{Altae-Tran2017,Simoes2018,Lee2019,Cai2020,Guo2021} and 
multitask learning methods~\cite{Dahl2014,Ramsundar2017,Xu2017,Rodriguez-Perez2019,Sosnin2019}
have been recently developed.
Inspired by the success of pretraining followed by finetuning in CV and NLP,
Goh \latin{et al.}~\cite{goh2018using} proposed ChemNet, where a deep neural network
is pretrained on a large set of compounds in a self-supervised manner
and then fine-tuned on individual activity/prediction tasks.
Following the same idea, Li and Fourches~\cite{li2020inductive}
proposed MolPMoFit which trains a Long Short Term Memory (LSTM)~\cite{Hochreiter1997} on
SMILES strings of compounds 
and then fine-tunes the pretrained model on specific tasks.
%
%
Although pretraining has been widely studied, 
existing work in cheminformatics do not demonstrate significant
performance improvement over the state-of-the-art 
supervised models in a single-task setting.
Moreover, models trained on SMILES strings
do not explicitly leverage the topological information of compounds.
However, our methods use molecular graphs as inputs, and hence explicitly
leverage the topological information.

Adversarial transfer learning has been rarely explored for 
SAR predictions and on graph-structured data.
To the best of our knowledge, only recently
Abbasi \latin{et al.}~\cite{abbasi2019deep} combined multitask networks and
adversarial domain adaptation
to learn transferable molecular representations from multiple source bioassays to improve the prediction performance on the target bioassay.
The authors evaluated their model on biophysics and physiology
data sets such as Tox21, SIDER, BACE, ToxCast and HIV.
Experimental results demonstrated that the proposed method outperforms 
no-transfer methods only on a few target tasks.
%
Moreover, experimental results do not clearly demonstrate the
contribution of the adversarial domain adaptation component
to the overall performance.
Overall, prior work on transfer-learning based SAR modeling
do not clearly suggest a performance gain over conventional SAR models
over a wide-array of target tasks.
%

\section{Results and Discussion}
\label{sec:results}
In this section,  we present the materials used for experimental evaluation (Section~\ref{sec:materials}),
followed by detailed experimental results and discussions (Section~\ref{sec:results:overall}-\ref{sec:results:ranking}).

\subsection{Materials}
\label{sec:materials}
In this section,  we describe the dataset generation,  baseline methods
and experimental protocols in detail.
\subsubsection{Dataset Generation}
\label{sec:materials:dataset}

We used the real screening data from PubChem to test our methods.
PubChem~\cite{kim2019pubchem, kim2021pubchem} is one of the largest public
chemical database with more than 271M substances, 111M unique
chemical structures and 293M bioassay data.
We selected a set of bioassays from PubChem bioassays [accessed on 2020-12-25] 
such that each bioassay has a sufficiently large number of active and inactive compounds.
Then, we generated pairs of bioassays for transfer learning in accordance 
to the protocols in below sections \ref{sec:materials:dataset:init} and \ref{sec:materials:dataset:pairs}.

\paragraph{Initial Bioassay Selection and Pruning}
\label{sec:materials:dataset:init}
\hfill \\
We first selected a set of 7,284 confirmatory bioassays that have a single protein target,
and are tested on chemical substances. These bioassays have 1,279 unique protein targets in total.
Among these protein targets, we were able to identify the organism and protein family information for 
961 protein targets within 435 protein families using UniProt~\cite{Bateman2021}. 
%
%
Among the 435 protein families, we further combined them into 278 families (e.g., Peptidase A1, Peptidase C12, and Peptidase C13 families
were combined into peptidase family). 
Among the 278 families, we selected 10 that have the most protein targets belonging to `Human' organisms. 
%
These top-10 protein families 
are 1) G-protein coupled receptor 1 family, 2) peptidase family, 
3) protein kinase family, 4) nuclear hormone receptor family, 5) protein-tyrosine phosphatase family,
6) ABC transporter family, 7) Cytochrome P450 family, 8) Bcl-2 family, 9) G-protein coupled receptor 3 family,
and 10) Histone deacetylase family. 
These protein families involved 269 unique protein targets and 
covered the major drug targets in drug discovery~\cite{Santos2016,Zdrazil2020}. 
%

According to the 10 protein families, 
bioassays with targets from these protein families were then processed as follows: 
\begin{enumerate}
\item We combined bioassays of a same target into one bioassay, resulting in 269 combined bioassays. 
\item For each combined bioassay, 
we selected its compounds
that were tested for inhibition against the target (i.e., the corresponding PubChem activity type
specified by the depositor was ``inhibitor").
%
%
\item From those inhibitive compounds, we selected the compounds that were specified as either ``active" or ``inactive" 
against the target, and discarded the compounds that were specified as 
``inconclusive'' or ``undetermined".
%
\item If the active/inactive compounds appeared multiple times in the bioassay with a same 
activity label, we retained one of their records. 
If the active/inactive compounds appeared multiple times in the bioassay with different activity labels, 
we removed the compounds from the bioassays (in our data set, about only 2.08\% of compounds for each bioassay on average
appear multiple times with different activity levels). 
We use canonical SMILES strings to detect identical compounds.  
%
%
%
\end{enumerate}

After the above processing, each combined bioassay has on average 17,005 unique 
compounds in total, with 188
active and 16,817 inactive. 
Furthermore, out of the 269 combined bioassays, 95 bioassays have more than 50 active compounds.
Among the 10 protein families involved in the 95 bioassays,
2 protein families had only 1 target with more than 50 active compounds. 
Thus, we removed these 2 protein families and only used the remaining 8 protein families and their 93 bioassays. 
This set of 93 bioassays has on average 40,115 compounds, with 521 active 
and 39,595 inactive.
This set of bioassays will be used to create bioassay pairs as will be described in the 
next section.
Table S1 in the Supplementary Materials presents the statistics of each of the 93 bioassays. 


\paragraph{Transferable Bioassay Pairing}
\label{sec:materials:dataset:pairs}
%
%
\hfill \\
From the 93 processed bioassays, we constructed 765 bioassay pairs such that in each pair, 
the protein targets of the two bioassays are from the same protein family.
%
We selected targets from a same protein family because based on the key intuition of 
chemical genomics~\cite{Harris2006, Klabunde2007}, proteins from a same family tend to have similar binding pockets and bind to similar compounds
-- this is the physicochemical foundation to enable possible information transfer across protein targets, and such targets
and their bioassays can be used to test transfer learning. 
%
We first ensured that each of the 765 pairs of bioassays had balanced active and inactive compounds as follows: 
\begin{enumerate}
%
\item In each pair of bioassays, we removed the compounds that appeared in both bioassays but with different activity labels
(on average, 2.09\% of all unique compounds in a pair of bioassays). 
This is to avoid any conflicting information across bioassays,
which could adversely affect our transfer learning method.
%
\item For compounds with same activity labels in both the bioassays (on average, 1.82\% of all unique compounds in a pair of bioassays), 
we randomly sampled half of them into one of the bioassays, 
and the other half into the other.
This is to avoid duplication of compounds across bioassays, which could
lead to overestimation of predictive performance.
%
\item After the above steps, for each bioassay of a pair, 
we used all its active compounds, 
and randomly sampled the same number of its inactive compounds. 
%
If the inactive compounds were not sufficient, 
we randomly sampled compounds from PubChem that were not active 
in the bioassay as additional inactive compounds for the bioassays. 
This is to ensure that each bioassay in a pair has equal number of active and inactive compounds, and thus 
the learning will not be dominated by either active or inactive compounds. Please note that a bioassay involved in 
two pairs may have different numbers of active and inactive compounds due to its paired, the other bioassay. 
\end{enumerate} 

After the above steps, we selected the bioassay pairs such that each bioassay in each pair had
at least 50 active compounds retained. There were 635 such pairs and involved 92 bioassays in total.  
Among the 635 pairs of bioassays, we further selected the pairs as follows 
such that the active compounds in each pair are similar to each other compared to its inactive compounds:
%
%
%
%
\begin{enumerate}
\item For a pair of bioassay $\bioassay_i$ and $\bioassay_j$, and their respective active and inactive compounds, 
denoted as $\SetCompounds_{\scriptsize \bioassay_i}^{+}$, $\SetCompounds_{\scriptsize \bioassay_i}^{-}$, 
$\SetCompounds_{\scriptsize \bioassay_j}^{+}$ and $\SetCompounds_{\scriptsize \bioassay_j}^{-}$, respectively, 
we calculated the following two types of average compound similarities using Taminoto coefficient~\cite{Willett1998} over 
Morgan-count  fingerprints (with radius 3 and dimension = 2,048): 
1) among compounds of same labels across the two bioassays: 
$\text{sim}(\SetCompounds_{\scriptsize \bioassay_i}^{+}, \SetCompounds_{\scriptsize \bioassay_j}^{+})$, and
$\text{sim}(\SetCompounds_{\scriptsize \bioassay_i}^{-}, \SetCompounds_{\scriptsize \bioassay_j}^{-})$; and 
2) among compounds of different labels across the two bioassays: 
$\text{sim}(\SetCompounds_{\scriptsize \bioassay_i}^{+}, \SetCompounds_{\scriptsize \bioassay_j}^{-})$, and
$\text{sim}(\SetCompounds_{\scriptsize \bioassay_i}^{-}, \SetCompounds_{\scriptsize \bioassay_j}^{+})$. 
%
\item Based on the similarities, we selected a set of bioassay pairs, denoted as $\mathcal{P}_0$, such that 
in each pair, the active compounds of the two bioassays are more similar, that is, 
\begin{equation*}
\mathcal{P}_0 = \{(\bioassay_i, \bioassay_j) | \text{sim}(\SetCompounds_{\scriptsize \bioassay_i}^{+}, \SetCompounds_{\scriptsize \bioassay_j}^{+}) > 
\text{sim}(\SetCompounds_{\scriptsize \bioassay_i}^{+}, \SetCompounds_{\scriptsize \bioassay_j}^{-}) 
~~\text{and}~~
\text{sim}(\SetCompounds_{\scriptsize \bioassay_i}^{+}, \SetCompounds_{\scriptsize \bioassay_j}^{+}) > 
\text{sim}(\SetCompounds_{\scriptsize \bioassay_i}^{-}, \SetCompounds_{\scriptsize \bioassay_j}^{+})
\}
\end{equation*}
We identified 329 such pairs. 
%
%
From $\mathcal{P}_0$, we further selected a set of bioassay pairs, denoted as $\mathcal{P}$, such 
that in each pair, the active compounds in the two bioassays have a similarity above a certain threshold, that is, 
%
%
\begin{eqnarray*}
& \mathcal{P} = &\{(\bioassay_i, \bioassay_j)  | 
(\bioassay_i, \bioassay_j) \in \mathcal{P}_0, \\
& & \text{sim}(\SetCompounds_{\scriptsize \bioassay_i}^{+}, \SetCompounds_{\scriptsize \bioassay_j}^{+}) - 
\text{sim}(\SetCompounds_{\scriptsize \bioassay_i}^{+}, \SetCompounds_{\scriptsize \bioassay_j}^{-}) 
+
\text{sim}(\SetCompounds_{\scriptsize \bioassay_i}^{+}, \SetCompounds_{\scriptsize \bioassay_j}^{+}) - 
\text{sim}(\SetCompounds_{\scriptsize \bioassay_i}^{-}, \SetCompounds_{\scriptsize \bioassay_j}^{+}) \ge 0.026
\}, 
\end{eqnarray*}
where 0.026 is the average value of $\text{sim}(\SetCompounds_{\scriptsize \bioassay_i}^{+}, \SetCompounds_{\scriptsize \bioassay_j}^{+}) - 
\text{sim}(\SetCompounds_{\scriptsize \bioassay_i}^{+}, \SetCompounds_{\scriptsize \bioassay_j}^{-}) 
+
\text{sim}(\SetCompounds_{\scriptsize \bioassay_i}^{+}, \SetCompounds_{\scriptsize \bioassay_j}^{+}) - 
\text{sim}(\SetCompounds_{\scriptsize \bioassay_i}^{-}, \SetCompounds_{\scriptsize \bioassay_j}^{+})$ among 
all the pairs. 
\end{enumerate}

After the above process, we identified 120 pairs of bioassays in $\mathcal{P}$, involving 59 bioassays and 
7 protein families, with 
278 active and 278 inactive compounds in each bioassay on average.
Table S2 presents all the pairs and their compound statistics.
\subsubsection{Baseline Methods}
\label{sec:materials:baselines}
We tested our \OurMethod and \OurMethodWithDisc methods with respect to two aspects: 
1) compound representations, and 2) transfer mechanisms. Compound representation is 
key to revealing information among compounds that can be leveraged to transfer across. 
Transfer machenisms are critical to enable effective transfer of revealed information across 
bioassays. 
%

\paragraph{Compound Representation Methods}
\hfill \\
Specifically, we compared our compound representation method \newgnn
{(i.e., the feature learner in Section{~\ref{sec:methods:conv}})}
 with the following compound representation methods:
%
\begin{itemize}
\item Binary Morgan fingerprint (\MORGAN)~\cite{rogers2010extended}: 
\MORGAN uses a binary feature vector to present a compound, in which each dimension of the 
feature vector corresponds to a pre-defined substructure, and the binary value in that dimension represents if 
the compound has that substructure or not. 
\item Morgan count fingerprints (\MORGANC)~\cite{rogers2010extended}: \MORGANC is very similar to \MORGAN 
except that the values in \MORGANC represent how many corresponding substructures the compound has. 
\item Directed Message Passing Network (\dmpnn)~\cite{Yang2019}: 
The \dmpnn method learns molecular structures by passing
messages along directed edges over molecular graphs.
It produces two representations for each bond
through message passing through the two directions along the bond. 
Then it learns atom representations from the incoming bond
representations and generates a compound representation using
mean pooling over the atom representations.
The \dmpnn\footnote{https://github.com/chemprop/chemprop [accessed  2021-01-22]} 
method is the state-of-the-art compound embedding learning approach for compound 
property prediction. 
%
%

\end{itemize}   

We generated \MORGAN and \MORGANC
(with radius = 3 and size = 2,048) using RDKit~\cite{landrum2020rdkit}.
%
%
In order to only compare the different compound representation methods, not the transfer learning mechanisms, 
we used a 2-layer fully-connected network as the classifier \classifier
over the above baseline feature representations to predict activity labels.
We used cross-entropy as the loss function for these baseline methods. 
%
The corresponding methods are denoted as \morganFCN, \morgancFCN and \dmpnnFCN, 
respectively. 
Note that  these three baseline methods do not have information transfer mechanisms -- they are 
single-task compound prediction methods.

\paragraph{Learning Methods for Compound Prediction:}
\hfill \\
We compared \OurMethod and \OurMethodWithDisc with a transfer learning baseline known as
domain-adversarial neural network, denoted as \DANN~\cite{Ganin2015}.
We selected \DANN because, to the best of our knowledge, there are no existing transfer learning 
methods over graph-structured data; and \DANN is a standard transfer learning baseline method
used on other data (e.g., images)~\cite{Wang2018}.
%
In particular, we adapted \DANN to learn compound features 
from graph-structured data via \gnn (e.g., \dmpnn or \newgnn). 

\DANN consists of three components: 
1) a feature extractor that
represents compounds via feature learning;
2) a label predictor that predicts activity labels from learned compound features;
3) a domain classifier that discriminates between the source and 
target compounds during training.
%
%
\DANN learns compound features that can generalize well
from one domain to another
such that the learned features contain little discriminative domain information, and   
enable \DANN to accurately predict activity labels.

The objective function in \DANN consists of two losses:
domain classification loss, and 
label prediction loss. 
\DANN uses a minimax optimization such that the domain classification loss
is minimized with respect to the domain classifier, and is
maximized with respect to the feature extractor.
Specifically, minimizing the domain classification loss will encourage
the domain classifier to correctly discriminate between the
source and target compounds.
On the other hand, maximizing the domain classification loss will
encourage the learning of generalizable compound features.

%
The feature learner and discriminators in \OurMethodWithDisc
are learned via a minimax optimization similarly to how the 
feature extractor and the domain classifier in \DANN are learned.
However, \OurMethodWithDisc is different from \DANN in that
\OurMethodWithDisc learns feature-wise transferability and compound-wise transferability, 
while \DANN only learns compound-wise transferability. 
%
Furthermore, following Ganin \latin{et al.}~\cite{Ganin2015},
\DANN is trained on labeled data from the source domain
and unlabeled data from the target domain.
%

%
%

\subsubsection{Experimental Protocols}
\label{sec:materials:experimental}
%

\paragraph{Experimental Settings}
\label{sec:materials:experimental:setting}
\hfill \\
In our experiments, we split each of the bioassays in a pair into 10 folds. 
For the target bioassay, we used 1 fold for modeling training, 1 fold for validation and 8 folds for testing. 
We did the above process 10 times, with a different 
training fold each time, and reported the average performance over the 10 times. 
The above 1:1:8 training-validation-testing ratio follows a typical setting in transfer learning~\cite{Wang2021}, 
where it is assumed that the training data is limited so it is needed to  
leverage other tasks via transfer. 
%
%
%
We used this cross validation setting because we did not have a benchmark test set for each bioassay; and 
10-fold across validation will reduce variance of the model performance. 
When we transferred the information from the source bioassay to the target bioassay, 
we used all the folds of the source bioassay and the training fold of the target bioassay in \OurMethod in order 
to maximize the information content in the source bioassay that could be leveraged. 
 
%
%

If the baseline methods do not have an information transfer mechanism (e.g., \morganFCN), we applied an additional setting 
to simulate information transfer: 
in addition to the target task \TargetTask's own training compounds, we also used all the compounds from the source task \SourceTask as 
training data of \TargetTask. 
Thus, \SourceTask's compounds will enrich \TargetTask's training data and bring (i.e., transfer) information from \source 
directly to \target. This setting is referred to as data transfer, denoted as \datatransfer. 
If we only use \TargetTask's compounds for training as in conventional single-task models, this setting is denoted as \notransfer.   


%
%
We trained each model using ADAM~\cite{Kingma2015} optimizer 
with an initial learning rate 1e-3. 
All the models are trained up to 40 epochs.
We used grid-search to tune all the hyper-parameters such as the 
dimension $d$ of the compound embedding \compr, 
hidden-layer dimension of the attention layer for \newgnn, 
hidden-layer dimension in \ldisc and \gdisc,
and batch size. 
We used the validation set to determine the optimal number of epochs. 
During training, we evaluated the performance of each model on the validation set 
at every epoch and choose the trained model at some epoch $k$
that gives the best performance on the validation set; 
thus we selected $k$ as the optimal number of epochs. 
We used \rocauc metric for the above performance evaluation. 
All evaluation metrics are discussed in the following section.
All the hyper-parameters are reported in Table S3 for reproducibility purposes.

\paragraph{Evaluation Metrics}
\label{sec:materials:metrics}
\hfill \\
We used the following evaluation metrics: 
area under the precision-recall curve (\prauc), area under the receiver operating characteristic curve (\rocauc), 
\precision,  \sensitivity, 
\accuracy and \fone score.

\begin{itemize}
\item Area under the Precision-Recall curve (\prauc): 
A Precision-Recall curve is generated by (precision, recall) value pairs
corresponding to variable thresholds. 
\prauc measures the area under the Precision-Recall curve, and 
provides an aggregate measure of performance across all possible thresholds.
\item Area under the Receiver Operating Characteristics curve (\rocauc):
A Receiver Operating Characteristic (ROC) curve is generated by  
true positive rates against false positive rates at
various threshold values.
\rocauc measures the area under the ROC curve.
\item \precision: it is the ratio of correctly predicted
positive instances out of all predicted positive instances 
(e.g. the ratio of predicted active compounds that are truly active). 
\item  \sensitivity: it is the ratio of correctly predicted positive
instances out of all ground-truth positive instances
(e.g. the ratio of active compounds that are correctly predicted as active).
%
%
\item \accuracy: it is the ratio of correctly predicted 
(positive and negative) instances out of all instances
(e.g. the ratio of compounds that are correctly predicted as active/inactive).
\item \fone-score: it is the harmonic mean of \precision and \sensitivity. 
\end{itemize}
If the above metrics have higher values, they indicate better performance. 

\subsubsection{Data and Software Availability}
All the data sets and source code are publicly available at \url{https://github.com/ninglab/TransferAct}.


\subsection{Overall Performance}
\label{sec:results:overall}
%


\begin{table}[h]
\centering
\caption{Overall Comparison}
\label{tbl:overall}
\begin{small}
  \begin{threeparttable}
      \begin{tabular}{
          @{\hspace{1pt}}l@{\hspace{1pt}}
          @{\hspace{1pt}}c@{\hspace{1pt}}
          @{\hspace{1pt}}c@{\hspace{1pt}}
          @{\hspace{1pt}}c@{\hspace{1pt}}
          @{\hspace{1pt}}c@{\hspace{1pt}}
          @{\hspace{1pt}}c@{\hspace{1pt}}
          @{\hspace{1pt}}c@{\hspace{1pt}}
          }
          \toprule
          method & \rocauc & \prauc & \precision 
          & \sensitivity 
          & \accuracy & \fone \\
          \midrule
          \morganFCN & 0.727$\pm$0.124 
          & 0.729$\pm$0.121 
          & 0.648$\pm$0.104 
          & \ul{0.742$\pm$0.131} 
          & 0.661$\pm$0.110 
          & \ul{0.683$\pm$0.105}
          \\
          \morgancFCN & 0.731$\pm$0.120 
          & 0.730$\pm$0.118 
          & 0.653$\pm$0.102 
          & 0.735$\pm$0.132 
          & 0.664$\pm$0.107 
          & 0.682$\pm$0.105
          \\
          \dmpnnFCN & 0.754$\pm$0.101 
          & 0.733$\pm$0.102 
          & 0.619$\pm$0.116 
          & 0.739$\pm$0.156 
          & 0.656$\pm$0.087 
          & 0.655$\pm$0.126
          \\
          \newgnnFCN & 0.755$\pm$0.112 
          & 0.729$\pm$0.112 
          & 0.660$\pm$0.119 
          & 0.712$\pm$0.165 
          & 0.665$\pm$0.101 
          & 0.651$\pm$0.136
          \\
          \dmpnnFCNDT & 0.754$\pm$0.104 
          & 0.735$\pm$0.105 
          & 0.687$\pm$0.106 
          & 0.686$\pm$0.213 
          & 0.669$\pm$0.088 
          & 0.655$\pm$0.140
          \\
          \newgnnFCNDT & \ul{0.763$\pm$0.108} 
          & \ul{0.745$\pm$0.109} 
          & \ul{0.702$\pm$0.108} 
          & 0.671$\pm$0.213 
          & \ul{0.672$\pm$0.092} 
          & 0.645$\pm$0.148
          \\
          \midrule
          \DANNWithDMPNN & 0.733$\pm$0.103
          & 0.715$\pm$0.103
          & 0.671$\pm$0.110
          & 0.647$\pm$0.215
          & 0.649$\pm$0.084
          & 0.623$\pm$0.144
          \\
          \DANNWithDMPNNA & 0.734$\pm$0.102
		 & 0.716$\pm$0.104
		 & 0.676$\pm$0.106
		 & 0.653$\pm$0.226
		 & 0.651$\pm$0.085
		 & 0.624$\pm$0.154
          \\
          \midrule
          \OurMethodWithDMPNN
          & 0.798$\pm$0.103 
          & 0.785$\pm$0.108 
          & 0.729$\pm$0.095 
          & 0.729$\pm$0.146 
          & \textbf{0.721$\pm$0.093}
          & 0.714$\pm$0.108
          \\
          \OurMethodWithDMPNNWithDisc & 0.798$\pm$0.102 
          & 0.784$\pm$0.107 
          & 0.729$\pm$0.094 
          & \textbf{0.731$\pm$0.142} 
          & 0.720$\pm$0.091 
          & \textbf{0.715$\pm$0.102}
          \\
           \OurMethodWithDMPNNA 
           & \textbf{0.801$\pm$0.102} 
           & \textbf{0.786$\pm$0.107} 
           & \textbf{0.731$\pm$0.094} 
           & 0.729$\pm$0.143
           & 0.720$\pm$0.090
           & 0.713$\pm$0.103
          \\
          \OurMethodWithDMPNNAWithDisc & 0.798$\pm$0.105 
          & 0.785$\pm$0.109 
          & 0.730$\pm$0.097 
          & 0.728$\pm$0.147 
          & 0.719$\pm$0.095 
          & 0.713$\pm$0.109
          \\
	\bottomrule
	\end{tabular}

	\begin{tablenotes}
        \setlength\labelsep{0pt}
		\begin{footnotesize}
		\item In this table, the columns \rocauc, \prauc, \precision, \sensitivity, 
		\accuracy and \fone-score 
		have the average and standard deviation 
		over all bioassays in each performance metric.
		The best performance values are \textbf{boldfaced}. The second best performance values are \ul{underlined}.
		\par
		\end{footnotesize}
	\end{tablenotes}

\end{threeparttable}
\end{small}
\end{table}

Table~\ref{tbl:overall} presents overall performance comparison
between \OurMethodWithDMPNN, \OurMethodWithDMPNNWithDisc,
\OurMethodWithDMPNNA, \OurMethodWithDMPNNAWithDisc,
and the baselines.
%
The columns has the average and standard deviation over all bioassays 
in respective evaluation metrics achieved by the optimal models.
Note that for each bioassay, the optimal model of each method
is the model that gives the best \rocauc value, and thus the performance of each method
in other metrics does not necessarily correspond to the optimal performance
in those metrics.


Table~\ref{tbl:overall} shows that, overall, \OurMethodWithDMPNNA achieves the best performance compared to all other methods.
Specifically, \OurMethodWithDMPNNA 
achieves the best average \rocauc, \prauc and \precision scores
of 0.801, 0.786 and 0.731, respectively.
This demonstrates that \OurMethodWithDMPNNA can learn effective compound features
for the target task
by leveraging source bioassay data, and correctly predict the compounds of the target bioassay.
Furthermore, all variants of \OurMethod and \OurMethodWithDisc,
especially \OurMethodWithDMPNN, \OurMethodWithDMPNNWithDisc,
and \OurMethodWithDMPNNAWithDisc
achieve similar performance on average across all metrics.
The performance of these three methods are not significantly different in most metrics. 
%
%
This suggests that learning feature-wise and compound-wise
transferability via \OurMethodWithDisc methods does not necessarily provide
performance boost on average.
However, compared to the best method \OurMethodWithDMPNNA,
\OurMethodWithDMPNNWithDisc and  \OurMethodWithDMPNNAWithDisc
improve \rocauc scores of 62\% and 39\% target tasks, respectively. 
%
%
On the whole, all variants of \OurMethod and \OurMethodWithDisc 
significantly outperform all baselines.
Specifically, \OurMethodWithDMPNNA improves the average \rocauc
by 4.9-10.1\%, and 
significantly improves \rocauc of at least 83\% of target tasks 
compared to any baseline method.
Each of the other variants such as \OurMethodWithDMPNN, \OurMethodWithDMPNNWithDisc
and \OurMethodWithDMPNNAWithDisc improves \rocauc
of at least 79\% of the target tasks
compared to any baseline method.
This indicates that these methods can effectively transfer relevant information
from the source task to the target task.
In particular, the transfer learning mechanism
in all variants of \OurMethod and \OurMethodWithDisc
can better leverage source domain data than the transfer learning mechanism
in other baselines. 
This is because both \OurMethod and \OurMethodWithDisc variants 
can better control the transferable information by incorporating 
varying degrees of task-relatedness between the source and target tasks during training.
Additionally, \OurMethodWithDisc variants can better extract relevant information from
source domain data by learning feature-wise (Section~\ref{sec:methods:variations:local_disc}) 
and compound-wise transferability (Section~\ref{sec:methods:variations:global_disc}).

%



\begin{table}[h]
\centering
\caption{Performance Comparison of \OurMethodWithDMPNNA vs \newgnnFCN}
\label{tbl:perf_imp_best}

\begin{threeparttable}
	\begin{tabular}{
		@{\hspace{2pt}}l@{\hspace{2pt}}
		@{\hspace{2pt}}c@{\hspace{2pt}}
		@{\hspace{2pt}}c@{\hspace{2pt}}
		@{\hspace{2pt}}c@{\hspace{2pt}}
		@{\hspace{2pt}}c@{\hspace{2pt}}
		@{\hspace{2pt}}c@{\hspace{2pt}}
       	@{\hspace{2pt}}c@{\hspace{2pt}}
      	}
        \toprule
        	method & \rocauc & \prauc & \precision 
      	& \sensitivity 
      	& \accuracy & \fone \\
     	\midrule
         \OurMethodWithDMPNNA 
       	& 0.801  & 0.786   & 0.731
       	& 0.729  & 0.720   & 0.713
      	\\
      	\newgnnFCNDT 
      	& 0.763  & 0.745  & 0.702  
        	& 0.671  & 0.672  & 0.645
 		\\
       	\diff & 4.980 & 5.503 & 4.131 & 8.644 & 7.143 & 10.543
      	\\
      	\multirow{2}{*}{\taskDiff} & 5.702 & 6.085 & 4.876 
      	& 25.281  & 7.727 & 18.464
       	\\
        	 & (2.80e-19) & (8.00e-21) & (1.69e-11) & (1.73e-09) & (1.19e-29) & (8.93e-20)
        	\\
        	\numImpv  & 199 (83\%) & 192 (80\%) & 157 (65\%) & 153 (64\%) & 201 (84\%) & 198 (82\%)  
        	\\
        	\multirow{2}{*}{\taskImpv} &  7.102 &  8.044 &  9.293 &  44.261 &  9.509 &  23.532 
        	\\
 		 &  (5.56e-22) &  (5.51e-26) & (2.81e-27) & (7.36e-25) & (5.02e-35) & (3.60e-26)
 		\\
 		\bottomrule
	\end{tabular}
 	
	\begin{tablenotes}
        \setlength\labelsep{0pt}
		\begin{footnotesize}
		\item In this table, the first two rows
		has the	performance from respective methods averaged
		over all bioassays in each performance metric.
		%
		The row \diff has the percentage difference of average performance in 
		each metric from \OurMethodWithDMPNNA
		over \newgnnFCNDT.
		The row \taskDiff has the average of task-wise percentage
		improvement from \OurMethodWithDMPNNA over
		\newgnnFCNDT in respective metrics, with corresponding p-value in parentheses below.
		The row \numImpv has the number and percentage of target tasks where \OurMethodWithDMPNNA
		performs better than \newgnnFCNDT in respective metrics.
		The row \taskImpv has the average of task-wise percentage
		improvement only among the corresponding improved tasks,
		with corresponding pvalues in parentheses below.
		\par
		\end{footnotesize}
	\end{tablenotes}

\end{threeparttable}
\end{table}

The best performance among the baseline methods is achieved by \newgnnFCNDT.
%
Table~\ref{tbl:perf_imp_best} presents the performance comparison
between {\OurMethodWithDMPNNA} and \mbox{{\newgnnFCNDT}}. 
%
The \diff values in Table~\ref{tbl:perf_imp_best} are calculated as
the percentage difference of average performance in each metric
from \OurMethodWithDMPNNA over \newgnnFCNDT,
where the average performance in each metric is calculated as the performance 
in that metric averaged over all the bioassays.
%
The \taskDiff values
are calculated as the average of task-wise performance improvement (in \%)
from \OurMethodWithDMPNNA over \newgnnFCNDT.
%
The \numImpv values denote the number 
of improved target tasks 
where \OurMethodWithDMPNNA
performs better than \newgnnFCNDT in respective metrics.
Considering only these \numImpv improved tasks, 
the average of task-wise performance improvement (in \%) are listed as \taskImpv values.
Similar to \taskDiff, the numbers presented in parentheses in this row 
are the corresponding p-values for \taskImpv.
A \pvalue less than 0.05 was considered to be statistically significant. 

Clearly, compared to the best baseline method \newgnnFCNDT,
\OurMethodWithDMPNNA improves average
\rocauc, \prauc, \precision, \sensitivity, \accuracy and \fone scores by
4.980\%, 5.503\%, 4.131\%, 8.644\%, 7.143\%, and 10.543\%,
respectively.
Furthermore, the average task-wise performance difference (i.e., \taskDiff)
from \OurMethodWithDMPNNA over \newgnnFCNDT across each metric is
5.702\%, 6.085\%, 4.876\%, 25.281\%, 7.727\%, and 18.464\%,
respectively; and these differences are positive and statistically significant
(as indicated by their corresponding p-values in parentheses)
-- hence suggesting that \OurMethodWithDMPNNA
significantly improves the task-wise performance over \newgnnFCNDT.
%
In particular, \OurMethodWithDMPNNA significantly improves the \rocauc
performance of 199 out of 240 (83\%) target tasks 
with an average task-wise improvement (i.e. \taskImpv) of 7.102\%
(\pvalue: 5.56e-22).
Such consistent and significant improvement
(demonstrated by \taskImpv and their corresponding p-values) 
across all evaluation metrics
on a large percentage of target tasks (demonstrated by \numImpv)
provides strong evidence that \OurMethodWithDMPNNA
clearly outperforms \newgnnFCNDT on majority of target tasks.
This further implies that the transfer mechanism in \OurMethodWithDMPNNA
is more effective than that in \newgnnFCNDT.
While \newgnnFCNDT pays equal attention to both the source and target tasks
during training,
\OurMethodWithDMPNNA can differentially focus on the two tasks
by varying the weightage on the source classification loss 
(i.e., the trade-off parameter $\alpha$ in equation~\ref{eqn:classifier}).
Note that \OurMethodWithDMPNNA with $\alpha=1$ is 
methodologically equivalent to \newgnnFCNDT.
By varying $\alpha$, \OurMethodWithDMPNNA can incorporate
different degrees of task relatedness between the source and target tasks during training.
If the two tasks are not that related, a lower $\alpha$ will encourage the 
learning to focus more on the target task. 
In essence, learned compound features are more specific to the target task.
On the other hand, $\alpha$ as high as 1 will enforce learning of compound features
that generalize well across the two tasks. 
Such features may encode little target task-specific information, and hence,
are not effective.
%


Furthermore, our experimental results in Table~\ref{tbl:overall} demonstrate the efficacy of our proposed attention
mechanism of \newgnn in learning better compound features.
Overall, both \newgnn-based methods (i.e., \newgnnFCN and \newgnnFCNDT)
outperform \dmpnn-based methods
(i.e., \dmpnnFCN and \dmpnnFCNDT).
Particularly, compared to \dmpnnFCNDT, 
\newgnnFCNDT improves about half of the 
target tasks 
improves \rocauc of 152 out of 240 (63\%) target tasks
and gives significant
performance improvement of 3.443\% (\pvalue: 2.64e-18) on those improved target tasks.
This demonstrates that the proposed attention mechanism in \newgnn 
enables more effective compound features 
since it can differentially score atoms based on their relevance towards the final task.
However, \newgnnFCN either achieves similar or slightly worse performance
compared to \dmpnnFCN.
This is because \newgnn -- with slightly more parameters than \dmpnn --
may struggle to capture relevant patterns during training, 
and thus \newgnnFCN can easily overfit to limited training data of the target task.
In essence, this can lead to poor generalization performance on the test data.
On the other hand, \newgnnFCNDT can generalize well since it 
is trained on the labeled source data along with the limited target data.
{Overall, the attention mechanism can better learn and effectively score the atoms
in {\newgnnFCNDT} but not in {\newgnnFCN} -- thereby 
achieving significant improvement in the former over \dmpnnFCNDT, and marginal improvement in the latter over \dmpnnFCN.
}
We will further demonstrate the efficacy of our proposed \newgnn
in the compound prioritization problem detailed in Section~\ref{sec:results:ranking}.
%


Furthermore, all \gnn-based baselines 
(i.e., \dmpnnFCN, \newgnnFCN , \dmpnnFCNDT
and 
\\
\newgnnFCNDT )
significantly outperform \DANN-based methods.
Our experimental results show that both \DANN-based methods 
yield poor or similar performance
compared to all other baseline methods.
%
Specifically, the best \DANN method (i.e. \DANNWithDMPNNA)
reduces the average performance by 3-4\% 
over the best baseline method \newgnnFCNDT
across all evaluation metrics.
%
Such poor performance 
may be due to the ineffectiveness of domain-invariant compound features
to encode necessary task-specific information.
%

Surprisingly, \DANN even performs worse than the fingerprint-based methods 
(i.e. \morganFCN and \morgancFCN).
%
As a matter of fact, overall, fingerprint-based methods 
performs relatively well compared to all other baselines.
Compared to \gnn-based methods, 
fingerprint-based methods achieve competitive or even better performance in most evaluation metrics.
This could be due to potential overfitting of \gnn-based methods
in low-data settings.
It is known that \gnns require large amounts of training data 
to extract relevant molecular substructures
and to effectively encode meaningful task-specific information.
In contrast, fingerprint-based methods are not data-hungry owing to fewer learnable parameters
and thus, these methods can perform reasonably well in low-data setting~\cite{Wu2018}.

\subsection{Top-N Task-wise Performance Comparison}
\label{sec:results:topn_comp}



\begin{table}[h]
\centering
\caption{Top-$N$ Performance Comparison (\%)}
\label{tbl:topn_task}
\begin{small}
  \begin{threeparttable}
      \begin{tabular}{
          @{\hspace{10pt}}l@{\hspace{10pt}}
          @{\hspace{10pt}}r@{\hspace{5pt}}
          @{\hspace{5pt}}r@{\hspace{5pt}}
          @{\hspace{5pt}}r@{\hspace{5pt}}
          @{\hspace{10pt}}r@{\hspace{5pt}}
          @{\hspace{5pt}}r@{\hspace{5pt}}
          @{\hspace{5pt}}r@{\hspace{5pt}}
          @{\hspace{10pt}}r@{\hspace{5pt}}
          @{\hspace{5pt}}r@{\hspace{5pt}}
          @{\hspace{5pt}}r@{\hspace{5pt}}
          }
          \toprule
          {method} 
          & \multicolumn{3}{c}{\rocauc} 
          & \multicolumn{3}{c}{\prauc} 
          & \multicolumn{3}{c}{\fone} 
          \\
          \cmidrule(lr){1-1}
          \cmidrule(lr){2-4}
          \cmidrule(lr){5-7}
          \cmidrule(lr){8-10}
          top-$N$ & 1 & 3 & 5
          & 1 & 3 & 5
          & 1 & 3 & 5
          \\
          \midrule
		 \morganFCN & 10 &  15 &  20 
		 & 13 &  21 &  30 
		 & 11 &  19 &  23 
		 \\
		 \morgancFCN & 2 &  13 &  18 
		 & 3 &  17 &  22 
		 & 4 &  17 &  22
		 \\
		 \dmpnnFCN & 5 &  9 &  17 
		 & 6 &  11 &  17 
		 & 10 &  19 &  30
		 \\
		 \newgnnFCN & 4 &  10 &  26 
		 & 2 &  7 &  20 
		 & 5 &  11 &  26
		 \\
		 \dmpnnFCNDT & 2 &  11 &  23 
		 & 4 &  13 &  24 
		 & 5 &  13 &  22
		 \\
		 \newgnnFCNDT & 5 &  18 &  33 
		 & 6 &  14 &  31 
		 & 2 &  8 &  23
		 \\
		 \midrule
		 \DANNWithDMPNN & 2 &  9 &  16 
		 & 1 &  5 &  15 
		 & 3 &  10 &  19 
		 \\
		 \DANNWithDMPNNA 
		 & 2 &  8 &  17 
		 & 1 &  8 &  15 
		 & 2 &  11 &  19
		 \\
		 \midrule
		\OurMethodWithDMPNN 
		& 18 &  52 &  85 
		& 17 &  48 &  81 
		& 12 &  49 &  \textbf{81}
		\\
		\OurMethodWithDMPNNWithDisc 
		& 12 &  45 &  79 
		& 15 &  48 &  80 
		& \textbf{17} &  \textbf{50} &  80
		\\
		\OurMethodWithDMPNNA 
		&  \textbf{22} &  \textbf{62} &  \textbf{89} 
		& 14 &  55 &  83 
		& 13 &  43 &  78
		\\
		\OurMethodWithDMPNNAWithDisc 
		& 16 &  51 &  81 
		& \textbf{20} &  \textbf{56} &  \textbf{86} 
		& 15 &  \textbf{50} &  \textbf{81}
		\\
	\bottomrule
	\end{tabular}

	\begin{tablenotes}
        \setlength\labelsep{0pt}
		\begin{footnotesize}
		\item In this table, the columns \rocauc, \prauc,
		and \fone have the percentage of tasks
		for which each method is ranked within the top-1, top-3 and top-5 best methods 
		in respective metrics.
		The best performance values are \textbf{boldfaced}. 
		\par
		\end{footnotesize}
	\end{tablenotes}

\end{threeparttable}
\end{small}
\end{table}


Table~\ref{tbl:topn_task} presents a fine-grained performance comparison
of top-performing methods over all 240 target tasks
across different evaluation metrics.
%
%
The columns corresponding to each evaluation metric has the 
percentage of tasks for which 
each method is among the top-$k$ ($k=1,3,5$) best methods with respect to the metric. 
%
%
%
Note that for each method, we consider the best performing model 
that achieves the optimal performance in each evaluation metric.
Therefore, for a given method, 
the models with the optimal performance in each metric
do not necessarily have the same set of corresponding hyperparameters.
%

%
Table~\ref{tbl:topn_task} shows that \OurMethod methods achieve the 
top-1 best performance among more tasks compared to other methods. For example, \OurMethodWithDMPNNA  is the best 
performing method in terms of \rocauc for 22\% of tasks, that is, more than 2 folds compared to the 
best baseline method \morganFCN (10\%). 
\OurMethodWithDMPNNA consistently achieves the top-3 and top-5 best performance in terms of \rocauc on significantly more 
tasks compare to other methods, with even more folds of difference.  
Similar trends hold for \prauc and \fone as the evaluation metrics. This indicates the strong performance of \OurMethod methods. 

Among the four \OurMethod variants, \OurMethodWithDMPNNA is the best in terms of \rocauc;
\OurMethodWithDMPNNAWithDisc is overall the best in terms of \prauc as it achieves top-1, top-3 and top-5 best performance 
on more tasks compared to other methods; 
and \OurMethodWithDMPNNWithDisc and \OurMethodWithDMPNNAWithDisc are the best in terms of \fone as they are either better than 
or similar to other methods. 
This indicates that, while different variants may have advantages of optimizing with respect to different evaluation metrics, 
\OurMethodWithDisc (Figure~\ref{fig:aada_L+G}, with the feature-wise and compound-wise discriminators) 
is actually also a very strong method or even better compared to \OurMethod. 

\subsection{Comparison on Discriminators}
\label{sec:results:ablation}


\begin{table}[!h]
\centering
\caption{Comparison on Discriminators (with \newgnn)}
\label{tbl:ablation}
\begin{small}
  \begin{threeparttable}
      \begin{tabular}{
          @{\hspace{5pt}}l@{\hspace{5pt}}
          @{\hspace{5pt}}r@{\hspace{5pt}}
          @{\hspace{5pt}}r@{\hspace{5pt}}
          @{\hspace{5pt}}r@{\hspace{5pt}}
          @{\hspace{5pt}}r@{\hspace{5pt}}
          @{\hspace{5pt}}r@{\hspace{5pt}}
          @{\hspace{5pt}}r@{\hspace{5pt}}
          }
          \toprule
          method &  \rocauc & \prauc & \precision 
          & \sensitivity 
          & \accuracy & \fone \\
          \midrule
          \OurMethod & 0.801 & 0.786 & 0.731 & 0.729 & 0.720 & 0.713 \\
		\midrule
		 \OurMethodWithDisc & 0.798    & 0.785    & 0.730    & 0.728    & 0.719    & 0.713    \\
		 \cmidrule(lr){2-7}
		 \multicolumn{1}{r}{\diff}  & -0.375   & -0.127   & -0.137   & -0.137   & -0.139   & 0.000    \\
		 \multicolumn{1}{r}{\taskDiff}  & -0.380   & -0.119   & -0.072   & 0.116  & -0.150   & -0.064   \\
		 & (2.59e-04) & (4.88e-01) & (7.26e-01) & (7.00e-01)    & (5.53e-01) & (8.04e-01) \\
		 \multicolumn{1}{r}{\numImpv}  & 93 (39\%) & 125 (52\%) & 119 (50\%) & 111 (46\%) & 99 (41\%) & 112 (47\%)      \\
		 \multicolumn{1}{r}{\taskImpv} & 0.921    & 1.373    & 2.756    & 7.037    & 2.230    & 3.729    \\
		 & (7.73e-18) & (4.13e-23) & (1.69e-18) & (5.85e-19)  & (4.26e-14) & (1.31e-15)\\
	   \midrule
	   \OurMethodWithGlobalDisc & 0.801 & 0.786 & 0.730 & 0.734 &  0.721 & 0.716 \\
	   \cmidrule(lr){2-7}
		\multicolumn{1}{r}{\diff} & 0.000 & 0.000 & -0.137 & 0.686 & 0.139 & 0.421 \\
		 \multicolumn{1}{r}{\taskDiff} & 0.010 & -0.080 & -0.130 & 1.583 & 0.128 & 0.516 \\
		& (7.78e-01) & (6.72e-01) & (6.32e-01) & (3.03e-01) & (4.01e-01) & (3.90e-01) \\
		 \multicolumn{1}{r}{\numImpv} & 135 (56\%) & 119 (50\%) & 123 (51\%) & 126 (52\%) & 128 (53\%) & 130 (54\%) \\
		 \multicolumn{1}{r}{\taskImpv} & 0.845 & 1.330 & 2.763 & 8.798 & 1.971 & 4.165 \\
		& (1.13e-23) & (5.03e-24) & (2.67e-17) & (8.22e-18) &  (4.75e-21) & (1.75e-15) \\
		 \midrule
		\OurMethodWithLocalDisc  & 0.799 & 0.785 & 0.732 & 0.722 & 0.721 & 0.711 \\
		\cmidrule(lr){2-7}
		\multicolumn{1}{r}{\diff} & -0.250 & -0.127 & 0.137 & -0.960 & 0.139 & -0.281 \\
		 \multicolumn{1}{r}{\taskDiff} & -0.192 & -0.091 & 0.218 & -0.768  & 0.170 & -0.326 \\
		& (4.00e-02) & (3.76e-01) & (5.44e-01) & (1.42e-01) & (3.19e-01) & (4.98e-01) \\
		 \multicolumn{1}{r}{\numImpv} & 100 (42\%) & 114 (48\%) & 124 (52\%) & 117 (49\%) & 125 (52\%) & 123 (51\%) \\
		 \multicolumn{1}{r}{\taskImpv} & 1.029 & 1.597 & 2.992 & 6.999 & 2.037 & 3.764 \\
		 & (3.22e-13) & (2.22e-21) & (3.85e-24) & (1.11e-19) & (9.65e-20) & (1.12e-16) \\
		 \bottomrule
		 \end{tabular}

	\begin{tablenotes}
        \setlength\labelsep{0pt}
		\begin{footnotesize}
		\item In this table, 
		the first row block has the average performance of \OurMethod.
		Each of the other row blocks has performance comparison of a \OurMethodWithDisc
		variant with respect to \OurMethod.
		%
		%
		The metric \diff represents the difference of average performance of each comparison 
		method with respect to \OurMethod;
		%
		\taskDiff represents the average of the task-wise improvement, with corresponding $p$-values in the 
		parentheses below; 
		%
		\numImpv represents the number of improved tasks and its proportion in the parentheses; 
		%
		and \taskImpv represents the average of the task-wise improvement only among the improved tasks, 
		with corresponding $p$-values in the parentheses below.
		\par
		\end{footnotesize}
	\end{tablenotes}

\end{threeparttable}
\end{small}
\end{table}


Table~\ref{tbl:ablation} presents detailed performance comparison 
between \OurMethod, \OurMethodWithDisc, \OurMethodWithGlobalDisc and \OurMethodWithLocalDisc, all with \newgnn.
We use \newgnn here because as in Table~\ref{tbl:topn_task}, \OurMethodWithDMPNNAWithDisc shows better performance 
on average compared to \OurMethodWithDMPNNWithDisc. 
\OurMethodWithGlobalDisc and \OurMethodWithLocalDisc are obtained by removing either the 
feature-wise discriminator \ldisc (Section~\ref{sec:methods:variations:local_disc}),
or the compound-wise discriminator \gdisc (Section~\ref{sec:methods:variations:global_disc}) 
from \OurMethodWithDisc.
%
%
Note that for each bioassay, the optimal model of each method is selected based on \rocauc. 
%
The \diff values in each row block are calculated as the difference (in \%) of average
performance in each metric from the \OurMethodWithDisc variant over \OurMethod.
The \taskDiff values in each row block are calculated as the average of 
task-wise performance improvements (in \%)
from the corresponding variant over \OurMethod.
The row \numImpv in each row block denotes the number
of improved target tasks where
the variant performs better than \OurMethod in respective metrics, 
and the average of task-wise performance improvement among only the improved tasks is 
calculated as \taskImpv (in \%). 
%
%

Compared to \OurMethod, 
\OurMethodWithDisc achieves similar but slightly worse performance overall (i.e., -0.375\% in \diff on \rocauc); 
on individual tasks, \OurMethodWithDisc has some statistically significant worse performance in terms of \rocauc 
but similar performance as \OurMethod on other evaluation metrics. 
%
%
%
%
In addition, \OurMethodWithDisc still provides significant task-wise 
improvement in about 40--50\% tasks across all evaluation metrics.
Particularly, it improves the \rocauc score for 93 out of 240 (39\%) tasks significantly by 
0.921\% on average (\pvalue = 7.73e-18) over \OurMethod.
This suggests that the learned feature-wise and 
compound-wise transferability together have the capacity of improving some target tasks.
%
%
%
\OurMethodWithGlobalDisc performs similarly to \OurMethod on average (i.e., 0.000 in \diff on \rocauc; no significant \taskDiff).  
%
However, \OurMethodWithGlobalDisc improves over more than half of the tasks with statistical significance on all the evaluation 
metrics. For example, \OurMethodWithGlobalDisc achieves better \rocauc on 135 out of 240 (56\%) tasks. 
This indicates that the global discriminator (Section~\ref{sec:methods:variations:global_disc}) that differentiate compounds for the source and target tasks
could help improve performance for some tasks. 
\OurMethodWithLocalDisc also shows improvement on about half of the tasks (\numImpv) with significant improvement 
that is even higher compared to that in \OurMethodWithGlobalDisc, but with overall performance (\diff) still slightly worse than 
that of \OurMethod. 
The fact that \OurMethodWithDisc, \OurMethodWithGlobalDisc and \OurMethodWithLocalDisc improve about half of the tasks 
over \OurMethod without discriminators indicates that they are suitable for certain tasks. 
%


%

We hypothesize that \mbox{\OurMethodWithGlobalDisc}
can effectively focus on similar compounds of source and target bioassays 
by learning compound-wise transferability via \gdisc.
We validate this hypothesis with an additional analysis on
model predictions and pairwise similarities of predicted compounds 
with source and target compounds.
We find and study the active compounds that are correctly classified as active
by \mbox{\OurMethodWithGlobalDisc} but
incorrectly classified as inactive by \OurMethod and its variants.
%
%
Table S4 presents the analysis for these
active compounds among target tasks which have at least one such active compound. 
%
%
%
For each of such active compounds in the target task, 
we calculated the mean pairwise similarities of 
that compound with its 5 most similar active compounds in the source task and in the target task, respectively. 
%
%
%
%

On average, \mbox{\OurMethodWithGlobalDisc} correctly classifies
5.4\% (i.e., average of values in {`cor\%'} column in Table S4) of active compounds
that are incorrectly classified by \OurMethod and its variants.
These compounds were found to be 12.4\% more similar to the active compounds in the source task
than to those in the target task.
Furthermore, in 47 out of 97 ({48}\%) tasks with at least one active compound only correctly classified by \OurMethodWithGlobalDisc, 
the similarity difference is statistically significant ($\pvalue < 0.05$). 
%
Overall, this analysis demonstrates that \mbox{\OurMethodWithGlobalDisc} 
can better learn the commonalities between source and target compounds,
and hence can enhance information transfer from the source task to the target task.
%

%
%

\subsection{Parameter Study}
\label{sec:results:parameter}


\begin{figure*}
\includegraphics[width=0.5\textwidth]{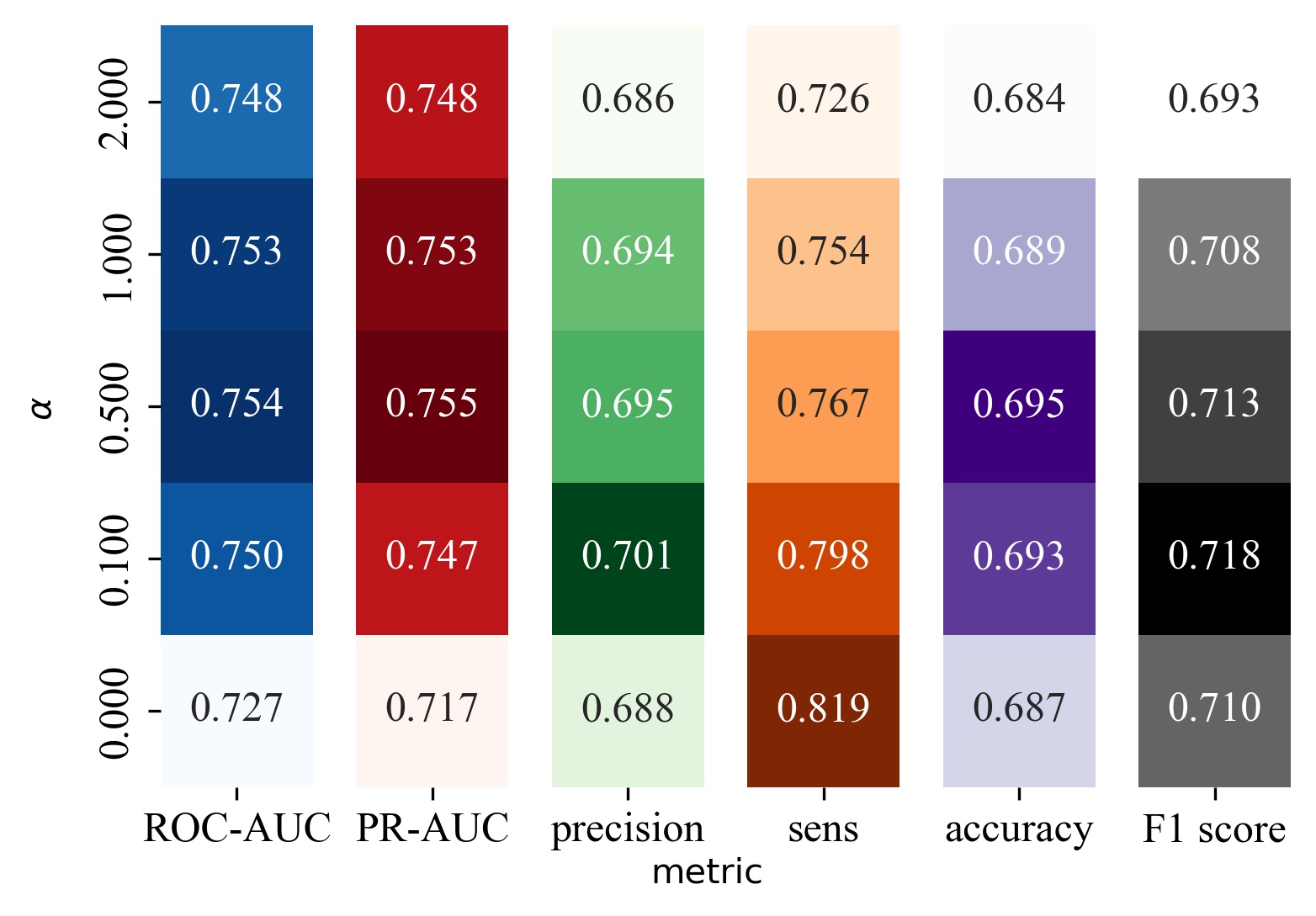}
\caption{Parameter Study of \OurMethodWithDMPNNA.
{The columns represent different evaluation metrics. 
The values in each cell has the average of the best performance achieved with given $\alpha$ and optimal choice of other hyperparameters.
Darker cells indicate better performance.}}
\label{fig:param_best_tasks_aada_dmpnna}
\end{figure*}

%
Figure~\ref{fig:param_best_tasks_aada_dmpnna} presents the parameter study in \OurMethodWithDMPNNA
on $\alpha$ (i.e., the trade-off parameter  
between the source and target classification losses as in Equation~\ref{equ:classifier_loss}). 
The study was conducted over the tasks for which  \OurMethodWithDMPNNA outperforms the other methods. 
%
%
%
The values in each cell in the figure represent the average of the best performance over the tasks with the optimal 
choice of other hyperparameters. 
%
%
%
%

Figure~\ref{fig:param_best_tasks_aada_dmpnna} shows that	
\OurMethodWithDMPNNA has the best average performance
 in \rocauc, \prauc, \precision, \sensitivity, \accuracy 
and \fone-score 
when $\alpha=$0.5, 0.5, 0.1, 0, 0.5 and 0.1, respectively. 
It indicates that weighing the source and target classification losses differently
has notable effects on the overall performance. 
This figure also demonstrates several trends:
1) best average performance is achieved with $\alpha =$ 0.1 and 0.5 (i.e., non-zero values) for all the metrics except \sensitivity;
2) performance degrades especially when $\alpha$ increases.
%
%
Non-zero values of $\alpha$ as the optimal values indicate that leveraging information from the source task is able to help improve
the target task. The fact that the optimal, non-zero $\alpha$ values are relatively small indicates that the training is still more focused
on the target tasks, while useful information is transferred from the source tasks. 
On the other hand, if $\alpha$ is too large (i.e., the source classification loss is given high weightage),
the training would be dominated by the source task, and thus the trained model could not well capture 
the patterns in the target task. That could explain why model performance decreases when $\alpha$ increases. 
%
%
%



\begin{figure}[h!]
    \centering
        \includegraphics[width=0.3\textwidth]{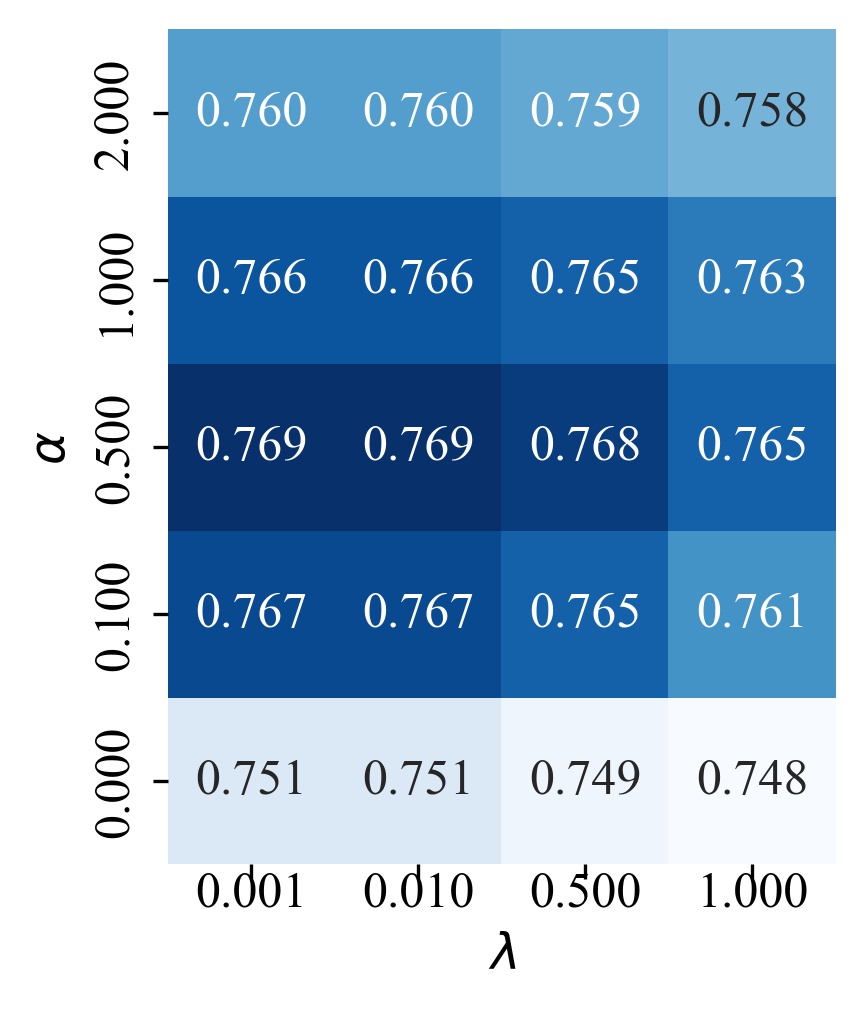}
      \caption{Parameter Study of \OurMethodWithDMPNNAWithDisc in terms of \rocauc}
    \label{fig:param_best_tasks_aada_dmpnna_LG}
\end{figure}


Figure~\ref{fig:param_best_tasks_aada_dmpnna_LG} presents the parameter study in \OurMethodWithDMPNNAWithDisc in terms of 
\rocauc 
on $\alpha$ (i.e., the trade-off parameter between source and target losses
in Equation~\ref{equ:classifier_loss})
and $\lambda$ (i.e., the trade-off parameter between the classification and discriminator losses in Equation~\ref{equ:loss}). 
%
Studies over other metrics are presented in Figure S1 in the Supplementary Materials.
The values in each cell of this figure represent the average of the best performance 
over the tasks where \OurMethodWithDMPNNAWithDisc outperforms all other methods, with 
corresponding $\alpha$ and $\lambda$,
and with optimal choice of other hyperparameters.
%

Figure~\ref{fig:param_best_tasks_aada_dmpnna_LG} shows that	
\OurMethodWithDMPNNAWithDisc has the best performance
in \rocauc (i.e., 0.769)
when $\alpha=0.5$ and $\lambda=0.01$ and $0.001$, that is, all non-zero values.
%
%
This demonstrates that a lower weight on the source classification loss
than the target classification loss,
and a lower weight on discriminator losses (sum of \LocalDiscLoss and \GlobalDiscLoss) 
will enable effective transfer of relevant information from the source domain.
%
Figure~\ref{fig:param_best_tasks_aada_dmpnna_LG}
also demonstrates that when $\alpha$ is too small or too large, 
regardless of what $\lambda$ is, there is a significant 
performance drop (as indicated in topmost rows).
This effect of $\alpha$
can be explained following the same reasoning presented in the previous section.
%
For the optimal $\alpha$ in each metric,
a $\lambda=0.01$ gives the best performance for most metrics.
This implies that \OurMethodWithDMPNNAWithDisc can effectively
leverage source task data to learn transferable compound features (using \ldisc) 
and to selectively focus on similar compounds (using \gdisc) during training.
Intuitively, for a given $\alpha$, higher $\lambda$ values
(i.e. higher weight on discriminator losses)
will encourage learning of more domain-invariant compound features. 
Such domain-invariant features contain little task-specific information,
and may not be relevant for
effective activity classification for the target task, and therefore the overall performance
degrades. 
\subsection{Case Studies: \OurMethodWithDMPNNA}
\label{sec:results:cases}

\subsubsection{Relation between Performance Improvement and Bioassay Similarity}
\label{sec:results:cases:impv_vs_sim}
%
\begin{figure*}[h!]
    \centering
    \begin{subfigure}{0.26\textwidth}
        \centering
        \includegraphics[width=.9\textwidth]{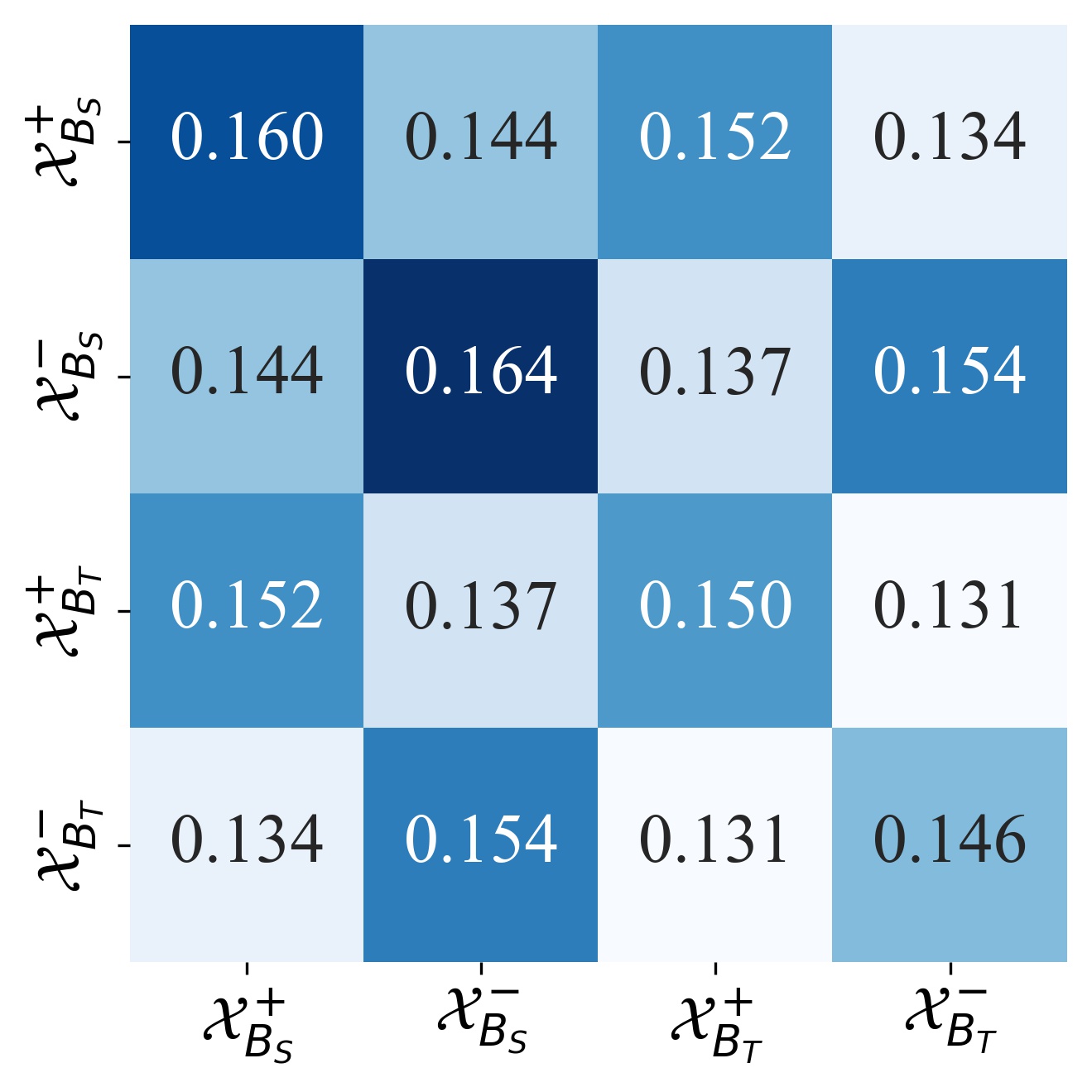}
        \caption{\mbox{\scriptsize{(NP\_000676, NP\_005152)}}}
        \label{fig:diffa}
    \end{subfigure}%
    \begin{subfigure}{0.26\textwidth}
        \centering
        \includegraphics[width=.9\textwidth]{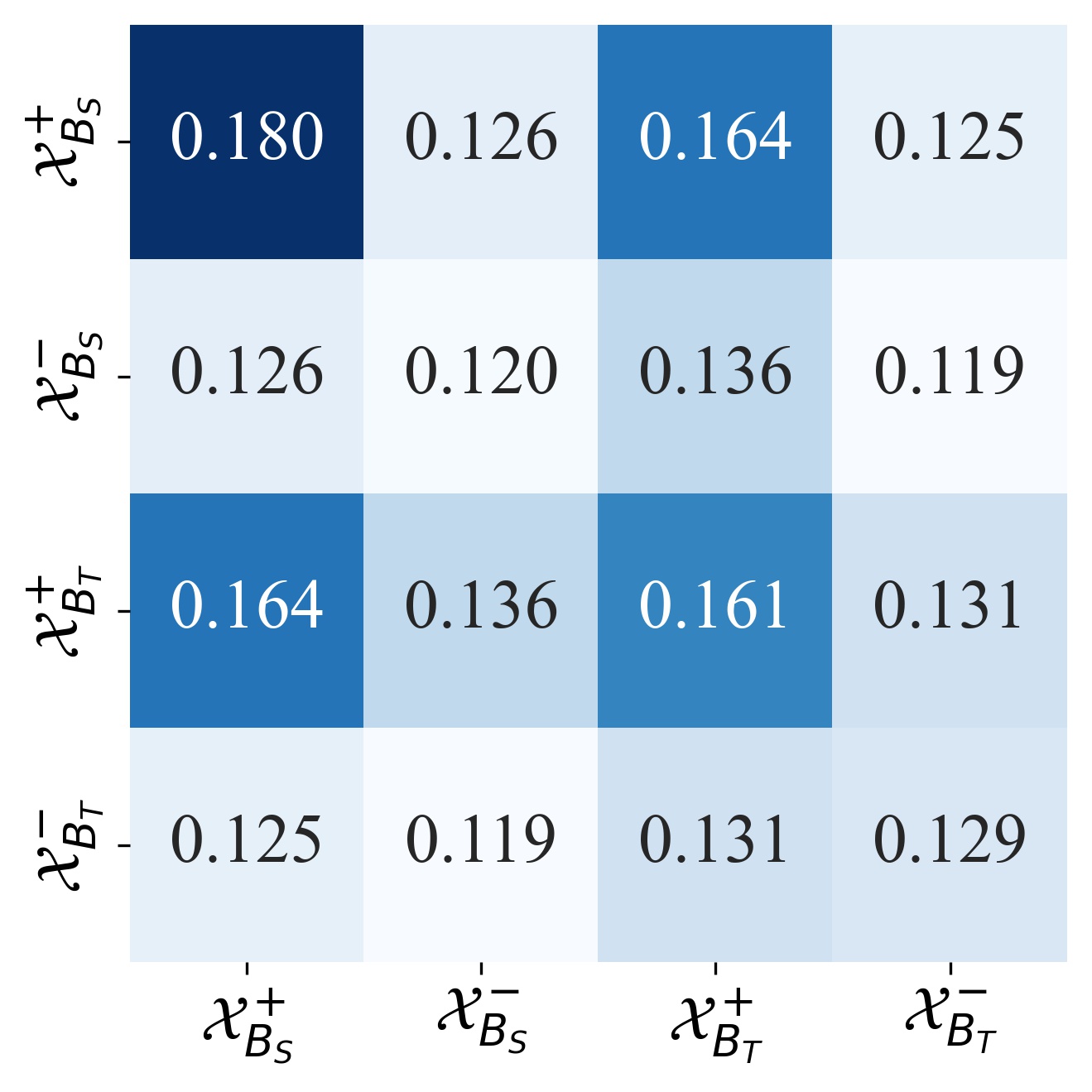}
      	\caption{\mbox{\scriptsize{(P00748, NP\_036559)}}}
	\label{fig:diffb}
    \end{subfigure}
\\
\vspace{10pt}
    \begin{subfigure}{0.26\textwidth}
        \centering
        \includegraphics[width=.9\textwidth]{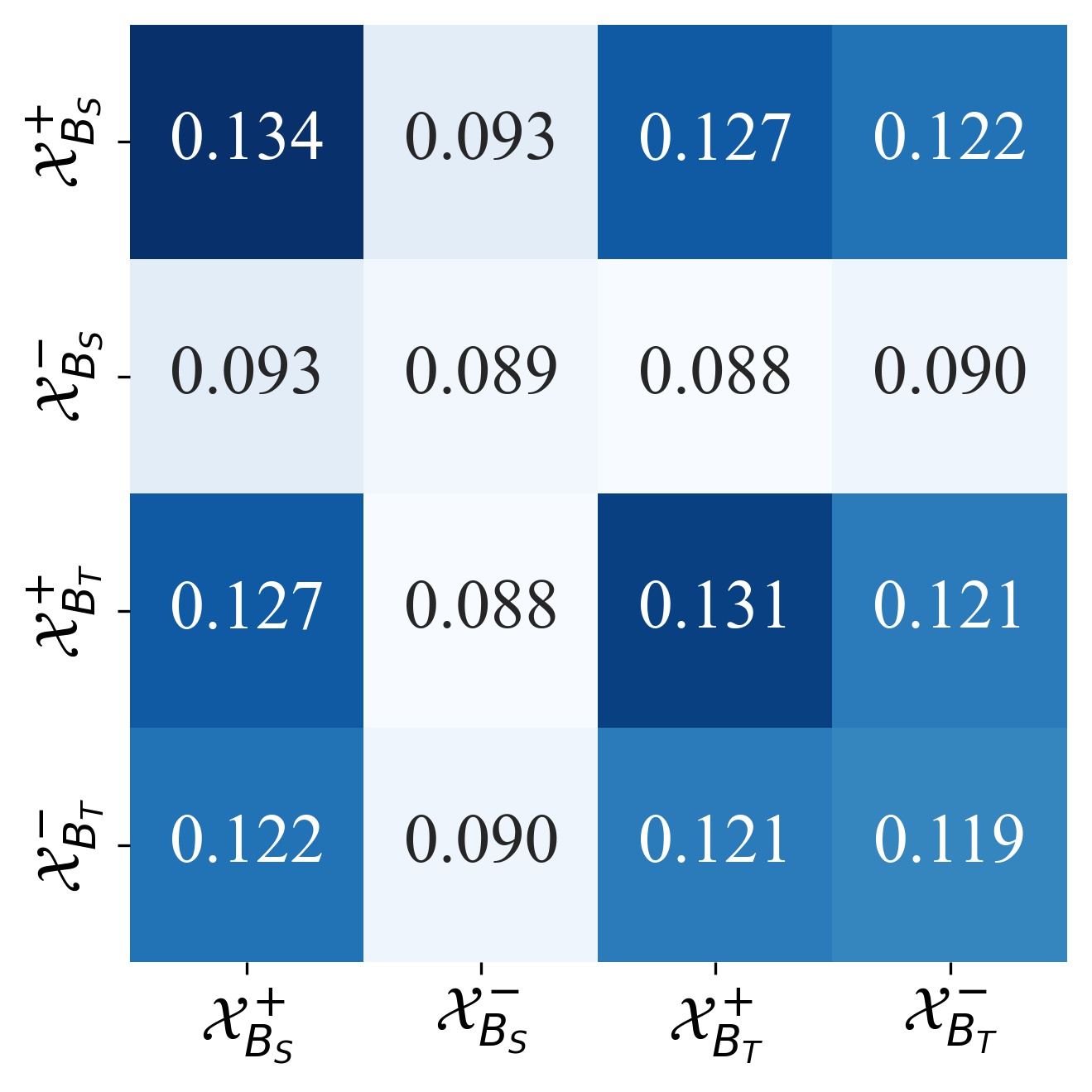}
       	\caption{\mbox{\scriptsize{(AAI28575, NP\_066285)}}}
	\label{fig:diffc}
    \end{subfigure}
    \begin{subfigure}{0.26\textwidth}
        \centering
        \includegraphics[width=.9\textwidth]{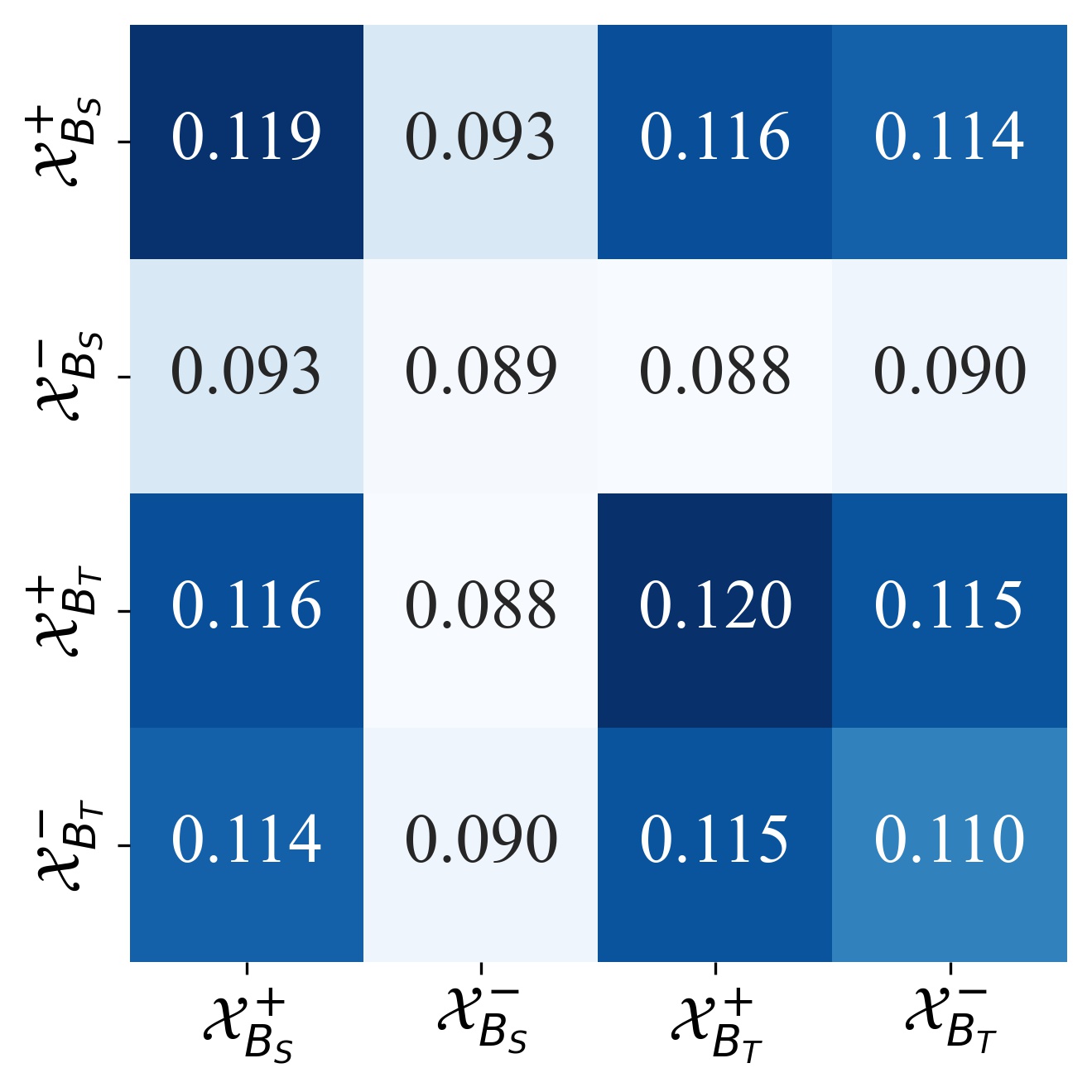}
     	\caption{\mbox{\scriptsize{(AAB26273, NP\_003605)}}}
	\label{fig:diffd}
    \end{subfigure}
    \caption{Similarity matrices of target pairs with significant \rocauc improvement{/degradation}}
    \label{fig:perf_diff_similarity}
\end{figure*}


Among 240 tasks, we identified and studied 4 tasks with significant
performance difference in \rocauc from \OurMethodWithDMPNNA over the
best no-transfer baseline method (i.e., \newgnnFCN).
%
Figure~\ref{fig:perf_diff_similarity} presents the average pairwise similarity matrices of the 4 task pairs
(captions include the corresponding bioassay PubChem AIDs of the source bioassay and the target bioasasy), 
where Figure~\ref{fig:diffa} and~\ref{fig:diffb} have the target tasks that are significantly improved by \OurMethodWithDMPNNA, 
and Figure~\ref{fig:diffc} and~\ref{fig:diffd} have the target tasks that are significantly degraded. 
%
%
%
%
In the figure, $\SetCompounds_{\scriptsize \bioassay_S}^{+}$, $\SetCompounds_{\scriptsize \bioassay_S}^{-}$, 
$\SetCompounds_{\scriptsize \bioassay_T}^{+}$ and $\SetCompounds_{\scriptsize \bioassay_T}^{-}$ denote the active ($^+$)
and inactive ($^-$) compounds for the source ($\bioassay_S$) and target ($\bioassay_T$) tasks,  
%
and average compound similarities ($\text{sim}$) were calculated using Tanimoto coefficient 
over Morgan-count fingerprints (with radius=3 and dimension=2,048). 

In Figure~\ref{fig:diffa} and~\ref{fig:diffb}, the performance of the target task NP\_005152 and NP\_036559 was improved from 
\OurMethodWithDMPNNA over \newgnnFCN by 34.13\% and 27.10\%, respectively. 
%
%
Figure~\ref{fig:diffa} and~\ref{fig:diffb} show that for these two target tasks, 
$\text{sim}(\SetCompounds_{\scriptsize \bioassay_S}^{+}, \SetCompounds_{\scriptsize \bioassay_T}^{+})$
(0.152 in Figure~\ref{fig:diffa}, 0.164 in Figure~\ref{fig:diffb})
is notably greater than both
$\text{sim}(\SetCompounds_{\scriptsize \bioassay_S}^{+}, \SetCompounds_{\scriptsize \bioassay_T}^{-})$ 
(0.134 in Figure~\ref{fig:diffa}, 0.125 in Figure~\ref{fig:diffb}) and
$\text{sim}(\SetCompounds_{\scriptsize \bioassay_S}^{-}, \SetCompounds_{\scriptsize \bioassay_T}^{+})$
(0.137 in Figure~\ref{fig:diffa}, 0.136 in Figure~\ref{fig:diffb}).
This indicates that if active compounds across bioassays are more similar than compounds 
with different activity labels across bioassays,
\OurMethodWithDMPNNA can better capture the commonalities among those
similar active compounds and can better transfer relevant information across bioassays. 
This transferred information can effectively improve the target task performance.
On the other hand, if compounds with different activity labels across bioassays are
more similar than compounds with same activity labels, \OurMethodWithDMPNNA
can cause transfer of conflicting information.
Such a transfer can result in performance degradation for the target task.
Such performance degradation in \rocauc
from \OurMethodWithDMPNNA over \newgnnFCN for the target tasks in pairs
(AAI28575, NP\_066285) in Figure~\ref{fig:diffc} was 5.74\% and (AAB26273, NP\_003605) in Figure~\ref{fig:diffd} was 2.34\%, 
respectively. 
%
In Figure~\ref{fig:diffc} and~\ref{fig:diffd}, 
$\text{sim}(\SetCompounds_{\scriptsize \bioassay_S}^{+}, \SetCompounds_{\scriptsize \bioassay_T}^{+})$ and
$\text{sim}(\SetCompounds_{\scriptsize \bioassay_S}^{+}, \SetCompounds_{\scriptsize \bioassay_T}^{-})$ values are relatively similar 
(0.127 vs 0.122 in Figure~\ref{fig:diffc}, 0.116 vs 0.114 in Figure~\ref{fig:diffd}). 
This indicates that 
when the similarities between $\SetCompounds_{\scriptsize \bioassay_S}^{+}$ and
$\SetCompounds_{\scriptsize \bioassay_T}^{-}$ compounds are relatively high, 
\OurMethodWithDMPNNA could lead to transfer of conflicting information, 
causing inactive compounds in the target bioassay to be incorrectly classified as active.
%

\begin{figure}[h!]
    \centering
        \includegraphics[width=0.5\textwidth]{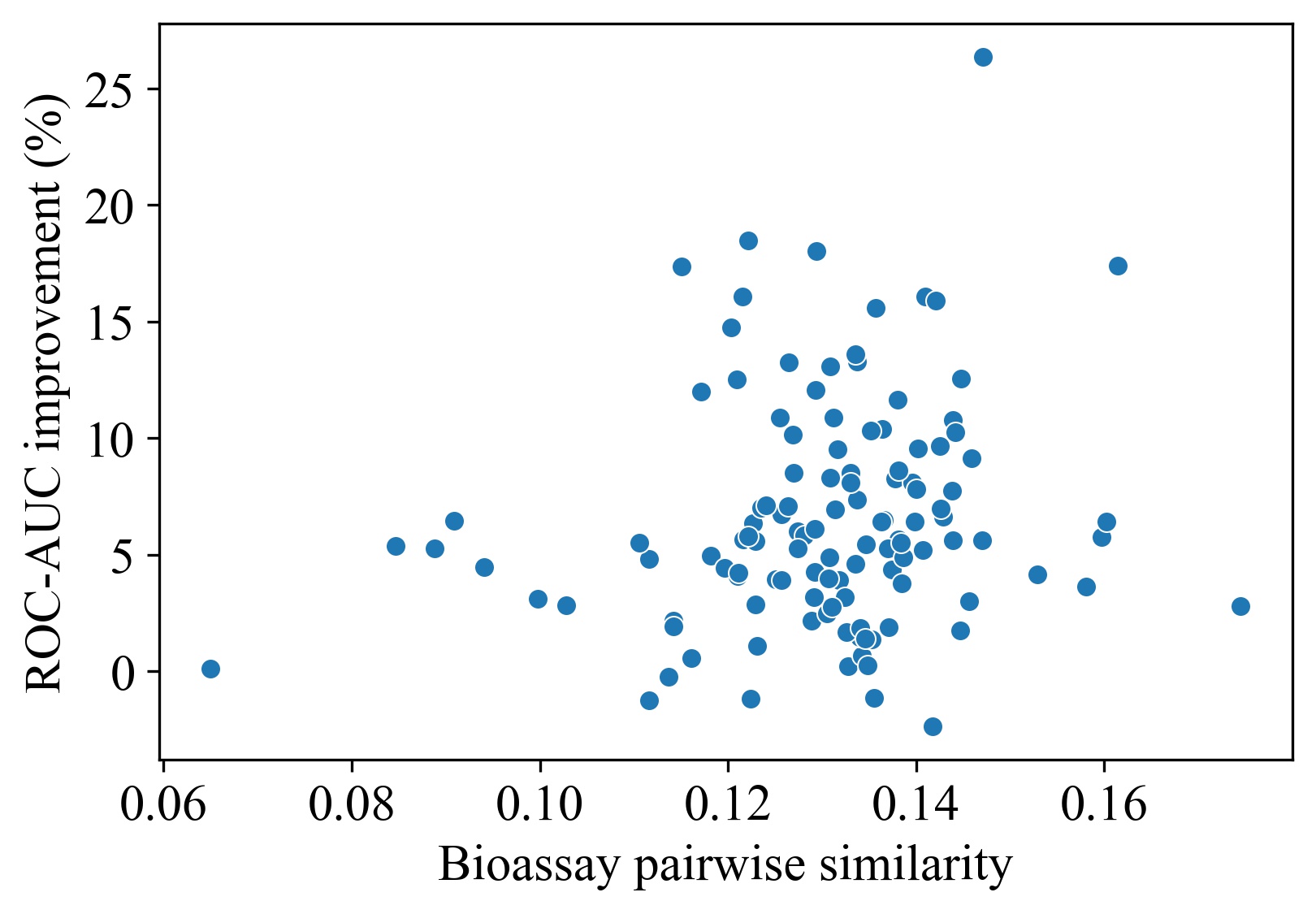}
      \caption{{\rocauc} improvement from {\OurMethod} over {\newgnnFCN} vs bioassay similarity}
    \label{fig:perf_impv_sim}
\end{figure}


Furthermore, we analyzed the relation between the task-wise {\rocauc} improvement from
{\OurMethod} over {\newgnnFCN}, and the bioassay similarities.
%
Figure{~\ref{fig:perf_impv_sim}} presents such relation.
Note that the bioassay similarities are calculated as the average of all pairwise compound similarities across two bioassays in the same way as discussed in 
section~\ref{sec:materials:dataset:pairs}.
%
%
Figure{~\ref{fig:perf_impv_sim}} demonstrates 
that there are significant task-wise {\rocauc} improvement
(e.g., in the upper right region)
when the pairwise similarities are relatively high 
(e.g., greater than 0.12);
and there are marginal improvement (e.g., in the lower left region)
when the pairwise similarities are low (e.g., lower than 0.12).
%
%
This suggests that if bioassay pairs are more similar,
{\OurMethod} can improve the performance over {\newgnnFCN}
by a large margin (e.g., more than 10\%).
On the other hand, if bioassay pairs are less similar,
{\OurMethod} achieves little or no improvement over {\newgnnFCN}. 
%
Indeed, there are some bioassay pairs that are more similar,
yet {\OurMethod} achieves marginal or negative improvement 
(e.g., in the lower middle region).
This is possibly due to the fact that the performance improvement is not only a function of bioassay similarity;
in fact, the improvement can be marginal or negative owing to poor generalization during testing.

\subsubsection{Correctly classified compounds possibly due to more similar compounds in the source bioassay}
\label{sec:results:cases:compounds}

\begin{figure*}[h!]
    \centering
    \begin{subfigure}{\textwidth}
        \centering
        \includegraphics[width=\textwidth]{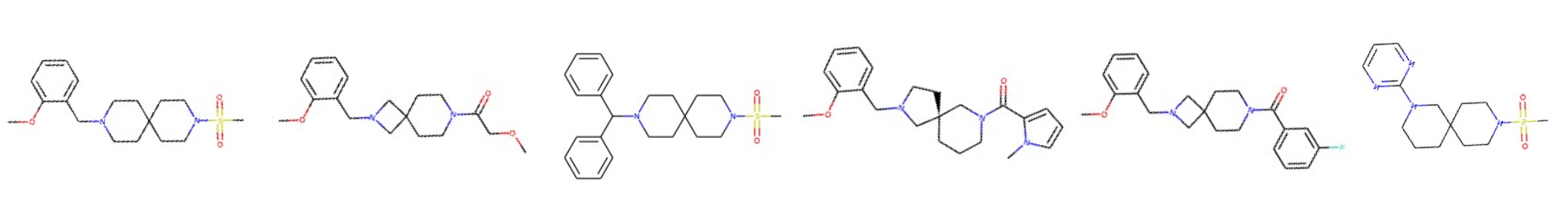}
        \caption{\scriptsize{A \emph{correctly} classified compound from the target bioassay NP\_000752 and its top-5 most similar compounds from the source bioassay NP\_000762}}
    \end{subfigure}
    
    \begin{subfigure}{\textwidth}
        \centering
        \includegraphics[width=\textwidth]{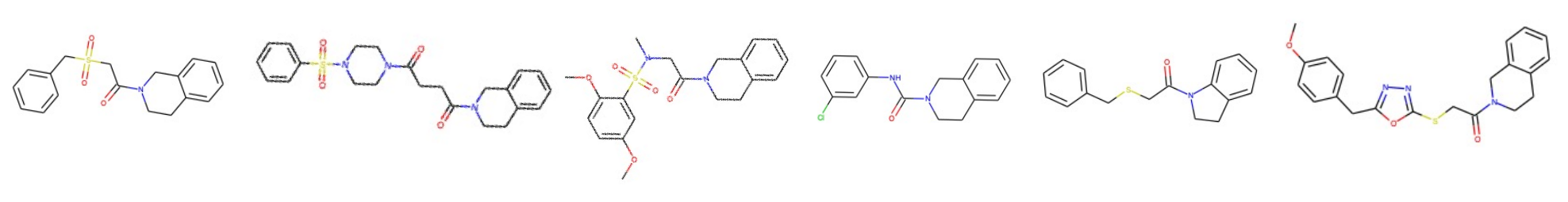}
        \caption{\scriptsize{A \emph{correctly} classified compound from the target bioassay NP\_660205 and its top-5 most similar compounds from the source bioassay NP\_004337}}
    \end{subfigure}
    
    \begin{subfigure}{\textwidth}
        \centering
        \includegraphics[width=\textwidth]{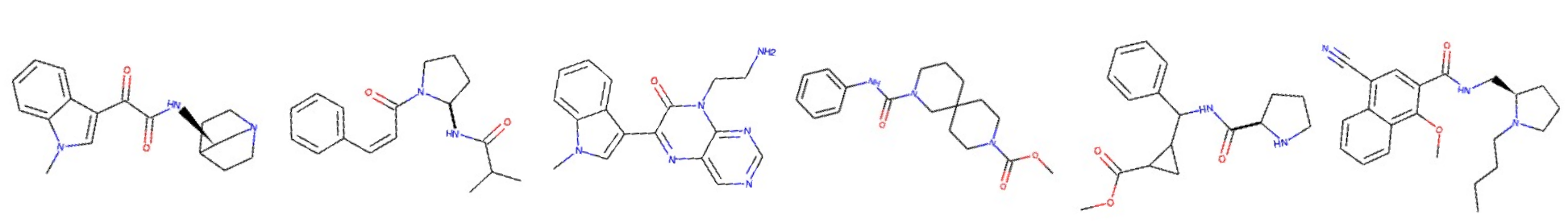}
        \caption{\scriptsize{An \emph{incorrectly} classified compound from the target bioassay NP\_000752 and its top-5 most similar compounds from the source bioassay NP\_000762}}
    \end{subfigure}
    
    \begin{subfigure}{\textwidth}
        \centering
        \includegraphics[width=\textwidth]{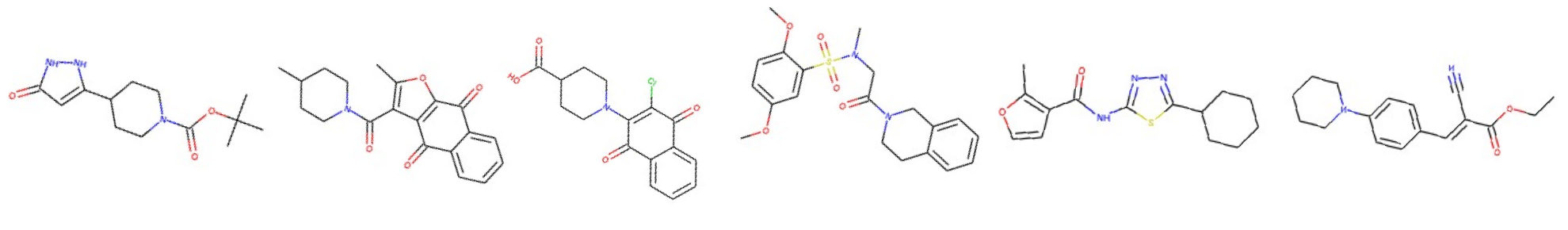}
         \caption{\scriptsize{An \emph{incorrectly} classified compound from the target bioassay NP\_660205 and its top-5 most similar compounds from the source bioassay NP\_004337}}
    \end{subfigure}

    \caption{Visualization of a few selected compounds from the target bioassay and their corresponding top-5 most similar compounds from the source bioassay.}
    \label{fig:compounds}
\end{figure*}


In this section, we identified (i) a few compounds
that were correctly classified by
{\OurMethod} but incorrectly classified by the baselines,
and (ii) a few compounds incorrectly classified by {\OurMethod} but correctly classified by the baselines.
Figure{~\ref{fig:compounds}} (a) and (b) present two such examples for (i); and
Figure{~\ref{fig:compounds}} (c) and (d) present two such examples for (ii).
In each figure, the left-most compound is the compound 
to be classified as active/inactive
from the target bioassay
(referred as $\mathbf{x}^{(T)}$)
and the others are the top-5 most similar compounds  
(referred as $\SetCompounds_{\scriptsize \bioassay_S}^{\ast}$)
to $\mathbf{x}^{(T)}$ from 
the corresponding source bioassay.
The mean pairwise Tanimoto coefficient between $\mathbf{x}^{(T)}$ and 
$\SetCompounds_{\scriptsize \bioassay_S}^{\ast}$
in Figure{~\ref{fig:compounds}} (a), (b), (c), and (d) are 
0.407, 0.428, 0.210 and 0.143, respectively.
Thus, in Figure{~\ref{fig:compounds}} (a) and (b), $\mathbf{x}^{(T)}$s are
structurally more similar to their corresponding $\SetCompounds_{\scriptsize \bioassay_S}^{\ast}$. 
Relatively, in  Figure{~\ref{fig:compounds}} (c) and (d), $\mathbf{x}^{(T)}$s are
less similar to their corresponding $\SetCompounds_{\scriptsize \bioassay_S}^{\ast}$. 
%
This suggests that
{\OurMethod} classifies some compounds correctly probably due to the fact that
those compounds have very similar compounds in the source bioassay.
\subsection{Compound Prioritization using \newgnn}
\label{sec:results:ranking}
We also explored the potential of using \newgnn for compound prioritization purposes. 
We develop a comprehensive 
learning-to-rank method \RankMethod for effective compound prioritization
that jointly learns molecular graph representations via \gnn and a scoring function
using the representations.
The learning methods for compound prioritization is described in Section 4 in the Supplementary
Materials.

%
\subsubsection{Materials}
\label{sec:results:ranking:materials}
\paragraph*{Baselines:}
%
We compare {\RankMethod} with the following feature vectors 
using the same scoring and loss functions:
(i) binary Morgan fingerprints (\MORGAN), 
(ii) morgan count fingerprints, (\MORGANC), 
(iii) bioassay-specific compound features~\cite{Liu2017Diff} computed using Tanimoto coefficient 
on binary Morgan fingerprints (\MORGANBA),
(iv) 200-dimensional  RDKit descriptors (\RDKITGD), and
(v) directed message passing network~\cite{Yang2019} (\dmpnn).
We generate binary Morgan fingerprints and Morgan-count fingerprints 
(with radius = 2 and size = 2,048) using RDKit~\cite{landrum2020rdkit}.
Codes for computing the RDKit descriptors are available in the Descriptastorus package~\cite{descriptastorus}.

\paragraph*{Experimental Protocol:}
In order to evaluate the overall ranking performance, 
we perform 5-fold cross validation. 
We randomly split each bioassay into five folds. 
%
%
%
In each run, four folds of each bioassay are used for training and the other fold is used for testing. 
%
We record optimal values of each performance metric averaged over the five folds. 
Finally, we report the average of all such recorded optimal values of each performance metric over all the bioassays.
For each bioassay, we train the models using Adam~\cite{Kingma2015} optimizer 
with an initial learning rate $\in~\{\text{5e-3, 1e-3, 5e-4}\}$. 
We use grid-search to tune all the hyperparameters such as 
the dimension of the graph representation $d$, 
hidden dimension of the attention layer and batch size.
Specifically, we use $d~\in~\{\text{25, 50, 100}\}$ for \dmpnn and \newgnn,
hidden dimension of the attention layer $\in~\{\text{5, 10, 20}\}$ for \newgnn.
We use batch size $\in~\{\text{128, 256, 512}\}$, and $\lambda \text{ = 1e-6}$
for all the models.   
All the models are trained for 50 epochs. 
%
%

\paragraph*{Evaluation:}
We evaluate all the methods using a set of 105 single-target
confirmatory bioassays from PubChem~\cite{kim2021pubchem}.
These bioassays all use IC$_{50}$ to measure compound binding affinities and have at least 50 active compounds.
For each bioassay, we only keep the active compounds, and remove duplicate compounds and those with identical IC$_{50}$ values. 
We evaluate the ranking performance using 
concordance index (CI), recall@$k$ (R@$k$),
Normalized Discounted Cumulative Gain@$k$ (ndcg@$k$)~\cite{Liu2017Diff}, where $k=3,5,10$.
We also use R@$k$\% and ndcg@$k$\%, where we consider top $k$\% ($k$=5,10) of the test fold compounds in $\mathit{r}$.

\subsubsection{Results}
\label{sec:results:ranking:results}

\begin{table}[h]
\sisetup{input-decimal-markers={.}, round-mode=places, round-precision=3}  
\centering
\caption{Overall Performance Comparison of \RankMethod} 
\label{tbl:best_ranking}
\begin{small}
  \begin{threeparttable}
      \begin{tabular}{
          @{\hspace{5pt}}l@{\hspace{5pt}}
          @{\hspace{5pt}}r@{\hspace{5pt}}
          @{\hspace{5pt}}r@{\hspace{5pt}}
          @{\hspace{5pt}}r@{\hspace{5pt}}
          %
          @{\hspace{5pt}}r@{\hspace{5pt}}
          @{\hspace{5pt}}r@{\hspace{5pt}}
          %
          @{\hspace{5pt}}r@{\hspace{5pt}}
          %
          @{\hspace{5pt}}r@{\hspace{5pt}} 
          }
          \toprule
          method & CI & 
          R\text{@3} & R\text{@5} & 
          ndcg\text{@3} & ndcg\text{@5} & 
          R\text{@5\%} & 
          \small{ndcg\text{@5\%}} 
          \\
          \midrule
			\MORGAN & 0.706 & 0.543 & 0.644 
			& 0.814 & 0.816 
			& 0.420 
			& 0.838 
			\\
			\MORGANC & 0.711 & 0.545 & 0.655 
			& 0.815 & 0.819 
			& 0.437 
			& 0.846 
			\\
			\MORGANBA & 0.687 & 0.500 & 0.626 
			& 0.789 & 0.797 
			& 0.375 
			& 0.816 
			\\
			\RDKITGD & 0.687 & 0.519 & 0.632 
			& 0.790 & 0.797 
			& 0.396 
			& 0.813 
			\\
			\dmpnn & \underline{0.731} & \underline{0.643} & \underline{0.709} 
			& \underline{0.854} & \underline{0.847} 
			& \underline{0.579} 
			& \underline{0.896} 
			\\
			\midrule
			\newgnn & \textbf{0.748} & \textbf{0.686} & \textbf{0.740} 
			& \textbf{0.881} & \textbf{0.867} 
			& \textbf{0.686} 
			& \textbf{0.936} 
			\\			
			\diff & 2.353 & 6.608 & 4.460 
			& 3.114 & 2.421 
			& 18.428 
			& 4.475 
			\\
			\taskDiff & 2.535 & 7.645 & 4.720 
			& 3.406 & 2.569 
			& 24.578 
			& 4.979 
			\\
			\pvalue & 1.14e-10 & 4.89e-10 & 9.87e-13 
			& 1.42e-12 & 2.25e-12 
			& 3.95e-15 
			& 1.83e-11 
			\\
 		\bottomrule
		\end{tabular}
		\begin{tablenotes}
        \setlength\labelsep{0pt}
		\begin{footnotesize}
		\item
		In this table, the columns have the respective average of each performance metric 
		over all bioassays obtained by the 	
		respective optimal hyperparameter settings.
		The best/second best performance under each metric is \textbf{bold}/\underline{underlined}. 
		\end{footnotesize}
		\end{tablenotes}
\end{threeparttable}
\end{small}
\end{table}

Table~\ref{tbl:best_ranking}  presents the performance comparison 
between {\newgnn}, {\dmpnn} and the baselines. 
Overall, \newgnn significantly performs better than all the baselines
including \dmpnn, across all performance metrics.
The average performance improvement from \newgnn over \dmpnn in terms of
CI, recall@3, recall@5, 
ndcg@3, ndcg@5, 
recall@5\%,
ndcg@5\% and 
is 2.353\%, 6.608\%, 4.460\%, 
3.114\%, 2.421\%, 
18.429\%, 
and 4.475\%, 
respectively.
Furthermore, compared to {\dmpnn}, 
the average bioassay-wise performance improvement from \newgnn is most significant 
in terms of recall@3, recall@5, ndcg@3, ndcg@5, recall@5\%, 
and ndcg@5\% 
(p-values: 4.89e-10, 9.87e-13, 1.42e-12,
2.25e-12, 3.95e-15, 
and 1.83e-11, 
respectively).
%
This indicates that \newgnn can rank the top-most 
compounds better than \dmpnn.
Unlike mean pooling in {\dmpnn}, attention mechanism in \newgnn 
can differentially focus on atoms based 
on the relevance of each atom to the prioritization problem. 
This demonstrates the ability of \newgnn to better differentiate compounds and to achieve effective compound prioritization.
Furthermore, {\newgnn} and  {\dmpnn} significantly outperforms 
all the fingeprint-based baselines across all performance metrics. 
Compared with the best performing fingerprint-based baseline {\MORGANC}, 
in terms of 
CI, recall@3, recall@5, 
ndcg@3, ndcg@5, 
recall@5\%, 
and ndcg@5\%, 
the average performance improvement from \newgnn is
5.247\%, 25.724\%, 13.094\%, 
8.140\%, 5.871\%, 
56.875\%, 
and 10.637\%, 
respectively;
and 
from \dmpnn in terms of 
CI, recall@3, recall@5, 
ndcg@3, ndcg@5, 
recall@5\%, 
and ndcg@5\% 
is 
2.827\%, 17.932\%, 8.266\%, 
4.874\%, 3.369\%, 
32.464\%, 
and 5.898\%, 
respectively. 
This demonstrates that the learned representation out of \RankMethod can effectively encode 
useful molecular substructure information,  
and thus, are more effective for compound prioritization.
%
%
%
%

%

\section{Conclusions}
\label{sec:conclusions}


We have developed \OurMethod that effectively leverages source bioassay data
to improve the performance of the target task. 
We also proposed a variant of \OurMethod, i.e., \OurMethodWithDisc that additionally 
learns feature-wise and compound-wise transferability.
We conducted an exhaustive array of experiments and analyses that suggest
that \OurMethodWithDMPNNA is the best-performing method on average across all target tasks.
The proposed variant is also a very strong method and even better compared to \OurMethod
on certain target tasks.
Furthermore, in ablation studies, 
we also showed that \OurMethodWithDMPNNAWithGlobalDisc can even improve 
performance for more than half of the target tasks compared to \OurMethodWithDMPNNA.
Our analyses further demonstrated that learning compound-wise transferability via \gdisc can
better encode the commonalities between compounds across bioassays.
We also provided a parameter study to demonstrate the effect of $\alpha$ and $\lambda$
on our proposed methods.
Additionally, we demonstrated the efficacy of our proposed \newgnn 
in both compound activity and compound prioritization problems
since it performed better than 
any other compound representation methods.

In this work, we paired the bioassays if their corresponding protein targets 
belong to the same protein family.
In other words, when we paired the bioassays, the corresponding pair of tasks
are assumed to be related.
We assumed that leveraging activity information from related protein targets 
(i.e., targets belonging to the same protein family)
can improve the target task performance.
However, we observed \OurMethod did not improve all the targets compared 
to the best no-transfer baseline method \newgnnFCN.
This suggests occurrence of potential negative transfer.
In future works, we will focus on developing a more principled approach 
to determine task-relatedness.
Given a target task, our current method only consider a single source task.
This severely limits the scope of transfer from only one related task, and can also 
impact the performance on the target task if the learning is too focused on the source task.
Our future work will incorporate multiple source tasks for each target task
by simultaneously learning task-relatedness in a data-driven manner.

%


\section{Computational Methods}
\label{sec:comp_methods}

\subsection{Notations and Definitions}
\label{sec:defs}

\begin{table}[hbt!]
\centering
\caption{Notations}
\label{tbl:defs}
\begin{small}
  \begin{threeparttable}
      \begin{tabular}{
          @{\hspace{5pt}}l@{\hspace{5pt}}
          @{\hspace{5pt}}l@{\hspace{5pt}}
          }
          \toprule
          method &  meanings \\ 
          \midrule

          \compound/\bioassay & compound/bioassay \\
          \MolGraph = (\SetAtoms, \SetBonds) & molecular graph with set of 
          atoms \SetAtoms and bonds \SetBonds  \\
          %
          $u$ & an atom in \MolGraph \\
          $(u, v)$ & a bond connecting atoms $u$ and $v$ in \MolGraph \\
          $\mathcal{N}(u)$ & neighbors of atom $u$ in \MolGraph \\
          %
          \SetCompounds & set of compounds in a bioassay \\
          \SetLabels & set of labels corresponding to \SetCompounds \\
          \FeatureSpace & input feature space \\
          \LabelSpace & label space \\
          $\domain = \{\FeatureSpace, P(\SetCompounds)\}$ & a domain consisting of
          \FeatureSpace and marginal probability distribution P(\SetCompounds) \\
          $\task = \{\LabelSpace, \omega(\cdot)\}$ & a task consisting of label space and a 
          decision function $\omega(\cdot)$  \\
          %
		 \hidden & hidden state \\
       	  \compr & molecular representation out of \gnn \\
       	  \scompr & scaled molecular representation \\

	\bottomrule
	\end{tabular}
\end{threeparttable}
\end{small}
\end{table}

%
In this section, we listed the notations and definitions used in this paper.
Table \ref{tbl:defs} presents a list of notations and their meanings.
%
We represent a compound \compound using a molecular graph $\MolGraph_{\scriptsize \compound}$.
$\MolGraph_{\scriptsize \compound}$ is denoted as 
$\MolGraph_{\scriptsize \compound} =  (\SetAtoms_{\scriptsize \compound}, \SetBonds_{\scriptsize \compound})$,
where $\SetAtoms_{\scriptsize \compound}$ is the set of atoms, and 
$\SetBonds_{\scriptsize \compound}$ is the set of corresponding bonds in \compound.
We denote the set of compounds in a bioassay \bioassay as  
$\SetCompounds_{\scriptsize \bioassay}$, and 
the activity labels of those compounds accordingly as $\SetLabels_{\scriptsize \bioassay}$.
In this paper, we use a label `1' or `0' to indicate that a compound is active or inactive in a bioassay, respectively.
%
%

We use the following definitions related to transfer learning.
\begin{itemize}
\item Domain: a domain \domain is a set of labeled compounds 
$\domain = \{\compound_i = (\InputFeature_i, \Label_i) | \InputFeature_i \in \FeatureSpace, \Label_i \in \LabelSpace, i = 1,..., |\domain|\}$, 
where the compounds $\{\InputFeature_i\}$ are represented in a feature space \FeatureSpace, 
and their activity labels $\{\Label_i\}$ are represented in a label space \LabelSpace; 
$|\domain|$ is the size of the domain (i.e., the number of $(\InputFeature_i, \Label_i)$ pairs).   
In our transfer learning, we will have two domains: 
a source domain, denoted as \SourceDomain, and a target domain, 
denoted as \TargetDomain.
In general, these two domains can have different numbers of compounds with different 
compound feature representations, and also different label sets. 
We use superscript $^{(S)}$ and $^{(T)}$ to represent information associated with the source domain and the target domain, 
respectively. For example, $\InputFeature^{(T)}$ represents a compound from the target domain. 
In addition, we use \SetCompounds to represent the set of compound features $\{\InputFeature_i\}$, that is, 
$\SetCompounds = \{\InputFeature_i| \InputFeature_i \in \FeatureSpace\}$, 
and \SetLabels to represent the set of compound labels, that is, $\SetLabels = \{\Label_i| \Label_i \in \LabelSpace\}$. 
Thus, \domain can also be represented as $\domain = (\SetCompounds, \SetLabels)$. 


%
\item Task: Given a domain $\domain = \{(\InputFeature_i, \Label_i)_{i=1, ..., |\scriptsize{\domain|}}\}$, 
a task \task is to learn a model that maps each $\InputFeature_i$ to its corresponding $\Label_i$. 
In our transfer learning, we will have two tasks: a source task, denoted as \SourceTask, and a target task, 
denoted as \TargetTask. \SourceTask and \TargetTask learn from the source domain \SourceDomain and 
the target domain \TargetDomain, respectively. 
\item Transfer Learning: Transfer learning learns and transfer information from source task \SourceTask to 
the target task \TargetTask, and helps improvement the performance of \TargetTask. 
The underlying assumptions are 1) the target domain \TargetDomain does not have 
sufficient information for \TargetTask to learn a good model; and 2) there are commonalities between 
\SourceDomain and \TargetDomain and such commonalities can be transferred from \SourceDomain to \TargetDomain, 
and used to improve \TargetTask. 
%

\end{itemize}
%

%
%
\subsection{Methods}
\label{sec:methods}


In this section, we present our two transfer learning methods:
{\OurMethod} and {\OurMethodWithDisc}.
We first introduce the overall architecture of {\OurMethod} in section{~\ref{sec:methods:overall}}, and then
discuss each component in detail in subsequent sections 
(i.e., sections{~\ref{sec:methods:conv}} and {\ref{sec:methods:classifier}}).
We discuss the end-to-end optimization process in section{~\ref{sec:methods:optimization}}.
We then introduce {\OurMethodWithDisc} with additional components 
that learns feature-wise and compound-wise transferability, and finally discuss
the optimization process in section{~\ref{sec:methods:variations}}.

\subsubsection{Overall Architecture of \OurMethod}
\label{sec:methods:overall}

\begin{figure*}
\includegraphics[width=\textwidth]{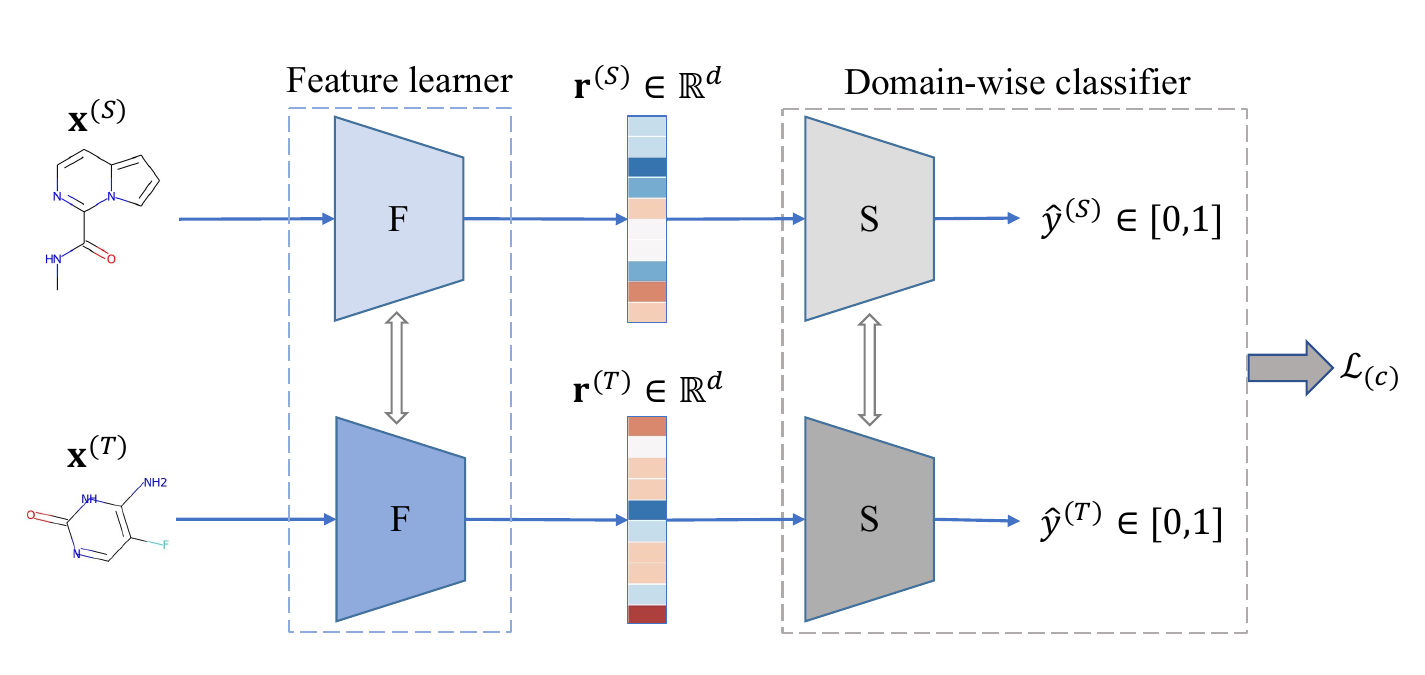}
\caption{{Proposed architecture of \OurMethod. 
{The feature learner {\fgen} learns compound representations {\compr} 
given the corresponding molecular graph. 
The domain-wise classifier {\classifier} classifies the compound as active/inactive.}}
}
\label{fig:aada}
\end{figure*}

\OurMethod learns to generate transferable features that can 
generalize well from one domain to another, 
and increase the predictive power for classification in the target domain.
Figure \ref{fig:aada} presents the overall architecture of the proposed \OurMethod.
%
\OurMethod consists of two components: 
1) a feature learner \fgen, 
that learns to represent chemical compounds; 
and 
2) a domain-wise classifier \classifier, 
that classifies chemical compounds of each domain.
%
Below, we discuss each component of \OurMethod in detail.
\subsubsection{Learning Compound Representations}
\label{sec:methods:conv}
{This section describes how the feature representations
of compounds are learned.
}
In \OurMethod, the feature representations of chemical compounds are learned in a data-driven fashion. 
Compared to using static fingerprints or fixed feature representations of molecular structures~\cite{rogers2010extended}, 
such learned features will be more adapted to the learning task and enable optimal performance. 
We leverage the popular idea of graph neural networks~\cite{Scarselli2009}, and use the Directed Message Passing Neural Network, 
denoted as \dmpnn, developed in Yang \latin{et al.}~\cite{Yang2019}. 
%
%
Given a molecular graph $\MolGraph_{\scriptsize{\compound}} = (\SetAtoms_{\scriptsize{\compound}}, \SetBonds_{\scriptsize{\compound}})$ 
for a compound \compound, 
\dmpnn 
learns a feature vector, also called an embedding, of \compound
using graph convolution,  
%
by passing messages along directed edges over molecular graphs. 
In \dmpnn, two representations for each bond are learned via message passing through the two directions along
the bond. 
Then atom representations are learned from the representations of their incoming bonds. 
In the end, the compound representation is generated via mean pooling over all the atom representations. 
Details about \dmpnn are presented in Section 1 in the Supplementary Materials, and also available 
in Yang \latin{et al.}~\cite{Yang2019} 

%
Based on \dmpnn, we further improve the compound representation learning by 
introducing an attention mechanism
{inspired from Graph Attention Networks{~\cite{Velickovic2018}}}. 
This new method is referred to as \dmpnn with attention, 
denoted as \newgnn. 
Specifically, we replace the mean-pooling in \dmpnn with an attention-based pooling mechanism as follows,
%
\begin{equation}
\label{equ:newgnn}
\compr_{\scriptsize{\compound}} = \sum_{u \in \scriptsize{\SetAtoms_{\scriptsize{\compound}}}} 
(\mathbbm{1} + w_u) \odot \embedding_u,
\end{equation}
where $\odot$ is the element-wise product, 
$\embedding_u$ is the learned representation of atom $u$ as in \dmpnn,
$w_u$ is the attention weight on atom $u$ calculated as follows,  
\begin{equation}
\label{eqn:attention}
w_u = \frac{\exp({\fatt(\embedding_u)})}
					{\sum_{v \in \scriptsize{\SetAtoms_{\scriptsize{\compound}}}} \exp({\fatt(\embedding_v)})} 
\end{equation}
where $\fatt(\cdot)$ is a 2-layer feed-forward network 
with a ReLU activation function after the hidden layer. That is, the attention learns a specific weight on each atom.
Thus, the attention mechanism in \newgnn
can differentially focus on atoms based on the relevance of each atom 
towards the final predictive task. The network to learn compound embeddings is denoted as \fgen (i.e., \fgen is \dmpnn or \newgnn). 

\subsubsection{Learning to Classify Compounds of Each Domain}
\label{sec:methods:classifier}
%
{This section describes how the compounds of each domain
 are classified as active/inactive using the learned feature representations.
}
Given the compound embedding \compr,
the domain-wise classifier classifies each compound in a given domain
as active or inactive with respect to that domain
using a two-layer fully-connected neural network \classifier
as follows,
\begin{equation}
\label{eqn:classifier}
\hat{y} = \classifier(\compr), 
\end{equation}
with ReLU at the hidden layer and sigmoid at the output layer.
The outputs of \classifier are the probabilities of input compounds
from source/target domain being active in the
source/target domain.
To learn \classifier, 
the loss function \ClassifierLoss for the classifier 
is defined as follows,
%
\begin{eqnarray}
\label{equ:classifier_loss}
\begin{aligned}
&\ClassifierLoss(\ParGnn, \ParClassifier) = 
				&&		-\alpha \frac{1}{n^{(S)}} \sum_{\scriptsize{\InputFeature^{(S)}} \in \scriptsize{\SetCompounds^{(S)}}} 
						[y^{(S)} \log (\hat{y}^{(S)})
						+ (1-y^{(S)}) \log (1-\hat{y}^{(S)})] \\
				&&&	-\frac{1}{n^{(T)}} \sum_{\scriptsize{\InputFeature^{(T)}} \in \scriptsize{\SetCompounds^{(T)}}} 
						[y^{(T)} \log (\hat{y}^{(T)})
						+ (1-y^{(T)}) \log (1-\hat{y}^{(T)})],
\end{aligned}
\end{eqnarray}
%
where $y^{(S)}/y^{(T)}$ is the ground-truth activity label of each compound in domain $S$/$T$,
$n^{(S)}$/$n^{(T)}$ is the number of compounds
in $\SetCompounds^{(S)}$/$\SetCompounds^{(T)}$ (i.e., $n = |\domain|$),
$\alpha$ is a hyperparameter to trade-off the two classification losses,
and 
$\ParGnn$ and $\ParClassifier$ are learnable parameters of \fgen and \classifier,
respectively.
%
%
Please note that both the source domain and target domain use the same classifier {\classifier}. 
Therefore, if the source and target domain have common compounds or very similar compounds, 
when these compounds have same labels in the two domains, they will induce small classification errors in the both domains; 
when these compounds have different labels in the two domains, they will induce large errors in one domain and small errors in the other. 
By minimizing the loss {$\ClassifierLoss(\ParGnn, \ParClassifier)$}, it will encourage common or similar compounds that have 
same labels in the two domains to be more focused on through learning, and prevent the transfer of conflicting 
information across domains.
\subsubsection{\OurMethod Model Optimization}
\label{sec:methods:optimization}
%
{This section presents the optimization process
of the proposed \OurMethod.
}
\OurMethod constructs an end-to-end transfer learning framework with the 
above two components: 1) feature learner \fgen, 
and 2) domain-wise classifier \classifier. 
We solve for \OurMethod through minimizing the loss function $\ClassifierLoss$.
In other words, we solve the following optimization problem:
\begin{equation}
\label{eqn:optim}
\widehat{\ParGnn}, \widehat{\ParClassifier} = 
\arg \min_{\scriptsize{\ParGnn, \ParClassifier}} \ClassifierLoss(\ParGnn, \ParClassifier),
\end{equation}
where \ParGnn and \ParClassifier are the learnable parameters of \fgen and \classifier, respectively.
Minimizing \ClassifierLoss will minimize the classification error in each domain
while preventing transfer of conflicting information across domains,
hence enabling the feature learner \fgen to learn better compound features 
for effective classification in each domain.
Since the same \fgen and \classifier are used for both the source and target tasks, 
minimizing \ClassifierLoss also enables transfer of relevant information 
through the shared parameters of \fgen and \classifier. 
Intuitively, the amount of transferable information from the source domain
to the target domain is determined by the degree of task-relatedness between those domains.
In this work, the degree of task-relatedness between the source and target domains 
is essentially controlled through the hyperparameter $\alpha$.
We will consider learning task-relatedness or $\alpha$ in
a data-driven manner in our future works.
%


\subsubsection{Variant of \OurMethod: \OurMethodWithDisc}
\label{sec:methods:variations}
\begin{figure*}
\includegraphics[width=\textwidth]{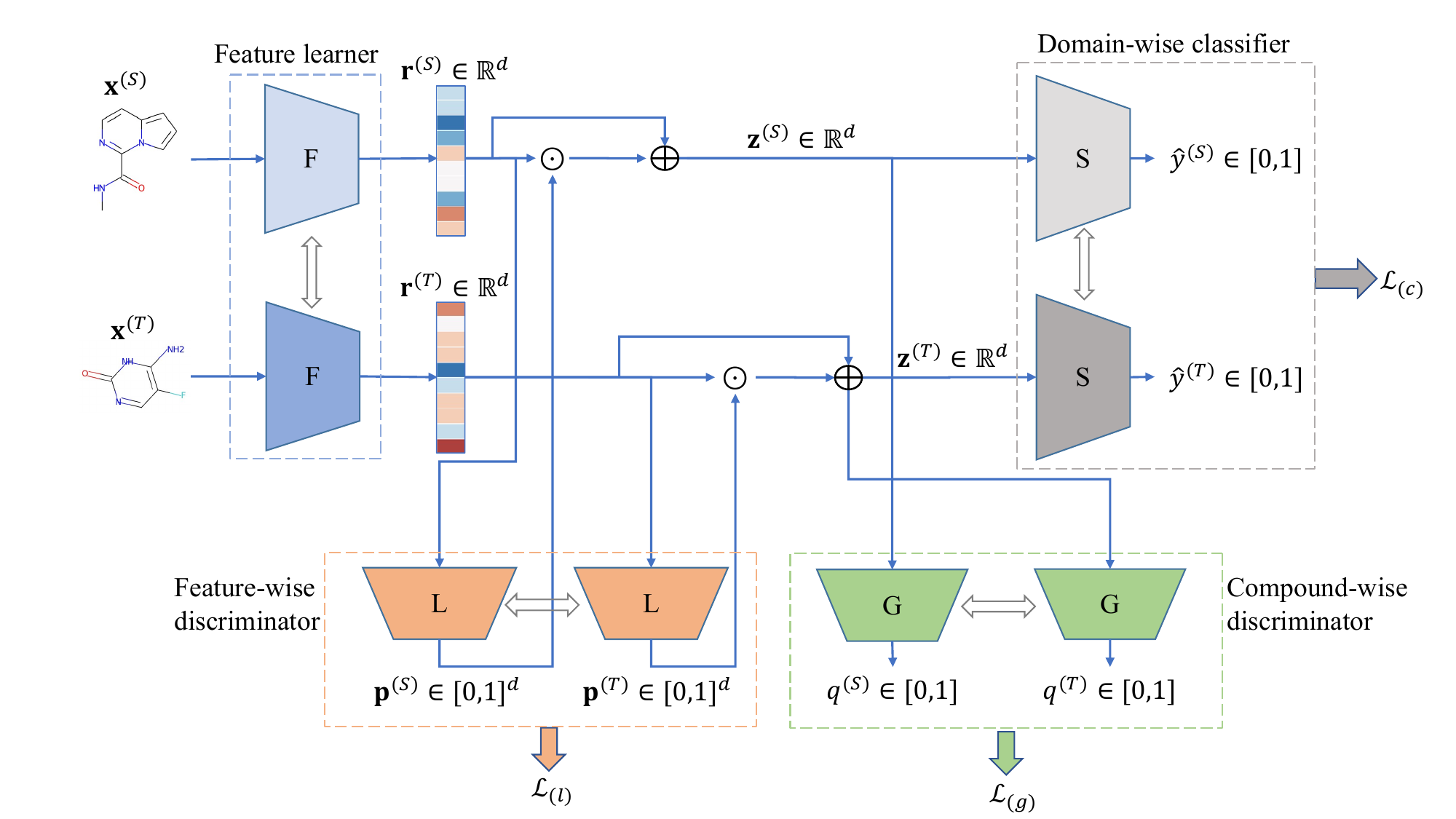}
\caption{Proposed architecture of \OurMethodWithDisc.
{The feature learner {\fgen} learns compound embedding {\compr} 
given the corresponding molecular graph.
The feature-wise discriminator {\ldisc} learns feature-wise transferability given
the learned compound embedding {\compr}.
{\compr} is further scaled into {\scompr} using its
feature entropy from $\mathbf{p}$ out of {\ldisc}.
The compound-wise discriminator {\gdisc} learns the compound-wise transferability
given {\scompr}. 
The domain-wise classifier {\classifier} classifies the compound as active/inactive.}
}
\label{fig:aada_L+G}
\end{figure*}

In this section, we propose a variant of \OurMethod where we
incorporate additional components to selectively learn feature-wise and compound-wise transferability.
We denote this variant as \OurMethodWithDisc. 
Figure \ref{fig:aada_L+G} presents the overall architecture of the 
proposed \OurMethodWithDisc.
In addition to the feature learner \fgen and the label classifier \classifier,
\OurMethodWithDisc consists of two more components: 
1) a feature-wise discriminator {\ldisc}, 
that learns the transferability of each learned feature {(Section{~\ref{sec:methods:variations:local_disc}})}; 
and
2) a compound-wise discriminator {\gdisc}, 
that separates chemical compounds into source and target domains {(Section{~\ref{sec:methods:variations:global_disc}})}.
We refer to \OurMethod with the feature-wise discriminator only as \OurMethodWithLocalDisc, 
and \OurMethod with the compound-wise discriminator only as \OurMethodWithGlobalDisc. 
Below, we discuss each component in detail and also the optimization of the 
proposed method.

\paragraph{Learning Transferability of Individual Features}
\label{sec:methods:variations:local_disc}
\hfill \\
Given the learned compound embedding
$\compr\in \mathbb{R}^d$ out of \fgen (discussed in Section~\ref{sec:methods:conv}), the feature-wise discriminator of \OurMethodWithDisc learns 
the transferability of each embedding feature in \compr using a two-layer neural network \ldisc as follows, 
%
%
%
\begin{equation}
\label{eqn:local}
\mathbf{p} = \ldisc(\compr), 
\end{equation}
where \ldisc has a hidden layer 
with ReLU, and an output layer with sigmoid. 
Note that \mbox{$\mathbf{p} = [p_1, p_2, \cdots, p_d]$} has the same 
dimension as \compr, and $p_i \in [0, 1]$ represents 
the probability that the $i$-th embedding feature  in $\compr$   
is specific to the source domain. 
Thus, the feature-wise discriminator {determines} whether the input compound features (not the input compounds) belong to the 
source domain or not. 
For bioactivity prediction problems, if \SourceDomain and \TargetDomain have compounds for protein targets that are from 
a same protein family, it is very likely that their active compounds are similar and share similar substructures (e.g., pharmacophores). 
In this case, intuitively, the feature-wise discriminator here could learn and represent such similar substructures. 

%
We further quantify the transferability of each embedding feature using its 
entropies as follows, 
\begin{equation}
\entropy(p_i) = -p_i \log p_i - (1-p_i) \log (1-p_i). 
\end{equation}
%
%
If $p_i$ is very large or very small and it will have a low entropy,
it indicates the $i$-th embedding feature is very likely or very unlikely to be specific to the source domain,
and thus it is less likely to be common across domains; 
if $p_i$ is close to 0.5 and with a high entropy, 
the feature is less specific to any of the domains and more likely to be common across domains, and therefore 
can be used for information transfer across domains. 
%

%
%
We then scale compound embedding $\compr$ into \scompr
using feature entropies as follows, 
\begin{equation}
\label{equ:scaled}
\scompr = (\mathbbm{1} + \boldsymbol{\entropy}) \odot  \compr,
\end{equation}
%
%
%
%
where $\boldsymbol{\entropy} = [\entropy(p_1), \entropy(p_2), ... ,\entropy(p_d)] \in \mathbb{R}^d$, and
$\odot$ represents element-wise dot-product. 
{Each feature is scaled with its entropy and added with itself.
Intuitively, the self-addition reduces the loss of informative features
due to improper scaling.}
%
%
Thus, in $\scompr$, domain-invariant embedding features are scaled larger than
domain-specific embedding features ($\boldsymbol{\entropy}\ge 0$). 
We will use \scompr as input to the following components. 

%

To learn the feature-wise discriminator, the loss function \LocalDiscLoss 
is defined as follows, 
%
\begin{eqnarray}
\label{eqn:localloss}
\begin{aligned}
& \LocalDiscLoss(\ParGnn, \ParLdisc)
				& =  & -\frac{1}{n^{(S)}}
				 \sum_{\scriptsize{\InputFeature^{(S)} \in \SetCompounds^{(S)}}} \frac{1}{d}\sum_{i=1...d} \log(p^{(S)}_i) \\
				 && &
				 -\frac{1}{n^{(T)}}
				 \sum_{\scriptsize{\InputFeature^{(T)} \in \SetCompounds^{(T)}}} \frac{1}{d}\sum_{i=1...d} \log(1-p^{(T)}_i), 
\end{aligned}
\end{eqnarray}
where $n^{(S)}$/$n^{(T)}$ is the number of compounds
in $\SetCompounds^{(S)}$/$\SetCompounds^{(T)}$ (i.e., $n = |\domain|$); 
\ParGnn and \ParLdisc are learnable parameters of
 \fgen (compound representation learning network as in Section~\ref{sec:methods:conv}) 
and  \ldisc (feature-wise discriminator network as in Equation~\ref{eqn:local}), respectively; 
and $d$ is the dimension of compound feature embeddings. 
Note that in Equation~\ref{eqn:localloss}, $p^{(S)}_i$ and $p^{(T)}_i$ both measure an embedding feature's probability of being specific 
to the \emph{source domain}; superscripts $^{(S)}$/$^{(T)}$ here indicate that the compounds, 
whose features are measured, are from source/target domain, respectively.
%

%
To have an accurate feature-wise discriminator, embedding features specific to the source/target domain should have large/small probabilities 
(i.e., large $p^{(S)}_i$ and small $p^{(T)}_i$) 
with respect to the source domain, and thus make the \LocalDiscLoss value small. 
Therefore, minimizing \LocalDiscLoss will encourage accurate probabilities.
Meanwhile, the feature learner \fgen should encourage the learning of more transferable embedding features, which will have probabilities close to 0.5
and thus  make the \LocalDiscLoss value large. Therefore, maximizing \LocalDiscLoss will encourage more transferable embedding features
being learned and learned well. 
To combine these two aspects, an adversarial optimization will be applied to \LocalDiscLoss as will be described later in Section~\ref{sec:methods:variations:full_optimization}. 

%
%

%

\paragraph{Learning Transferability of Compounds}
\label{sec:methods:variations:global_disc}
\hfill \\
%
{Inspired by the principle that similar compounds tend to bind to similar protein targets,
our method identifies such similar compounds that have
same activity labels across two targets, and hence learns compound-wise transferability.
}
Given the scaled compound embedding \scompr of compound \compound,
the compound-wise discriminator classifies whether the compound
is from the source domain using a two-layer fully-connected neural network \gdisc as follows,
\begin{equation}
\label{eqn:global}
q = \gdisc(\scompr), 
\end{equation}
with ReLU at the hidden layer and sigmoid at the output layer. 
%
If $q$ is very large or very small, \compound  is very likely or 
very unlikely to belong to the source domain (it is equivalent to calculate the value with respect to the target domain, since 
there are only two domains to consider).
If $q$ is close to 0.5, \compound is likely to be common 
across domains (e.g., identical or similar compounds in the two domains)  and thus can 
be used for information transfer across domains.
%
%

%
To learn the compound-wise discriminator, the loss function \GlobalDiscLoss 
is defined as follows,
%
%
\begin{eqnarray}
\label{eqn:globaldisc}
\begin{aligned}
\GlobalDiscLoss(\ParGnn, \ParGdisc)
 =& - \frac{1}{n^{(S)}}
			\sum_{\scriptsize{\InputFeature^{(S)}} \in \scriptsize{\SetCompounds^{(S)}}}
			\log(q^{(S)}) \\
		&	- \frac{1}{n^{(T)}}
			\sum_{\scriptsize{\InputFeature^{(T)}} \in \scriptsize{\SetCompounds^{(T)}}}
			\log(1 - q^{(T)}), 
\end{aligned}
\end{eqnarray}
where $n^{(S)}$/$n^{(T)}$ is the number of compounds
in $\SetCompounds^{(S)}$/$\SetCompounds^{(T)}$ (i.e., $n = |\domain|$); 
\ParGnn and \ParGdisc are learnable parameters of
 \fgen (compound representation learning network as in Section~\ref{sec:methods:conv}) 
and  \gdisc (Equation~\ref{eqn:global}); 
and $d$ is the dimension of compound feature embeddings.
Note that in Equation~\ref{eqn:globaldisc}, 
$q^{(S)}$ and $q^{(T)}$ represent the probability of $\compound^{(S)}$ and $\compound^{(T)}$ belonging to the \emph{source domain}. 
Also, all the compounds from the source and target domains will be predicted using a same {\gdisc}. 
%

%
In order to identify similar compounds across domains, the discriminator needs to identify compounds with their $q$ values close to 0.5; 
when the $q$ values are close 0.5, {\GlobalDiscLoss} will be maximized. Therefore, maximizing \GlobalDiscLoss will encourage more 
transferable compounds being learned and learned well.   
Meanwhile, to have an accurate compound-wise discriminator, compounds specific to the source/target domain should have large/small 
probabilities (i.e., large $q^{(S)}_i$ and small $q^{(T)}_i$) with respect to the source domain, and thus make the \GlobalDiscLoss value small. 
Therefore, minimizing \GlobalDiscLoss will encourage accurate probabilities.
To combine these two aspects, similarly as to \LocalDiscLoss, 
an adversarial optimization will be applied to \GlobalDiscLoss as will be described later in Section~\ref{sec:methods:variations:full_optimization}. 

%
According to \gdisc, a compound that is common in the two domains, or is similar to compounds in the other domain, 
could be transferable ($q$ value close to 0.5; not specific to source or target domain). 
However, such common or similar compounds may have different activity labels in the two domains. 
Using transfered information 
from common/similar compounds with conflicting labels in \TargetTask will confuse any learners adversely. 
The compound-wise discriminator \gdisc does not consider activity label information in learning compound transferability, and thus 
possibly induces conflicting information into \TargetTask. 
However, in the downstream domain-wise classification (Section~\ref{sec:methods:classifier}), the minimization of domain-specific 
classification errors will prevent the transfer of conflicting information. 

However, the input to the domain-wise classifier \classifier in \OurMethodWithDisc
is \scompr instead of \compr as in Section~\ref{sec:methods:classifier}.
Given the scaled compound embedding \scompr,
the domain-wise classifier classifies each compound in a given domain
as active or inactive with respect to that domain
using a two-layer fully-connected neural network \classifier
as follows,
\begin{equation}
\label{equ:classifier_fullmodel}
\hat{y} = \classifier(\scompr), 
\end{equation}
with ReLU at the hidden layer and sigmoid at the output layer.
As discussed in Section~\ref{sec:methods:classifier}, 
minimizing the loss {$\ClassifierLoss(\ParGnn, \ParClassifier)$}
enables correct classification in each domain,
and prevent the transfer of conflicting information across domains.

\paragraph{\OurMethodWithDisc Model Optimization}
\label{sec:methods:variations:full_optimization}
\hfill \\
{This section presents the optimization process
of the proposed \OurMethodWithDisc.
}
{\OurMethodWithDisc} constructs an end-to-end adversarial transfer learning framework with the 
above four components: 1) compound feature presentation learning network \fgen, 
2) feature-wise discriminator \ldisc, 3) compound-wise discriminator \gdisc, and 4) domain-wise classifier
\classifier. We solve for {\OurMethodWithDisc} through optimizing the following loss function:
\begin{eqnarray}
\label{equ:loss}
\begin{aligned}
 \FinalLoss(\ParGnn, \ParLdisc, \ParGdisc, \ParClassifier) =
 		-\lambda [( \LocalDiscLoss(\ParGnn, \ParLdisc) + \GlobalDiscLoss(\ParGnn, \ParGdisc)] + {\ClassifierLoss(\ParGnn,\ParClassifier)}, 
\end{aligned}
\end{eqnarray}
where \ParGnn, \ParLdisc, \ParGdisc and \ParClassifier are learnable
parameters of \fgen, \ldisc, \gdisc and \classifier, respectively, 
and $\lambda$ is a trade-off parameter. 
This loss function combines the three loss functions for \ldisc, \gdisc and \classifier, and will be optimized in an 
adversarial way as follows:

(Step 1). Minimize \FinalLoss with respect to \ParGnn and \ParClassifier via solving the following optimization problem:
\begin{equation}
\label{eqn:min}
\widehat{\ParGnn}, \widehat{\ParClassifier} = 
\arg \min_{\scriptsize{\ParGnn, \ParClassifier}} \FinalLoss(\ParGnn, \ParLdisc, \ParGdisc, \ParClassifier). 
\end{equation}
By minimizing \FinalLoss, we essentially minimize \ClassifierLoss, and maximize 
\LocalDiscLoss and \GlobalDiscLoss. As discussed in \ldisc (Section~\ref{sec:methods:variations:local_disc}) and 
\gdisc (Section~\ref{sec:methods:variations:global_disc}), maximizing \LocalDiscLoss and \GlobalDiscLoss will encourage 
learning of transferable features and compounds that can be used to help the tasks; as discussed in \classifier
(Section~\ref{sec:methods:classifier}), minimizing \ClassifierLoss will prevent the transfer of conflicting information, in addition to 
minimizing the classification errors in each task. 

(Step 2). Maximize \FinalLoss with respect to \ParLdisc and \ParGdisc via solving the following optimization problem:
\begin{equation}
\label{eqn:max}
\widehat{\ParLdisc}, \widehat{\ParGdisc} = 
\arg \max_{\scriptsize{\ParLdisc, \ParGdisc}} \FinalLoss(\ParGnn, \ParLdisc, \ParGdisc, \ParClassifier). 
\end{equation}
By maximizing \FinalLoss, we essentially minimize \LocalDiscLoss and \GlobalDiscLoss (\ClassifierLoss is fixed in this step). 
As discussed in \ldisc (Section~\ref{sec:methods:variations:local_disc}) and 
\gdisc (Section~\ref{sec:methods:variations:global_disc}), minimizing \LocalDiscLoss and \GlobalDiscLoss will encourage that
\ldisc and \gdisc accurately learn features and compounds that are specific to each domain so as to improve 
the classification performance of each domain. 

(Step 3). The above two steps are iterated until the learning converges. 
%
%
%

Thus, the optimization problem consists of a maximization with respect to 
some parameters and a minimization with respect to the others.
In order to tackle such a mini-max optimization,
we insert the gradient reversal layer (GRL)~\cite{Ganin2015}
between \fgen and the discriminators \ldisc and \gdisc.
GRL reverses the gradients during the backward propagation,
and hence optimizes parameters \ParGnn by 
maximizing the discriminator loss.
%


\section{Acknowledgment}
This project was made possible, in part, by support from the
National Science Foundation under Grant Number IIS-1855501, IIS-1827472 and IIS-2133650, and AWS
Machine Learning Research Award. Any opinions, findings,
and conclusions or recommendations expressed in this
material are those of the authors and do not necessarily reflect
the views of the funding agencies.

\bibliography{ref_abbv}  

\end{document}


\section{Directed Message Passing Neural Networks}
\label{suppsec:dmpnn}

\dmpnn incorporates atom features $\AtomFeature_u$ 
for each atom $u \in \SetAtoms_{\scriptsize{\compound}}$
and bond features $\BondFeature_{uv}$ 
for each bond $(u, v) \in \SetBonds_{\scriptsize{\compound}}$, and
captures molecular substructures by propagating messages 
along directed edges in $\MolGraph_{\scriptsize{\compound}}$.
\dmpnn initializes atom features $\AtomFeature_u$ using the atom's
physicochemical properties including mass, formal charge, chirality, type, 
number of connected bonds, etc.  
%
\dmpnn initializes bond features $\BondFeature_{uv}$ using
bond type, stereo configuration, etc.  
%
 
%
In particular, in $\MolGraph_{\scriptsize{\compound}}$,
each bond $(u, v)$ is associated with two messages $\mes_{uv}$ and $\mes_{vu}$
encoding messages from atom $u$ to $v$ and vice versa.
%
%
Each message $\mes_{uv}^{(t+1)}$ in the $(t+1)$-th iteration of \dmpnn 
is aggregated as follows,
\begin{equation}
\label{equ:message_passing}
\mes_{uv}^{(t+1)} = \sum_{k \in \mathcal{N}(u) \setminus v} \hidden_{ku}^{(t)}
\end{equation}
where $\mathcal{N}(u)$ is the set of atoms connected to $u$,
$\hidden_{ku}^{(t)}$ is the hidden state of edge $(k, u)$ in the $t$-th iteration. 
%
%
In the $(t+1)$-th iteration of message passing, the hidden state $\hidden_{uv}^{(t+1)}$ 
for each edge $(u, v)$ is updated as follows,
\begin{equation}
\label{equ:update}
\hidden_{uv}^{(t+1)} = \text{ReLU}(\hidden_{uv}^{(0)} + W \mes_{uv}^{(t+1)}), 
\end{equation}
where $W$ is a learnable parameter matrix, and 
$\hidden_{uv}^{(0)}$ is the initial hidden state of edge $(u,v)$ initialized as follows,
\begin{equation}
\label{equ:initial}
\hidden_{uv}^{(0)} = \text{ReLU}(W_0 [\AtomFeature_u, \BondFeature_{uv}])
\end{equation}
where $\AtomFeature_u$ and $\BondFeature_{uv}$ are atom and 
bond feature vectors, 
respectively, $[\AtomFeature_u, \BondFeature_{uv}]$ is the concatenation of 
$\AtomFeature_u$ and $\BondFeature_{uv}$ and $W_0$ is a learnable parameter. 

After the final iteration of message passing,
the hidden states for edges incident to an atom $u$ are 
aggregated to generate an intermediate representation $\hidden_u$ for that atom as follows, 
\begin{equation}
\label{eqn:hiddenu}
\hidden_u = \sum_{k \in \mathcal{N}(u)} \hidden_{ku}^{(\tau)}, 
\end{equation}
where $\tau$ is the total number of message passing iterations. These intermediate atom representations 
are then used to generate another atom representation that also incorporates atom features $\AtomFeature_u$ as follows, 
%
%
\begin{equation}
\label{eqn:embeddingm}
\embedding_u = \text{ReLU}(W_e [\AtomFeature_u, \hidden_u]),
\end{equation}
where $W_e$ 
is a learnable parameter and $[\AtomFeature_u, \hidden_u]$ is the concatenation of
$\AtomFeature_u$ and $\hidden_u$.
Thus, the atom representation $\hidden_u$ captures structural information
about atom $u$'s $\tau$-hop neighbors, thereby enhancing the representation power. 
Given the representation $\hidden_u$ for each atom in \compound, \dmpnn 
produces an embedding for \compound using mean pooling over all the atom representations as follows, 
%
\begin{equation}
\label{equ:mean-pooling}
\compr_{\scriptsize{\compound}} = \frac{1}{|\SetAtoms_{\scriptsize{\compound}}|} 
\sum_{u \in \scriptsize{\SetAtoms_{\scriptsize{\compound}}}} \embedding_u, 
\end{equation}
where $|\SetAtoms_{\scriptsize{\compound}}|$ is the number of atoms in \compound.

\section{Supplementary Materials}

\subsection{Assay Information}
\label{suppsec:assay}
Table~\ref{tbl:supp:final_bio_stat} presents the assay statistics of 93 processed bioassays
with their associated target protein accession IDs in PubChem; their
number of total, active and inactive compounds, 
and the corresponding protein family for each associated target.

\begin{ThreePartTable}
\centering
\begin{TableNotes}
\item In this table, the columns protacxn, \#total, \#active and \#inactive correspond to the target protein's accession number from PubChem, number of total compounds, active compounds and inactive compounds in the bioassay, respectively.
The column Protein family denotes the corresponding protein family for each target.
\par
\end{TableNotes}

\begin{longtable}[h]
		{
		@{\hspace{5pt}}c@{\hspace{5pt}}
		@{\hspace{2pt}}r@{\hspace{2pt}}
		@{\hspace{2pt}}r@{\hspace{2pt}}
		@{\hspace{2pt}}r@{\hspace{2pt}}
		@{\hspace{10pt}}c@{\hspace{10pt}}
		}
		
		\caption{Compound statistics of processed bioassays}
		\label{tbl:supp:final_bio_stat} \\
		\midrule
        	protacxn & \#total & \# active &  \# inactive & Protein family\\ 
        \midrule
        \endfirsthead

        \multicolumn{5}{c}%
        {{\bfseries \tablename\ \thetable{} -- continued from previous page}} \\
	\midrule
        	protacxn & \#total & \# active &  \# inactive  & Protein family\\ 
        \midrule
        \endhead
        \midrule
        \multicolumn{5}{|r|}{{Continued on next page}} \\
        \endfoot
        \midrule
        \insertTableNotes
        \endlastfoot
        NP\_001017535 & 380,711 & 5,152 & 375,559 & Nuclear hormone receptor family \\
AAH18745 & 374,923 & 1,230 & 373,693 & Peptidase family \\
NP\_005021 & 349,900 & 10,198 & 339,702 & Protein kinase superfamily \\
NP\_004196 & 336,578 & 238 & 336,340 & Peptidase family \\
NP\_057051 & 292,415 & 97 & 292,318 & Protein-tyrosine phosphatase family \\
NP\_000903 & 292,382 & 151 & 292,231 & G-protein coupled receptor 1 family \\
NP\_005292 & 292,044 & 207 & 291,837 & G-protein coupled receptor 1 family \\
NP\_004081 & 290,998 & 1,674 & 289,324 & Protein-tyrosine phosphatase family \\
NP\_001121649 & 275,994 & 116 & 275,878 & Nuclear hormone receptor family \\
NP\_997055 & 208,843 & 2,526 & 206,317 & G-protein coupled receptor 1 family \\
NP\_002736 & 182,123 & 1,207 & 180,916 & Protein kinase superfamily \\
NP\_542155 & 112,687 & 906 & 111,781 & Protein-tyrosine phosphatase family \\
NP\_775180 & 96,456 & 153 & 96,303 & Nuclear hormone receptor family \\
P55210 & 71,790 & 77 & 71,713 & Peptidase family \\
NP\_150634 & 70,986 & 61 & 70,925 & Peptidase family \\
NP\_000483 & 10,100 & 179 & 9,921 & ABC transporter superfamily \\
ABB72139 & 6,968 & 64 & 6,904 & Nuclear hormone receptor family \\
ADZ17337 & 6,738 & 86 & 6,652 & Nuclear hormone receptor family \\
NP\_000762 & 7,759 & 1,163 & 6,596 & Cytochrome P450 family \\
EAW77416 & 7,131 & 626 & 6,505 & Cytochrome P450 family \\
ADZ17384 & 6,425 & 55 & 6,370 & Nuclear hormone receptor family \\
NP\_000760 & 7,676 & 1,713 & 5,963 & Cytochrome P450 family \\
NP\_000752 & 7,670 & 4,008 & 3,662 & Cytochrome P450 family \\
AAF64255 & 2,728 & 1,107 & 1,621 & Bcl-2 family \\
NP\_002084 & 2,285 & 682 & 1,603 & Protein kinase superfamily \\
NP\_063937 & 2,390 & 823 & 1,567 & Protein kinase superfamily \\
AAI28575 & 1,824 & 305 & 1,519 & Nuclear hormone receptor family \\
AAB26273 & 1,669 & 276 & 1,393 & G-protein coupled receptor 1 family \\
NP\_000947 & 1,350 & 139 & 1,211 & G-protein coupled receptor 1 family \\
NP\_036559 & 1,260 & 109 & 1,151 & Peptidase family \\
NP\_065717 & 1,630 & 661 & 969 & Protein kinase superfamily \\
AAI27629 & 1,065 & 160 & 905 & G-protein coupled receptor 1 family \\
P51449 & 978 & 138 & 840 & Nuclear hormone receptor family \\
NP\_004040 & 824 & 91 & 733 & Bcl-2 family \\
P00748 & 761 & 160 & 601 & Peptidase family \\
ABD72211 & 1,164 & 605 & 559 & ABC transporter superfamily \\
AAH04460 & 993 & 445 & 548 & Peptidase family \\
NP\_004950 & 954 & 456 & 498 & Nuclear hormone receptor family \\
NP\_000676 & 667 & 209 & 458 & G-protein coupled receptor 1 family \\
NP\_000466 & 489 & 55 & 434 & Nuclear hormone receptor family \\
NP\_005152 & 672 & 259 & 413 & G-protein coupled receptor 1 family \\
NP\_001391 & 729 & 345 & 384 & G-protein coupled receptor 1 family \\
NP\_005217 & 456 & 82 & 374 & G-protein coupled receptor 1 family \\
NP\_660205 & 1,167 & 795 & 372 & Peptidase family \\
NP\_000789 & 457 & 100 & 357 & G-protein coupled receptor 1 family \\
NP\_001027450 & 543 & 203 & 340 & Peptidase family \\
NP\_004521 & 445 & 107 & 338 & Peptidase family \\
NP\_004960 & 387 & 61 & 326 & Peptidase family \\
NP\_005424 & 693 & 368 & 325 & Protein kinase superfamily \\
P53779 & 362 & 57 & 305 & Protein kinase superfamily \\
AAC63054 & 517 & 223 & 294 & Peptidase family \\
AAF04852 & 973 & 683 & 290 & Peptidase family \\
NP\_004337 & 1,055 & 781 & 274 & Peptidase family \\
AAA51985 & 366 & 105 & 261 & Peptidase family \\
NP\_004358 & 289 & 58 & 231 & G-protein coupled receptor 1 family \\
NP\_004221 & 276 & 69 & 207 & G-protein coupled receptor 1 family \\
NP\_000901 & 292 & 88 & 204 & G-protein coupled receptor 1 family \\
NP\_002522 & 349 & 156 & 193 & G-protein coupled receptor 1 family \\
NP\_644806 & 248 & 57 & 191 & Peptidase family \\
EAW86722 & 539 & 350 & 189 & G-protein coupled receptor 1 family \\
NP\_000900 & 291 & 104 & 187 & G-protein coupled receptor 1 family \\
AAI07736 & 242 & 59 & 183 & Bcl-2 family \\
AAI29989 & 981 & 799 & 182 & Peptidase family \\
BAH02301 & 462 & 317 & 145 & Nuclear hormone receptor family \\
NP\_000448 & 292 & 149 & 143 & Nuclear hormone receptor family \\
NP\_004841 & 229 & 88 & 141 & Protein kinase superfamily \\
AAH14970 & 248 & 126 & 122 & G-protein coupled receptor 1 family \\
NP\_002721 & 190 & 71 & 119 & Protein kinase superfamily \\
NP\_003813 & 211 & 92 & 119 & Nuclear hormone receptor family \\
AAH36651 & 210 & 92 & 118 & Protein kinase superfamily \\
Q05397 & 210 & 110 & 100 & Protein kinase superfamily \\
NP\_004570 & 151 & 52 & 99 & Protein kinase superfamily \\
NP\_000918 & 189 & 102 & 87 & ABC transporter superfamily \\
NP\_112168 & 304 & 217 & 87 & Protein kinase superfamily \\
NP\_066285 & 233 & 173 & 60 & Nuclear hormone receptor family \\
BAB91222 & 142 & 91 & 51 & G-protein coupled receptor 1 family \\
NP\_003605 & 232 & 181 & 51 & G-protein coupled receptor 1 family \\
P28566 & 95 & 51 & 44 & G-protein coupled receptor 1 family \\
NP\_037457 & 331 & 290 & 41 & Peptidase family \\
Q6L5J4 & 432 & 391 & 41 & G-protein coupled receptor 1 family \\
NP\_000732 & 98 & 58 & 40 & G-protein coupled receptor 1 family \\
NP\_612200 & 314 & 274 & 40 & G-protein coupled receptor 1 family \\
NP\_001387 & 307 & 270 & 37 & Protein kinase superfamily \\
NP\_004818 & 125 & 89 & 36 & ABC transporter superfamily \\
EAW70217 & 188 & 154 & 34 & Protein kinase superfamily \\
AAH14460 & 91 & 69 & 22 & Peptidase family \\
NP\_001124480 & 81 & 64 & 17 & Protein-tyrosine phosphatase family \\
NP\_001516 & 171 & 156 & 15 & G-protein coupled receptor 1 family \\
NP\_003984 & 95 & 83 & 12 & Protein kinase superfamily \\
NP\_004705 & 94 & 82 & 12 & Protein kinase superfamily \\
NP\_004062 & 96 & 88 & 8 & Protein kinase superfamily \\
NP\_002825 & 166 & 160 & 6 & Protein-tyrosine phosphatase family \\
NP\_004942 & 74 & 69 & 5 & G-protein coupled receptor 1 family
\end{longtable}
\end{ThreePartTable}


\subsection{Assay Pairs Information}
\label{suppsec:assay_pairs}

Table~\ref{tbl:supp:pairs_stat} presents the compound statistics of all 120 pairs in $\mathcal{P}$
with their number of total, 
active and inactive compounds in each assay of every pair.

\begin{ThreePartTable}
\centering
\begin{TableNotes}
\item In this table, the first two columns target(P) and target(Q) 
correspond to the accession numbers of the two target proteins 
corresponding to the two bioassays P and Q of the pair, respectively.
The columns 
$|\SetCompounds_{\scriptsize \bioassay_A}|$ 
$|\SetCompounds_{\scriptsize \bioassay_A}^{+}|$ 
and $|\SetCompounds_{\scriptsize \bioassay_A}^{-}|$ 
correspond to the number of
total compounds, active compounds and inactive compounds in the bioassay $A$  
where ($A = \{P, Q\}$ of each pair).
\par
\end{TableNotes}

\begin{longtable}[h]
	{
      	  @{\hspace{8pt}}r@{\hspace{8pt}}
          @{\hspace{8pt}}r@{\hspace{8pt}}
          @{\hspace{8pt}}r@{\hspace{8pt}}
          @{\hspace{8pt}}r@{\hspace{8pt}}
          @{\hspace{8pt}}r@{\hspace{8pt}}
          @{\hspace{8pt}}r@{\hspace{8pt}}
          @{\hspace{8pt}}r@{\hspace{8pt}}
          @{\hspace{8pt}}r@{\hspace{8pt}}
        }

	\caption{Compound statistics of selected bioassay pairs}
	\label{tbl:supp:pairs_stat} \\
	\midrule
        	target(P) & target(Q) & $|\SetCompounds_{\scriptsize \bioassay_P}|$
        	& $|\SetCompounds_{\scriptsize \bioassay_P}^{+}|$ 
        	& $|\SetCompounds_{\scriptsize \bioassay_P}^{-}|$
        & $|\SetCompounds_{\scriptsize \bioassay_Q}|$
        	& $|\SetCompounds_{\scriptsize \bioassay_Q}^{+}|$ 
        	& $|\SetCompounds_{\scriptsize \bioassay_Q}^{-}|$
        \\ 
        \midrule
        \endfirsthead

        \multicolumn{8}{c}%
        {{\bfseries \tablename\ \thetable{} -- continued from previous page}} \\
	\midrule
        	target(P) & target(Q) & $|\SetCompounds_{\scriptsize \bioassay_P}|$
        	& $|\SetCompounds_{\scriptsize \bioassay_P}^{+}|$ 
        	& $|\SetCompounds_{\scriptsize \bioassay_P}^{-}|$
        & $|\SetCompounds_{\scriptsize \bioassay_Q}|$
        	& $|\SetCompounds_{\scriptsize \bioassay_Q}^{+}|$ 
        	& $|\SetCompounds_{\scriptsize \bioassay_Q}^{-}|$
        	\\
        \midrule
        \endhead
        \midrule
        \multicolumn{8}{|r|}{{Continued on next page}} \\
        \endfoot
        \midrule
        \insertTableNotes
        \endlastfoot        
        
NP\_005021 & NP\_004705 & 20,396 & 10,198 & 10,198 & 164 & 82 & 82 \\
NP\_000903 & EAW86722 & 302 & 151 & 151 & 666 & 333 & 333 \\
NP\_000903 & BAB91222 & 302 & 151 & 151 & 166 & 83 & 83 \\
NP\_005292 & BAB91222 & 414 & 207 & 207 & 166 & 83 & 83 \\
NP\_005292 & NP\_004942 & 414 & 207 & 207 & 134 & 67 & 67 \\
NP\_004081 & NP\_002825 & 3,288 & 1,644 & 1,644 & 248 & 124 & 124 \\
NP\_997055 & NP\_004942 & 5,052 & 2,526 & 2,526 & 134 & 67 & 67 \\
NP\_775180 & ADZ17337 & 306 & 153 & 153 & 170 & 85 & 85 \\
NP\_000483 & ABD72211 & 356 & 178 & 178 & 1,188 & 594 & 594 \\
NP\_000483 & NP\_000918 & 358 & 179 & 179 & 204 & 102 & 102 \\
NP\_000483 & NP\_004818 & 358 & 179 & 179 & 178 & 89 & 89 \\
ADZ17337 & AAI28575 & 156 & 78 & 78 & 592 & 296 & 296 \\
ADZ17337 & NP\_000448 & 172 & 86 & 86 & 298 & 149 & 149 \\
ADZ17337 & NP\_066285 & 172 & 86 & 86 & 342 & 171 & 171 \\
NP\_000762 & NP\_000752 & 888 & 444 & 444 & 1,918 & 959 & 959 \\
NP\_000760 & NP\_000752 & 1,356 & 678 & 678 & 2,070 & 1,035 & 1,035 \\
NP\_002084 & P53779 & 1,358 & 679 & 679 & 112 & 56 & 56 \\
AAI28575 & P51449 & 608 & 304 & 304 & 244 & 122 & 122 \\
AAI28575 & BAH02301 & 608 & 304 & 304 & 626 & 313 & 313 \\
AAI28575 & NP\_000448 & 608 & 304 & 304 & 298 & 149 & 149 \\
AAI28575 & NP\_066285 & 610 & 305 & 305 & 346 & 173 & 173 \\
AAB26273 & NP\_000947 & 536 & 268 & 268 & 272 & 136 & 136 \\
AAB26273 & AAI27629 & 552 & 276 & 276 & 320 & 160 & 160 \\
AAB26273 & NP\_000676 & 552 & 276 & 276 & 416 & 208 & 208 \\
AAB26273 & NP\_001391 & 548 & 274 & 274 & 688 & 344 & 344 \\
AAB26273 & NP\_004358 & 552 & 276 & 276 & 116 & 58 & 58 \\
AAB26273 & NP\_000901 & 552 & 276 & 276 & 174 & 87 & 87 \\
AAB26273 & EAW86722 & 552 & 276 & 276 & 700 & 350 & 350 \\
AAB26273 & NP\_000900 & 552 & 276 & 276 & 206 & 103 & 103 \\
AAB26273 & AAH14970 & 552 & 276 & 276 & 252 & 126 & 126 \\
AAB26273 & BAB91222 & 552 & 276 & 276 & 182 & 91 & 91 \\
AAB26273 & NP\_003605 & 552 & 276 & 276 & 360 & 180 & 180 \\
AAB26273 & NP\_612200 & 552 & 276 & 276 & 542 & 271 & 271 \\
AAB26273 & NP\_001516 & 552 & 276 & 276 & 310 & 155 & 155 \\
NP\_000947 & AAI27629 & 278 & 139 & 139 & 318 & 159 & 159 \\
NP\_000947 & EAW86722 & 278 & 139 & 139 & 700 & 350 & 350 \\
NP\_000947 & BAB91222 & 278 & 139 & 139 & 182 & 91 & 91 \\
NP\_000947 & NP\_001516 & 278 & 139 & 139 & 308 & 154 & 154 \\
NP\_036559 & P00748 & 168 & 84 & 84 & 274 & 137 & 137 \\
NP\_036559 & AAC63054 & 184 & 92 & 92 & 410 & 205 & 205 \\
NP\_036559 & AAF04852 & 218 & 109 & 109 & 1,362 & 681 & 681 \\
NP\_036559 & NP\_004337 & 218 & 109 & 109 & 1,556 & 778 & 778 \\
NP\_036559 & AAA51985 & 174 & 87 & 87 & 180 & 90 & 90 \\
NP\_036559 & AAI29989 & 218 & 109 & 109 & 1,592 & 796 & 796 \\
NP\_036559 & AAH14460 & 202 & 101 & 101 & 120 & 60 & 60 \\
NP\_065717 & NP\_001387 & 1,028 & 514 & 514 & 278 & 139 & 139 \\
NP\_065717 & EAW70217 & 1,268 & 634 & 634 & 260 & 130 & 130 \\
AAI27629 & EAW86722 & 320 & 160 & 160 & 698 & 349 & 349 \\
AAI27629 & NP\_004942 & 320 & 160 & 160 & 138 & 69 & 69 \\
P00748 & NP\_660205 & 320 & 160 & 160 & 1,586 & 793 & 793 \\
P00748 & NP\_004521 & 320 & 160 & 160 & 214 & 107 & 107 \\
P00748 & AAC63054 & 266 & 133 & 133 & 382 & 191 & 191 \\
P00748 & AAF04852 & 320 & 160 & 160 & 1,364 & 682 & 682 \\
P00748 & NP\_004337 & 320 & 160 & 160 & 1,558 & 779 & 779 \\
P00748 & AAA51985 & 244 & 122 & 122 & 154 & 77 & 77 \\
P00748 & AAI29989 & 320 & 160 & 160 & 1,596 & 798 & 798 \\
P00748 & AAH14460 & 296 & 148 & 148 & 114 & 57 & 57 \\
ABD72211 & NP\_000918 & 1,210 & 605 & 605 & 202 & 101 & 101 \\
ABD72211 & NP\_004818 & 1,210 & 605 & 605 & 178 & 89 & 89 \\
AAH04460 & AAH14460 & 890 & 445 & 445 & 138 & 69 & 69 \\
NP\_000676 & NP\_005152 & 154 & 77 & 77 & 172 & 86 & 86 \\
NP\_000676 & NP\_000901 & 416 & 208 & 208 & 176 & 88 & 88 \\
NP\_000676 & EAW86722 & 418 & 209 & 209 & 700 & 350 & 350 \\
NP\_000676 & NP\_001516 & 418 & 209 & 209 & 310 & 155 & 155 \\
NP\_000676 & NP\_004942 & 418 & 209 & 209 & 138 & 69 & 69 \\
NP\_005152 & NP\_004358 & 512 & 256 & 256 & 108 & 54 & 54 \\
NP\_005152 & NP\_001516 & 518 & 259 & 259 & 310 & 155 & 155 \\
NP\_001391 & NP\_000901 & 682 & 341 & 341 & 174 & 87 & 87 \\
NP\_001391 & EAW86722 & 690 & 345 & 345 & 700 & 350 & 350 \\
NP\_001391 & NP\_000900 & 682 & 341 & 341 & 206 & 103 & 103 \\
NP\_001391 & BAB91222 & 690 & 345 & 345 & 182 & 91 & 91 \\
NP\_001391 & NP\_001516 & 690 & 345 & 345 & 308 & 154 & 154 \\
NP\_660205 & NP\_004337 & 914 & 457 & 457 & 642 & 321 & 321 \\
NP\_660205 & AAI29989 & 986 & 493 & 493 & 642 & 321 & 321 \\
NP\_005424 & NP\_001387 & 736 & 368 & 368 & 540 & 270 & 270 \\
NP\_005424 & NP\_003984 & 736 & 368 & 368 & 166 & 83 & 83 \\
NP\_005424 & NP\_004705 & 736 & 368 & 368 & 164 & 82 & 82 \\
NP\_005424 & NP\_004062 & 736 & 368 & 368 & 176 & 88 & 88 \\
AAC63054 & AAF04852 & 446 & 223 & 223 & 1,364 & 682 & 682 \\
AAC63054 & NP\_004337 & 446 & 223 & 223 & 1,558 & 779 & 779 \\
AAC63054 & AAA51985 & 402 & 201 & 201 & 182 & 91 & 91 \\
AAC63054 & AAI29989 & 446 & 223 & 223 & 1,594 & 797 & 797 \\
AAC63054 & AAH14460 & 438 & 219 & 219 & 130 & 65 & 65 \\
AAF04852 & NP\_004337 & 654 & 327 & 327 & 718 & 359 & 359 \\
AAF04852 & AAI29989 & 682 & 341 & 341 & 696 & 348 & 348 \\
NP\_004337 & AAI29989 & 804 & 402 & 402 & 758 & 379 & 379 \\
AAA51985 & AAI29989 & 210 & 105 & 105 & 1,598 & 799 & 799 \\
AAA51985 & AAH14460 & 194 & 97 & 97 & 118 & 59 & 59 \\
NP\_004358 & NP\_000901 & 116 & 58 & 58 & 176 & 88 & 88 \\
NP\_004358 & EAW86722 & 116 & 58 & 58 & 700 & 350 & 350 \\
NP\_004358 & NP\_000900 & 116 & 58 & 58 & 208 & 104 & 104 \\
NP\_004358 & AAH14970 & 116 & 58 & 58 & 244 & 122 & 122 \\
NP\_004358 & BAB91222 & 116 & 58 & 58 & 182 & 91 & 91 \\
NP\_004358 & NP\_001516 & 116 & 58 & 58 & 312 & 156 & 156 \\
NP\_004221 & EAW86722 & 138 & 69 & 69 & 700 & 350 & 350 \\
NP\_004221 & NP\_001516 & 138 & 69 & 69 & 312 & 156 & 156 \\
NP\_004221 & NP\_004942 & 138 & 69 & 69 & 138 & 69 & 69 \\
NP\_000901 & EAW86722 & 174 & 87 & 87 & 700 & 350 & 350 \\
NP\_000901 & BAB91222 & 176 & 88 & 88 & 182 & 91 & 91 \\
NP\_000901 & NP\_001516 & 176 & 88 & 88 & 312 & 156 & 156 \\
EAW86722 & AAH14970 & 700 & 350 & 350 & 250 & 125 & 125 \\
EAW86722 & BAB91222 & 700 & 350 & 350 & 182 & 91 & 91 \\
EAW86722 & NP\_001516 & 700 & 350 & 350 & 312 & 156 & 156 \\
NP\_000900 & BAB91222 & 208 & 104 & 104 & 182 & 91 & 91 \\
NP\_000900 & NP\_001516 & 208 & 104 & 104 & 312 & 156 & 156 \\
NP\_000900 & NP\_004942 & 208 & 104 & 104 & 138 & 69 & 69 \\
AAH14970 & BAB91222 & 252 & 126 & 126 & 182 & 91 & 91 \\
AAH14970 & NP\_001516 & 252 & 126 & 126 & 310 & 155 & 155 \\
AAH14970 & NP\_004942 & 252 & 126 & 126 & 138 & 69 & 69 \\
NP\_002721 & NP\_001387 & 142 & 71 & 71 & 540 & 270 & 270 \\
NP\_002721 & NP\_003984 & 142 & 71 & 71 & 166 & 83 & 83 \\
NP\_002721 & NP\_004705 & 142 & 71 & 71 & 164 & 82 & 82 \\
NP\_002721 & NP\_004062 & 142 & 71 & 71 & 176 & 88 & 88 \\
BAB91222 & NP\_004942 & 182 & 91 & 91 & 138 & 69 & 69 \\
NP\_003605 & NP\_004942 & 362 & 181 & 181 & 138 & 69 & 69 \\
NP\_001387 & EAW70217 & 488 & 244 & 244 & 260 & 130 & 130 \\
EAW70217 & NP\_003984 & 278 & 139 & 139 & 136 & 68 & 68 \\
EAW70217 & NP\_004705 & 280 & 140 & 140 & 132 & 66 & 66 \\
EAW70217 & NP\_004062 & 280 & 140 & 140 & 142 & 71 & 71 \\
NP\_001124480 & NP\_002825 & 128 & 64 & 64 & 318 & 159 & 159
\end{longtable}
\end{ThreePartTable}

\section{Supplementary Results}
\subsection{Hyperparmeter Configurations}
\label{suppsec:hyper}

\begin{table}[h]
\centering
\caption{Hyperparameter Configurations}
\label{tbl:supp:parameters}
  \begin{threeparttable}
      \begin{tabular}{
          @{\hspace{5pt}}l@{\hspace{5pt}}
          @{\hspace{5pt}}c@{\hspace{5pt}}
          }
          \toprule
          parameter & values \\
          \midrule
          $\alpha$ & {0, 0.1, 0.5, 1, 2} \\
          $\lambda$ & {0.001, 0.01, 0.5, 1} \\
          dimension $d$ & {25, 50, 100} \\
          \# message passing steps $\tau$   &    {2, 3, 4} \\
          hidden layer size in $\fatt(\cdot)$	& 		{100} \\
		 hidden layer size in $\ldisc$, $\gdisc$ and $\classifier$  &   {100} \\
          batch size   	& 	{10} \\
          learning rate & {1e-3} \\
          \bottomrule
		 \end{tabular}
		%
%
	
\end{threeparttable}
\end{table}
Table~\ref{tbl:supp:parameters} presents the hyperparameter configurations for all our methods.
$\alpha$ and $\lambda$ correspond to the trade-off parameter  
between the source and target classification losses as in Equation 4)
and the trade-off parameter between the classification and discriminator losses in Equation 13),
respectively.
$\alpha$ is only associated with \OurMethod and \OurMethodWithDisc;
whereas $\lambda$ is only associated with \OurMethodWithDisc and its variants.
We tried 3 different values for $d$ which represents the dimension of the compound representation out of 
\gnn and \newgnn; and
3 different values for the number of message passing steps $\tau$.
Hence, these two hyperparameters are associated with all methods except \morganFCN and \morgancFCN.
The hidden layer size in the attention network denoted by $\fatt(\cdot)$,
the feature-wise discriminator denoted by \ldisc, and
the compound-wise discriminator denoted by \gdisc
is fixed to 100.
We used a fixed batch size of 10 and an initial learning rate of 1e-3
with exponential decay every epoch from 1e-3 to 1e-4 at the end of training.
We trained each model for 40 epochs with an early-stopping criteria based on the \rocauc
performance on the validation set.
Specifically, during training, we evaluated the \rocauc performance of each model
on the validation set at every epoch; 
and we choose the trained model at some epoch $k$ that gives the best performance
on the validation set.
%

\subsection{Prediction Analysis}
\label{suppsec:pred_analysis}

Table~\ref{tbl:supp:TP_preds_analysis_TAc-c} presents the pairwise similarity analysis
for active compounds which were correctly classified by \OurMethodWithGlobalDisc,
but incorrectly classified as inactive by \OurMethod, \OurMethodWithDisc and its variants.

\begin{ThreePartTable}
\centering
\begin{TableNotes}
\item In this table, the columns source(S) and target(T) correspond to 
the protein accession numbers of the corresponding bioassays of source and target tasks, respectively. 
The columns \#cor, \#act and cor\%
have the count of correctly classified active compounds,
total number of active compounds and 
the percentage of correctly classified active compounds, respectively.
Such compounds are denoted by the set $\tilde{\SetCompounds}_{\scriptsize \bioassay_T}^{+}$.
The columns
$\text{sim}(\tilde{\SetCompounds}_{\scriptsize \bioassay_T}^{+}, \SetCompounds_{\scriptsize \bioassay_S}^{+})$ and
$\text{sim}(\tilde{\SetCompounds}_{\scriptsize \bioassay_T}^{+}, \SetCompounds_{\scriptsize \bioassay_T}^{+})$
present the average pairwise similarities 
of correctly classified compounds in $\tilde{\SetCompounds}_{\scriptsize \bioassay_T}^{+}$ 
with their top-5 most similar active compounds 
from the source and target bioassay, respectively.
The column \simDiff has the percentage difference of the 
average pairwise similarities from the source compounds
over the target compounds.
The column \eSimDiff has the average of element-wise percentage difference
of similarities from the source active compounds over the target active compounds.
The column \pvalue has the corresponding p-values for \eSimDiff.
\par
\end{TableNotes}

\begin{small}
\begin{longtable}[h]
		{
		 @{\hspace{1pt}}c@{\hspace{1pt}}
         @{\hspace{1pt}}c@{\hspace{0pt}}
         @{\hspace{0pt}}r@{\hspace{1pt}}
         @{\hspace{1pt}}r@{\hspace{1pt}}
         @{\hspace{1pt}}r@{\hspace{1pt}}
         @{\hspace{1pt}}r@{\hspace{1pt}}
         @{\hspace{1pt}}r@{\hspace{1pt}}
         @{\hspace{1pt}}r@{\hspace{1pt}}
         @{\hspace{1pt}}r@{\hspace{1pt}}
         @{\hspace{1pt}}r@{\hspace{1pt}}
         }
        \caption{Similarity analysis of correct predictions by \OurMethodWithGlobalDisc}
		\label{tbl:supp:TP_preds_analysis_TAc-c} \\
		\midrule
        	source(S)  &  target(T) & \#cor & \#act & cor\% &
        \footnotesize{$\text{sim}(\tilde{\SetCompounds}_{\tiny \bioassay_T}^{+}, \SetCompounds_{\tiny \bioassay_S}^{+})$}
        & \footnotesize{$\text{sim}(\tilde{\SetCompounds}_{\tiny \bioassay_T}^{+}, \SetCompounds_{\tiny \bioassay_T}^{+})$}
        	& \simDiff & \eSimDiff & \pvalue \\
        \midrule
        \endfirsthead
        \multicolumn{10}{c}%
        {{\bfseries \tablename\ \thetable{} -- continued from previous page}} \\
		\midrule
        source(S)  &  target(T) & \#cor & \#act & cor\% &
        \footnotesize{$\text{sim}(\tilde{\SetCompounds}_{\tiny \bioassay_T}^{+}, \SetCompounds_{\tiny \bioassay_S}^{+})$}
        & \footnotesize{$\text{sim}(\tilde{\SetCompounds}_{\tiny \bioassay_T}^{+}, \SetCompounds_{\tiny \bioassay_T}^{+})$}
        	& \simDiff & \eSimDiff & \pvalue \\
        \midrule
        \endhead
        \midrule
        \multicolumn{10}{|r|}{{Continued on next page}} \\
        \endfoot
        \midrule
        \insertTableNotes
        \endlastfoot
EAW86722 & AAI27629 & 36 & 128 & 28.125 & 0.231 & 0.194 & 19.401 & 21.692 & 1.68e-05 \\
NP\_004818 & ABD72211 & 131 & 485 & 27.010 & 0.191 & 0.218 & -12.557 & -12.312 & 3.05e-20 \\
NP\_000901 & NP\_001516 & 33 & 124 & 26.613 & 0.213 & 0.182 & 17.135 & 19.886 & 2.57e-05 \\
NP\_000483 & NP\_000918 & 18 & 82 & 21.951 & 0.262 & 0.444 & -41.090 & -32.693 & 1.56e-05 \\
NP\_004942 & NP\_003605 & 29 & 145 & 20.000 & 0.164 & 0.174 & -5.916 & -2.832 & 4.15e-01 \\
NP\_004221 & NP\_001516 & 21 & 124 & 16.935 & 0.212 & 0.203 & 4.530 & 6.081 & 2.45e-01 \\
AAC63054 & AAF04852 & 85 & 546 & 15.568 & 0.255 & 0.249 & 2.490 & 3.969 & 1.25e-01 \\
NP\_001516 & NP\_000947 & 15 & 112 & 13.393 & 0.244 & 0.193 & 26.449 & 28.965 & 2.92e-04 \\
EAW86722 & NP\_004358 & 6 & 46 & 13.043 & 0.239 & 0.144 & 66.156 & 66.256 & 5.06e-06 \\
AAH04460 & AAH14460 & 6 & 55 & 10.909 & 0.228 & 0.154 & 47.925 & 50.102 & 2.68e-05 \\
P00748 & NP\_004521 & 8 & 85 & 9.412 & 0.227 & 0.209 & 8.621 & 11.614 & 2.38e-01 \\
P00748 & AAF04852 & 50 & 546 & 9.158 & 0.237 & 0.267 & -11.053 & -8.667 & 1.22e-04 \\
P00748 & NP\_660205 & 57 & 634 & 8.991 & 0.243 & 0.242 & 0.662 & 1.593 & 6.55e-01 \\
NP\_004942 & NP\_000676 & 15 & 168 & 8.929 & 0.183 & 0.216 & -15.296 & -11.831 & 8.28e-03 \\
EAW86722 & NP\_000947 & 10 & 112 & 8.929 & 0.240 & 0.208 & 15.663 & 17.138 & 1.41e-01 \\
AAC63054 & AAI29989 & 55 & 638 & 8.621 & 0.234 & 0.249 & -6.064 & -3.701 & 1.69e-02 \\
NP\_001391 & NP\_001516 & 10 & 123 & 8.130 & 0.269 & 0.201 & 33.649 & 34.022 & 1.13e-06 \\
EAW86722 & AAH14970 & 8 & 99 & 8.081 & 0.266 & 0.212 & 25.838 & 28.634 & 1.58e-02 \\
AAI29989 & AAF04852 & 22 & 273 & 8.059 & 0.315 & 0.213 & 48.214 & 50.797 & 3.80e-10 \\
NP\_036559 & AAI29989 & 51 & 636 & 8.019 & 0.244 & 0.258 & -5.124 & -4.478 & 5.37e-03 \\
BAB91222 & AAH14970 & 8 & 101 & 7.921 & 0.175 & 0.165 & 5.943 & 8.653 & 3.60e-01 \\
P00748 & NP\_036559 & 5 & 67 & 7.463 & 0.271 & 0.184 & 47.440 & 43.716 & 8.58e-02 \\
NP\_004358 & AAH14970 & 7 & 98 & 7.143 & 0.207 & 0.167 & 24.161 & 23.704 & 9.37e-03 \\
AAA51985 & NP\_036559 & 5 & 70 & 7.143 & 0.226 & 0.167 & 35.308 & 35.516 & 1.30e-03 \\
BAB91222 & NP\_000901 & 5 & 71 & 7.042 & 0.252 & 0.166 & 51.322 & 52.216 & 2.77e-02 \\
NP\_004358 & NP\_001516 & 8 & 124 & 6.452 & 0.195 & 0.196 & -0.357 & 2.666 & 9.42e-01 \\
AAH14460 & AAC63054 & 11 & 176 & 6.250 & 0.240 & 0.237 & 1.395 & 4.762 & 8.31e-01 \\
NP\_003984 & EAW70217 & 7 & 112 & 6.250 & 0.176 & 0.207 & -15.381 & -10.736 & 2.51e-01 \\
NP\_001516 & NP\_000676 & 10 & 168 & 5.952 & 0.215 & 0.177 & 21.813 & 25.027 & 4.71e-02 \\
NP\_000900 & NP\_001391 & 16 & 273 & 5.861 & 0.207 & 0.187 & 10.512 & 13.962 & 5.45e-02 \\
EAW86722 & NP\_000901 & 4 & 70 & 5.714 & 0.211 & 0.153 & 37.288 & 39.267 & 3.66e-04 \\
NP\_004062 & NP\_005424 & 16 & 294 & 5.442 & 0.178 & 0.191 & -6.695 & -4.963 & 1.83e-01 \\
NP\_001516 & AAB26273 & 12 & 222 & 5.405 & 0.192 & 0.173 & 11.478 & 13.326 & 2.22e-02 \\
AAH14460 & P00748 & 6 & 118 & 5.085 & 0.209 & 0.176 & 19.259 & 19.193 & 9.09e-03 \\
ADZ17337 & NP\_000448 & 6 & 119 & 5.042 & 0.216 & 0.378 & -42.872 & -26.487 & 7.48e-02 \\
EAW86722 & AAB26273 & 11 & 222 & 4.955 & 0.171 & 0.172 & -0.639 & -0.201 & 9.25e-01 \\
AAB26273 & AAH14970 & 5 & 101 & 4.950 & 0.237 & 0.191 & 23.628 & 30.287 & 1.59e-01 \\
AAI28575 & ADZ17337 & 3 & 62 & 4.839 & 0.195 & 0.122 & 59.508 & 59.696 & 6.99e-03 \\
NP\_004358 & AAB26273 & 10 & 222 & 4.505 & 0.180 & 0.181 & -0.222 & 1.167 & 9.65e-01 \\
P00748 & NP\_004337 & 27 & 623 & 4.334 & 0.244 & 0.271 & -9.934 & -7.917 & 2.86e-03 \\
AAB26273 & NP\_000901 & 3 & 70 & 4.286 & 0.200 & 0.158 & 26.692 & 33.966 & 1.20e-01 \\
AAB26273 & NP\_000676 & 7 & 167 & 4.192 & 0.217 & 0.294 & -26.226 & -17.244 & 1.02e-01 \\
AAA51985 & P00748 & 4 & 98 & 4.082 & 0.254 & 0.185 & 37.324 & 38.576 & 3.26e-02 \\
AAA51985 & AAI29989 & 26 & 639 & 4.069 & 0.227 & 0.243 & -6.579 & -5.651 & 1.71e-02 \\
AAB26273 & EAW86722 & 11 & 280 & 3.929 & 0.253 & 0.359 & -29.579 & -7.283 & 1.26e-01 \\
NP\_036559 & AAF04852 & 21 & 545 & 3.853 & 0.224 & 0.237 & -5.495 & -3.310 & 2.10e-01 \\
AAH14460 & AAH04460 & 13 & 356 & 3.652 & 0.154 & 0.136 & 12.702 & 12.870 & 1.06e-04 \\
NP\_004358 & NP\_000900 & 3 & 83 & 3.614 & 0.211 & 0.181 & 16.860 & 15.934 & 1.37e-01 \\
NP\_000901 & EAW86722 & 10 & 280 & 3.571 & 0.251 & 0.503 & -50.099 & -46.770 & 1.25e-04 \\
NP\_004818 & NP\_000483 & 5 & 143 & 3.497 & 0.205 & 0.190 & 8.065 & 8.009 & 1.57e-01 \\
AAF04852 & AAC63054 & 6 & 179 & 3.352 & 0.318 & 0.215 & 48.069 & 49.310 & 5.66e-04 \\
AAI29989 & NP\_660205 & 12 & 395 & 3.038 & 0.287 & 0.191 & 49.869 & 52.166 & 3.98e-05 \\
NP\_775180 & ADZ17337 & 2 & 67 & 2.985 & 0.181 & 0.143 & 26.606 & 25.601 & 2.58e-01 \\
NP\_000448 & ADZ17337 & 2 & 68 & 2.941 & 0.114 & 0.060 & 91.779 & 95.379 & 7.93e-02 \\
ADZ17337 & NP\_066285 & 4 & 137 & 2.920 & 0.221 & 0.204 & 8.640 & 10.717 & 2.39e-01 \\
AAC63054 & NP\_004337 & 18 & 623 & 2.889 & 0.264 & 0.271 & -2.731 & 0.234 & 5.85e-01 \\
NP\_004358 & NP\_000901 & 2 & 71 & 2.817 & 0.154 & 0.123 & 25.345 & 26.218 & 3.06e-01 \\
AAF04852 & NP\_004337 & 8 & 287 & 2.787 & 0.321 & 0.218 & 47.407 & 50.401 & 6.15e-03 \\
AAB26273 & NP\_000947 & 3 & 108 & 2.778 & 0.232 & 0.158 & 46.717 & 52.124 & 7.21e-02 \\
AAB26273 & NP\_003605 & 4 & 144 & 2.778 & 0.219 & 0.151 & 44.746 & 43.328 & 3.17e-02 \\
NP\_000901 & BAB91222 & 2 & 73 & 2.740 & 0.266 & 0.584 & -54.493 & -54.476 & 2.78e-02 \\
AAH14970 & AAB26273 & 6 & 222 & 2.703 & 0.191 & 0.164 & 16.190 & 16.216 & 7.96e-02 \\
BAB91222 & AAB26273 & 6 & 222 & 2.703 & 0.160 & 0.167 & -4.540 & -3.455 & 2.74e-01 \\
BAB91222 & NP\_000947 & 3 & 112 & 2.679 & 0.206 & 0.214 & -3.745 & -2.894 & 7.53e-01 \\
P00748 & AAC63054 & 4 & 153 & 2.614 & 0.259 & 0.249 & 3.974 & 5.017 & 7.69e-01 \\
BAB91222 & NP\_001391 & 7 & 276 & 2.536 & 0.158 & 0.183 & -13.803 & -13.149 & 1.29e-02 \\
AAH14460 & NP\_036559 & 2 & 81 & 2.469 & 0.246 & 0.186 & 32.453 & 32.468 & 6.44e-02 \\
NP\_004337 & P00748 & 3 & 128 & 2.344 & 0.270 & 0.191 & 41.575 & 43.957 & 1.01e-02 \\
AAI29989 & NP\_036559 & 2 & 87 & 2.299 & 0.227 & 0.150 & 52.107 & 51.804 & 3.36e-01 \\
NP\_004337 & AAC63054 & 4 & 179 & 2.235 & 0.283 & 0.197 & 43.357 & 42.784 & 4.74e-03 \\
NP\_000760 & NP\_000752 & 17 & 828 & 2.053 & 0.265 & 0.197 & 34.518 & 40.100 & 4.00e-05 \\
AAA51985 & AAC63054 & 3 & 160 & 1.875 & 0.263 & 0.212 & 24.034 & 24.481 & 8.67e-03 \\
NP\_000903 & EAW86722 & 5 & 267 & 1.873 & 0.236 & 0.507 & -53.336 & -52.539 & 1.96e-03 \\
NP\_000900 & AAB26273 & 4 & 222 & 1.802 & 0.192 & 0.172 & 11.874 & 21.471 & 5.88e-01 \\
NP\_612200 & AAB26273 & 4 & 222 & 1.802 & 0.205 & 0.192 & 7.035 & 12.079 & 6.27e-01 \\
NP\_004705 & EAW70217 & 2 & 112 & 1.786 & 0.176 & 0.153 & 15.465 & 15.582 & 3.20e-03 \\
AAI29989 & AAC63054 & 3 & 179 & 1.676 & 0.275 & 0.195 & 40.963 & 43.788 & 1.15e-01 \\
EAW86722 & NP\_000903 & 2 & 121 & 1.653 & 0.201 & 0.119 & 68.483 & 73.823 & 1.13e-02 \\
BAH02301 & AAI28575 & 4 & 243 & 1.646 & 0.199 & 0.155 & 28.479 & 30.254 & 2.86e-02 \\
AAI29989 & P00748 & 2 & 128 & 1.563 & 0.307 & 0.187 & 63.821 & 68.477 & 2.76e-01 \\
NP\_660205 & P00748 & 2 & 128 & 1.563 & 0.223 & 0.199 & 12.042 & 18.287 & 3.47e-01 \\
NP\_000762 & NP\_000752 & 11 & 768 & 1.432 & 0.252 & 0.223 & 12.713 & 16.400 & 7.20e-02 \\
BAB91222 & EAW86722 & 4 & 280 & 1.429 & 0.212 & 0.395 & -46.403 & -45.265 & 1.19e-02 \\
NP\_000918 & NP\_000483 & 2 & 143 & 1.399 & 0.181 & 0.167 & 8.313 & 8.153 & 2.06e-01 \\
NP\_066285 & AAI28575 & 3 & 244 & 1.230 & 0.224 & 0.175 & 28.138 & 28.295 & 3.79e-02 \\
P00748 & AAI29989 & 7 & 638 & 1.097 & 0.234 & 0.325 & -27.930 & -17.446 & 1.11e-01 \\
AAB26273 & NP\_001391 & 3 & 274 & 1.095 & 0.236 & 0.220 & 7.039 & 8.095 & 7.11e-01 \\
NP\_004942 & NP\_997055 & 19 & 2,020 & 0.941 & 0.174 & 0.233 & -25.354 & -24.636 & 1.43e-07 \\
NP\_002721 & NP\_001387 & 2 & 216 & 0.926 & 0.157 & 0.262 & -40.099 & -15.747 & 5.69e-01 \\
NP\_001387 & NP\_065717 & 3 & 411 & 0.730 & 0.303 & 0.358 & -15.284 & -11.201 & 3.77e-01 \\
AAF04852 & AAI29989 & 2 & 278 & 0.719 & 0.233 & 0.190 & 22.866 & 23.111 & 5.88e-02 \\
NP\_000947 & EAW86722 & 2 & 280 & 0.714 & 0.219 & 0.166 & 31.671 & 31.738 & 6.30e-04 \\
NP\_004337 & AAI29989 & 2 & 303 & 0.660 & 0.272 & 0.211 & 28.856 & 28.826 & 6.91e-02 \\
NP\_004705 & NP\_005021 & 50 & 8,158 & 0.613 & 0.163 & 0.249 & -34.671 & -33.500 & 5.32e-14 \\
NP\_000752 & NP\_000760 & 3 & 543 & 0.552 & 0.299 & 0.247 & 21.330 & 23.816 & 3.93e-01 \\
NP\_036559 & NP\_004337 & 2 & 622 & 0.322 & 0.252 & 0.246 & 2.153 & 3.971 & 8.75e-01 \\
NP\_002825 & NP\_004081 & 2 & 1,316 & 0.152 & 0.261 & 0.254 & 2.678 & 2.762 & 8.02e-03
\end{longtable}
\end{small}
\end{ThreePartTable}



\subsection{Parameter Study: \OurMethodWithDMPNNAWithDisc}
\label{suppsec:param_study}

\begin{figure*}[h!]
    \centering
    \begin{subfigure}{0.32\textwidth}
        \centering
        \includegraphics[width=\textwidth]{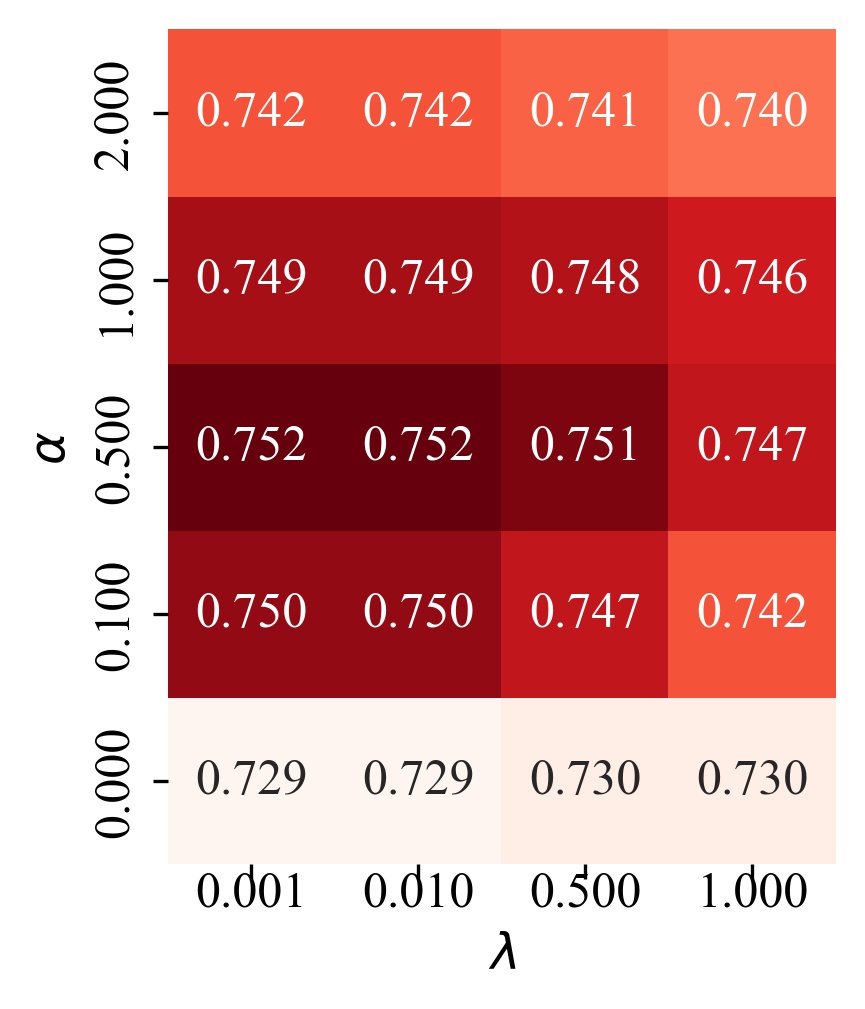}
        \caption{\prauc}
    \end{subfigure}
     ~ 
    \begin{subfigure}{0.32\textwidth}
        \centering
        \includegraphics[width=\textwidth]{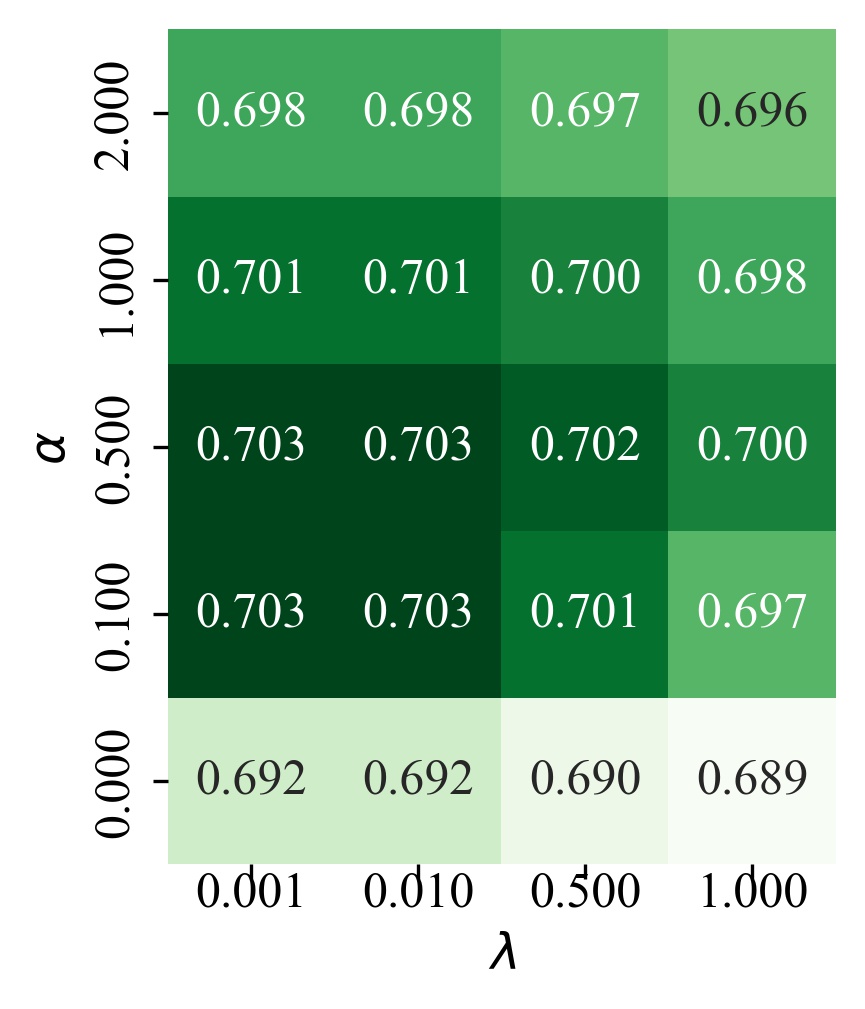}
        \caption{\precision}
    \end{subfigure}
     ~
    \begin{subfigure}{0.32\textwidth}
        \centering
        \includegraphics[width=\textwidth]{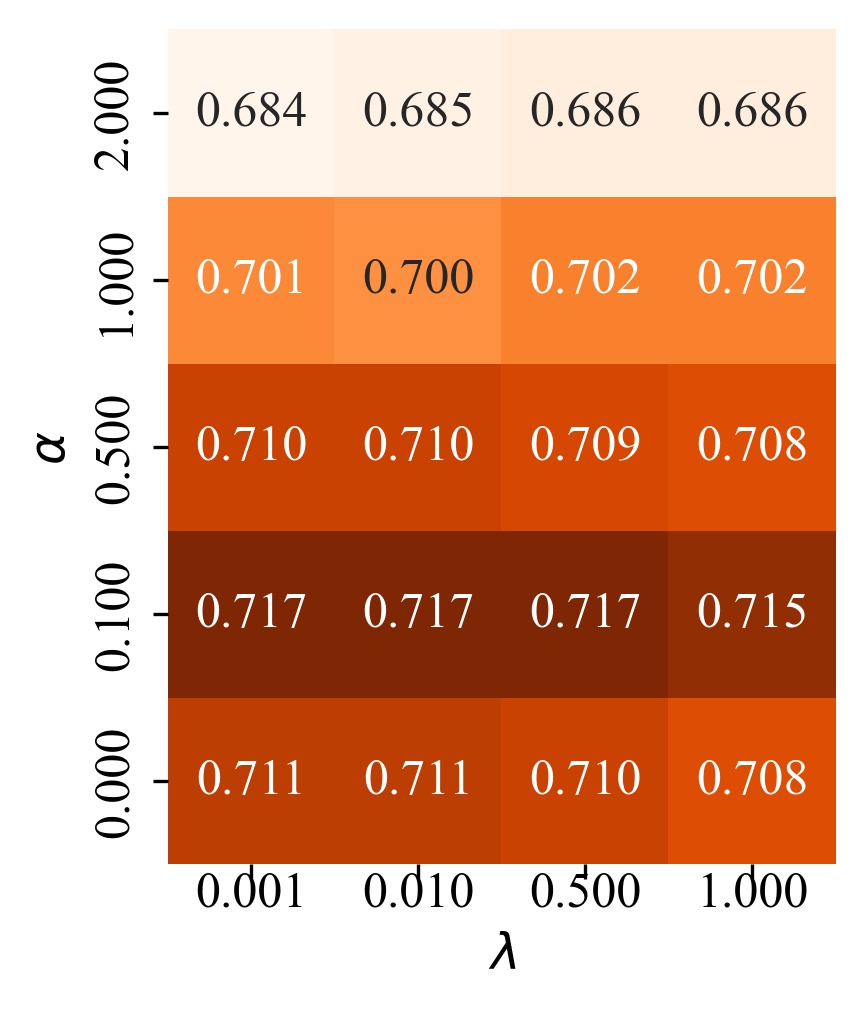}
        \caption{\sensitivity}
    \end{subfigure}%
    
    \begin{subfigure}{0.32\textwidth}
        \centering
        \includegraphics[width=\textwidth]{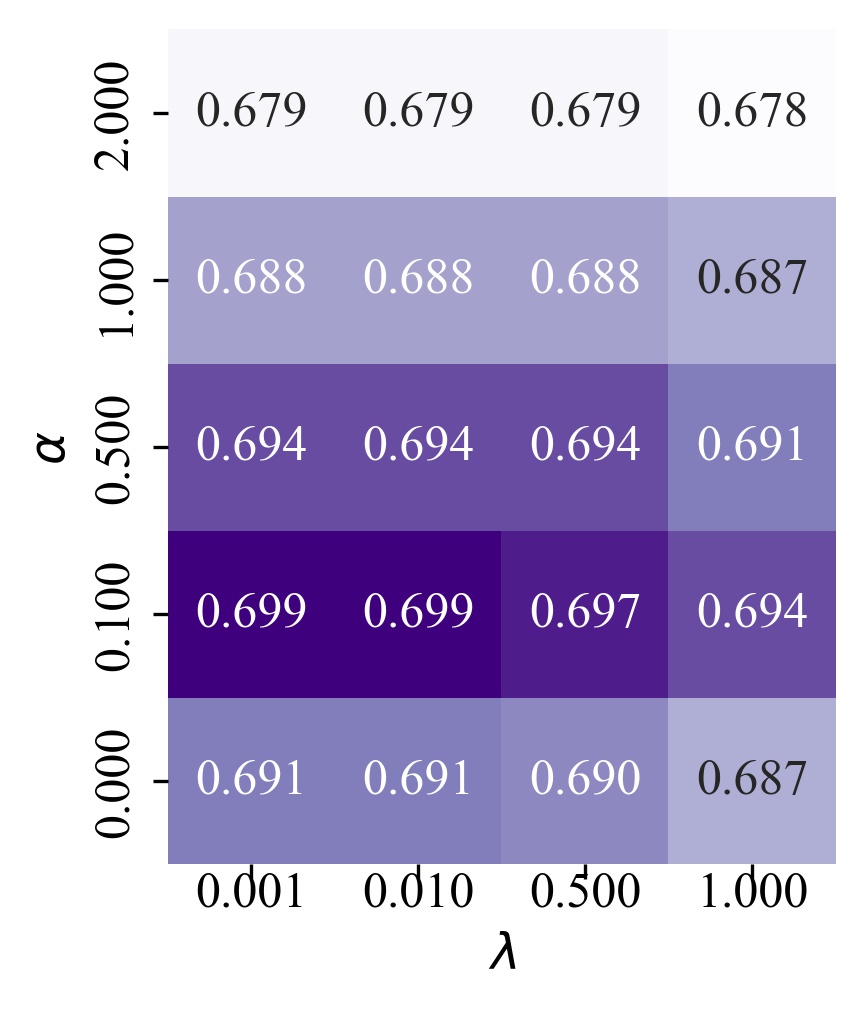}
        \caption{\accuracy}
    \end{subfigure}
     ~ 
    \begin{subfigure}{0.32\textwidth}
        \centering
        \includegraphics[width=\textwidth]{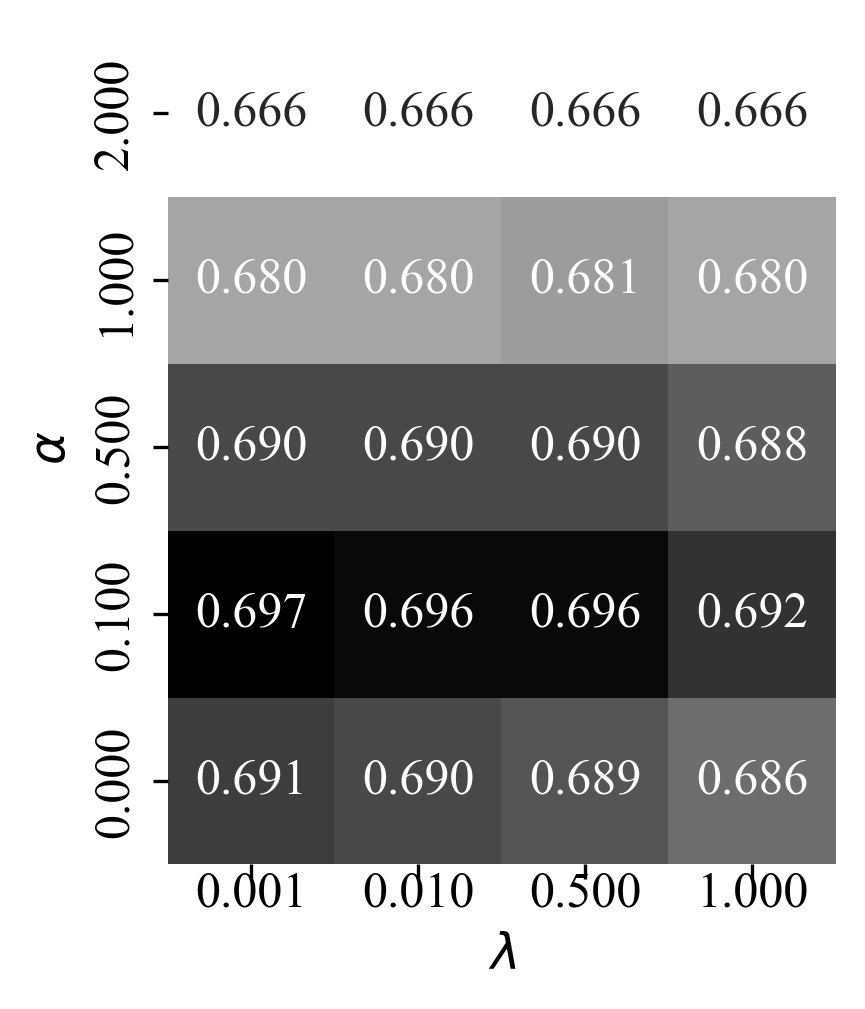}
        \caption{\fone score}
    \end{subfigure}
    \caption{Parameter Study of \OurMethodWithDMPNNAWithDisc}
    \label{supp:fig:param_best_tasks_aada_dmpnna_LG}
\end{figure*}

Figure~\ref{supp:fig:param_best_tasks_aada_dmpnna_LG} presents the parameter study
in \OurMethodWithDMPNNAWithDisc
in terms of \prauc, \precision, \sensitivity, \accuracy and \fone on $\alpha$ 
(i.e., the trade-off parameter  
between the source and target classification losses as in Equation 4)
and $\lambda$ (i.e., the trade-off parameter between the classification and discriminator losses in Equation 13).
The study was conducted over the tasks for which  \OurMethodWithDMPNNA outperforms the other methods in respective metrics. 
The values in each cell in the figure represent the average of the best performance 
over the tasks with the optimal choice of other hyperparameters.

Figure~\ref{supp:fig:param_best_tasks_aada_dmpnna_LG} shows that \OurMethodWithDMPNNAWithDisc
has the best performance in \prauc when
$\alpha = 0.5$ and $\lambda =$ 0.001 and 0.01.
This aligns with our observations in the parameter study for \rocauc.
\OurMethodWithDMPNNAWithDisc
has the best performance in \precision, \sensitivity, \accuracy and \fone when
$\alpha = 0.1$ and $\lambda =$ 0.001 and 0.01.
Although for the metrics except \rocauc and \prauc, 
the best performance is achieved at a lower $\alpha$
than 0.5, optimal $\alpha$ values are still non-zero.
This provides strong evidence that the target task benefits from the 
transferred information from the source task.
Similar to our observed trends from the parameter study for \rocauc,
there is a significant performance drop for too high or too low $\alpha$, 
regardless of what $\lambda$ is;
for optimal $\alpha$, a $\lambda =$ 0.001 or 0.01 gives the best performance;
for a given $\alpha$, higher $\lambda$ values degrades the performance.

\section{Compound Prioritization using \newgnn}
\label{suppsec:ranking:methods}
In this section, we develop a comprehensive learning-to-rank method 
for effective compound prioritization
that jointly learns molecular graph representations via \gnn 
and a scoring function $\phi(\cdot)$
using the representations in an end-to-end manner. 
We denote our method as \RankMethod.
We consider the compound prioritization problem to correctly rank compounds 
in terms of their activities with respect to a protein target. 
To achieve so, \RankMethod represents compounds using latent features that are learned 
from molecular graphs via a new, attention directed message passing neural network (\newgnn) (explained in Section 4.2.2).
%
We use a linear scoring function 
$\phi(\compr_{\scriptsize{\compound}}): \mathbb{R}^d \rightarrow \mathbb{R}$ 
to score the compounds as follows,
\begin{equation}
\label{equ:scoring}
\phi(\compr_{\scriptsize{\compound}}) = \mathbf{w}^\mathsf{T}\compr_{\scriptsize{\compound}},
\end{equation}
where $\mathbf{w}$ is a learnable parameter.
%
Our proposed method \RankMethod will produce a ranking of compounds 
induced by their predicted scores (computed using 
$\phi(\compr_{\scriptsize{\compound}})$).
Compounds with higher predicted scores will be ranked higher than those 
with lower predicted scores. 
Higher scores will be assigned to more active compounds 
in order to achieve the best ranking quality.
To quantify the ranking quality, we use the popular metric non-concordance index ($nCI$). 
$nCI$ measures the fraction of incorrectly ranked compound pairs as follows,
\begin{equation}
\label{equ:nci}
nCI(r, \phi) = \frac{1}{|\{c_i, c_j | c_i \succ_{\mathit{r}} c_j\}|} 
\sum_{\{c_i, c_j | c_i \succ_{\mathit{r}} c_j\}} \mathbb{I} (c_i \preceq_{\mathit{\phi}} c_j),
\end{equation}
where $\mathbb{I}$ is the indicator function.
In equation~\ref{equ:nci}, $c_i \succ_{\mathit{r}} c_j$ represents a pair of compounds 
$c_i$ and $c_j$ such that $c_i$ is ranked higher than $c_j$ 
in the ground truth ranking $\mathit{r}$, and
$c_i \preceq_{\mathit{\phi}} c_j$ represents that $c_i$ is ranked lower than $c_j$ in the predicted ranking structure induced by $\mathit{\phi}$.
In essence, a lower $nCI$ would indicate better ranking performance. 
Following the work~\cite{Liu2017Diff},
we use $nCI$ over the predicted ranking structure induced by $\mathit{\phi}$ as 
our loss function $\mathcal{L}_{rank}$.

Since the indicator function in equation~\ref{equ:nci} is discontinuous, 
we use the logistic loss as a surrogate function~\cite{Li2011} as follows,
\begin{equation}
\label{equ:surrogate}
\mathbb{I}(a \preceq b) \approx \log{(1 + \exp{(-(a-b))})}.
\end{equation}
The loss term $\mathcal{L}_{rank}$ for the set of compounds in a given bioassay 
is defined as follows,
\begin{equation}
\label{equ:rank_loss}
\begin{split}
\mathcal{L}_{rank} = \frac{1}{|\{c_i, c_j | c_i \succ_{\mathit{r}} c_j\}|} 
\sum_{\{c_i, c_j | c_i \succ_{\mathit{r}} c_j, c_i \prec_{\mathit{\phi}} c_j\}}
\log{[1 + \exp{(-(\mathbf{w}^{\mathsf{T}}\mathbf{z}_{c_i} - \mathbf{w}^{\mathsf{T}}\mathbf{z}_{c_j}))}]}.
\end{split}
\end{equation}
We solve the following optimization:
\begin{equation}
\label{equ:rank_objective}
\min_{\params} \mathcal{L}_{rank} + \lambda ||\params||_2^2,
\end{equation}
where $\lambda$ is the regularization parameter and $\params$ is the set of trainable parameters.
The above optimization encourages correct ranking of 
compound pairs and hence,
higher scores being assigned to more active compounds. 

\bibliography{ref_abbv}  